\documentclass[twoside,11pt]{article}

\usepackage{blindtext}

\usepackage[preprint]{jmlr2e}

\usepackage[utf8]{inputenc} 
\usepackage[T1]{fontenc}    
\usepackage{url}            
\usepackage{booktabs}       
\usepackage{amsfonts}       
\usepackage{nicefrac}       
\usepackage{microtype}      
\usepackage{enumitem}
\usepackage{xcolor}
\definecolor{gr}{HTML}{84bd00}
\definecolor{re}{HTML}{e4002b}
\definecolor{dg}{HTML}{DA1884}
\usepackage{acronym}
\acrodef{vdm}[VDM]{Video Diffusion Model}
\acrodef{arvdm}[AR-VDM]{Auto-Regressive Video Diffusion Model}
\acrodef{natvdm}[NAT-VDM]{Non-AuToregressive Video Diffusion Models}
\acrodef{natvdm}[NAT-VDM]{Non-AuTo-regressive VDM}
\acrodef{ar}[AR]{Auto-Regressive}
\acrodef{sde}[SDE]{Stochastic Differential Equation}
\acrodef{fvd}[FVD]{Fréchet Video Distance}
\acrodef{fid}[FID]{Fréchet Image Distance}
\acrodef{t2v}[T2V]{Text-To-Video}
\acrodef{srr}[SRR]{Successful Retrieval Rate}
\newcommand{\init}{{\text{init}}}
\newcommand{\ar}{{\text{ar}}}
\newcommand{\est}{{\text{est}}}

\usepackage{makecell}
\usepackage{subcaption}
\usepackage{bbm}
\usepackage{multirow}
\newtheorem{requirement}{Requirement} 
\usepackage{changepage}
\usepackage{wrapfig}
\definecolor{MAEblue}{RGB}{47 112 182}
\definecolor{SDEblue}{RGB}{28 58 88}
\definecolor{cc1}{rgb}{1.0, 0.44, 0.37}
\definecolor{cc2}{rgb}{0.0, 0.2, 0.6}
\definecolor{cc3}{RGB}{255, 191, 0}
\definecolor{cc4}{RGB}{0, 128, 128}
\definecolor{gred}{RGB}{219,68,55}
\hypersetup{
	colorlinks=true,
	urlcolor=magenta,
	citecolor=cc4,
	linkcolor=cc2,
	bookmarks=true,
	bookmarksnumbered=true,
	bookmarksopen=true,
	linktoc=all,
}

\usepackage{lastpage}
\jmlrheading{23}{2022}{1-\pageref{LastPage}}{1/21; Revised 5/22}{9/22}{21-0000}{Jing Wang, Fengzhuo Zhang, Xiaoli Li, Vincent Y. F. Tan, Tianyu Pang, Chao Du, Aixin Sun and Zhuoran Yang}

\ShortHeadings{AR-VDM}{Wang et al.}

\usepackage[mathscr]{eucal}
\usepackage[cmex10]{amsmath}
\usepackage{epsfig,epsf,psfrag}
\usepackage{amssymb,amsmath,amsfonts,latexsym}
\usepackage{amsmath,graphicx,bm,xcolor,url,overpic}
\usepackage{fixltx2e}
\usepackage{array}
\usepackage{verbatim}
\usepackage{bm}
\usepackage{algorithmic}
\usepackage{algorithm}
\usepackage{verbatim}
\usepackage{textcomp}
\usepackage{epstopdf}
\usepackage{acronym}
\usepackage{diagbox}
\usepackage{tikz}
\usetikzlibrary{graphs, positioning, quotes, shapes.geometric}

\newcommand{\openone}{\leavevmode\hbox{\small1\normalsize\kern-.33em1}} 

\catcode`~=11 \def\UrlSpecials{\do\~{\kern -.15em\lower .7ex\hbox{~}\kern .04em}} \catcode`~=13 

\allowdisplaybreaks[4]

\def\tv{\mathop{\mathrm{TV}}}
\newcommand{\unif}{{\text{Unif}}}
\newcommand{\kl}{\mathrm{KL}}

\newcommand{\calD}{\mathcal{D}}

\newcommand{\calG}{\mathcal{G}}
\newcommand{\calH}{\mathcal{H}}

\newcommand{\calL}{\mathcal{L}}

\newcommand{\calN}{\mathcal{N}}

\newcommand{\calP}{\mathcal{P}}

\newcommand{\calR}{\mathcal{R}}
\newcommand{\calS}{\mathcal{S}}

\newcommand{\rmA}{\mathrm{A}}

\newcommand{\rmd}{\mathrm{d}}

\newcommand{\rmE}{\mathrm{E}}

\newcommand{\rmI}{\mathrm{I}}

\newcommand{\rmO}{\mathrm{O}}

\newcommand{\rmR}{\mathrm{R}}

\newcommand{\bbE}{\mathbb{E}}

\newcommand{\bbN}{\mathbb{N}}

\newcommand{\bbR}{\mathbb{R}}

\DeclareMathAlphabet{\mathbsf}{OT1}{cmss}{bx}{n}
\DeclareMathAlphabet{\mathssf}{OT1}{cmss}{m}{sl}

\DeclareSymbolFont{bsfletters}{OT1}{cmss}{bx}{n}  
\DeclareSymbolFont{ssfletters}{OT1}{cmss}{m}{n}
\DeclareMathSymbol{\bsfGamma}{0}{bsfletters}{'000}
\DeclareMathSymbol{\ssfGamma}{0}{ssfletters}{'000}
\DeclareMathSymbol{\bsfDelta}{0}{bsfletters}{'001}
\DeclareMathSymbol{\ssfDelta}{0}{ssfletters}{'001}
\DeclareMathSymbol{\bsfTheta}{0}{bsfletters}{'002}
\DeclareMathSymbol{\ssfTheta}{0}{ssfletters}{'002}
\DeclareMathSymbol{\bsfLambda}{0}{bsfletters}{'003}
\DeclareMathSymbol{\ssfLambda}{0}{ssfletters}{'003}
\DeclareMathSymbol{\bsfXi}{0}{bsfletters}{'004}
\DeclareMathSymbol{\ssfXi}{0}{ssfletters}{'004}
\DeclareMathSymbol{\bsfPi}{0}{bsfletters}{'005}
\DeclareMathSymbol{\ssfPi}{0}{ssfletters}{'005}
\DeclareMathSymbol{\bsfSigma}{0}{bsfletters}{'006}
\DeclareMathSymbol{\ssfSigma}{0}{ssfletters}{'006}
\DeclareMathSymbol{\bsfUpsilon}{0}{bsfletters}{'007}
\DeclareMathSymbol{\ssfUpsilon}{0}{ssfletters}{'007}
\DeclareMathSymbol{\bsfPhi}{0}{bsfletters}{'010}
\DeclareMathSymbol{\ssfPhi}{0}{ssfletters}{'010}
\DeclareMathSymbol{\bsfPsi}{0}{bsfletters}{'011}
\DeclareMathSymbol{\ssfPsi}{0}{ssfletters}{'011}
\DeclareMathSymbol{\bsfOmega}{0}{bsfletters}{'012}
\DeclareMathSymbol{\ssfOmega}{0}{ssfletters}{'012}

\newcommand{\tilB}{\tilde{B}}

\newcommand{\hatP}{\hat{P}}

\newcommand{\tilP}{\tilde{P}}

\newcommand{\tilt}{\tilde{t}}

\newcommand{\tilX}{\tilde{X}}

\newcommand{\tilY}{\tilde{Y}}

\newcommand{\bart}{\bar{t}}

\newcommand{\barT}{\bar{T}}

\DeclareMathOperator*{\argmin}{argmin}

\def\##1\#{\begin{align}#1\end{align}}
\def\$#1\${\begin{align*}#1\end{align*}}

\usepackage{thmtools, thm-restate}

\newtheorem{fact}{Fact}
\newtheorem{assumption}{Assumption}
\usepackage{amsmath,amssymb,amsfonts}
\usepackage{algorithmic}
\usepackage{graphicx}
\usepackage{textcomp}
\usepackage{xcolor,hyperref}
\def\BibTeX{{\rm B\kern-.05em{\sc i\kern-.025em b}\kern-.08em
    T\kern-.1667em\lower.7ex\hbox{E}\kern-.125emX}}

\begin{document}

\title{Error Analyses of Auto-Regressive Video Diffusion Models}

\author{\name Jing Wang$^{1,2,*}$ \email jing005@e.ntu.edu.sg \\
       \addr $^{1}$Nanyang Technological University, $^{2}$A$^{*}$STAR
       \AND
       \name Fengzhuo Zhang$^{3,*,\dag}$ \email fzzhang@u.nus.edu \\
       \addr $^{3}$National University of Singapore
       \AND
       \name Xiaoli Li$^{6}$ \email xiaoli\_li@sutd.edu.sg \\
       \addr $^{6}$Singapore University of Technology and Design
       \AND
       \name Vincent Y. F. Tan$^{3}$ \email vtan@nus.edu.sg \\
       \addr $^{3}$National University of Singapore
       \AND
       \name Tianyu Pang$^{4}$ \email tianyupang@sea.com \\
       \addr $^{4}$Sea AI Lab
       \AND
       \name Chao Du$^{4,\ddag}$ \email duchao@sea.com \\
       \addr $^{4}$Sea AI Lab
       \AND
       \name Aixin Sun$^{1}$ \email axsun@ntu.edu.sg \\
       \addr $^{1}$Nanyang Technological University
       \AND
       \name Zhuoran Yang$^{5}$ \email zhuoran.yang@yale.edu \\
       \addr $^{5}$Yale University}

\footnotetext[1]{$*$ Equal contribution}
\footnotetext[2]{$\dag$ Project Lead}
\footnotetext[3]{$\ddag$ Corresponding author}
\footnotetext[4]{Work done by Jing Wang and Fengzhuo Zhang as associate members at Sea AI Lab}

\editor{My editor}

\maketitle

\begin{abstract}
Auto-Regressive Video Diffusion Models (AR-VDMs) have shown strong capabilities in generating long, photorealistic videos, but suffer from two key limitations: (i) \emph{history forgetting}, where the model loses track of previously generated content, and (ii) \emph{temporal degradation}, where frame quality deteriorates over time. Yet a rigorous theoretical analysis of these phenomena is lacking, and existing empirical understanding remains insufficiently grounded. In this paper, we introduce \texttt{Meta-ARVDM}, a unified analytical framework that studies both errors through the shared autoregressive structure of AR-VDMs. We show that history forgetting is characterized by the conditional mutual information between the generated output and preceding frames, conditioned on inputs, and prove that incorporating more past frames \emph{monotonically alleviates} history forgetting, thereby theoretically justifying a common belief in existing works. Moreover, our theory reveals that standard metrics fail to capture this effect, motivating a new evaluation protocol based on a ``needle-in-a-haystack'' task in closed-ended environments (DMLab and Minecraft). We further show that temporal degradation can be quantified by the cumulative sum of per-step errors, enabling prediction of degradation for different schedulers without video rollout. Finally, our evaluation uncovers a strong empirical correlation between history forgetting and temporal degradation, a connection not previously reported.

\end{abstract}

\section{Introduction}
The \ac{arvdm} has emerged as a particularly effective approach for generating long
videos~\citep{kim2024fifo,henschel2024streamingt2v}. By modeling video data in an
\ac{ar} manner, these models capture complex motion patterns and rich textures that
closely reflect real-world dynamics. Their versatility enables a wide range of
applications, including video game synthesis~\citep{che2024gamegen}, world modeling~\citep{chen2024diffusion},
and robotic perception and planning~\citep{zhang2024autoregressive}.

\begin{figure}
  \begin{center}
    \includegraphics[width=1.0\textwidth]{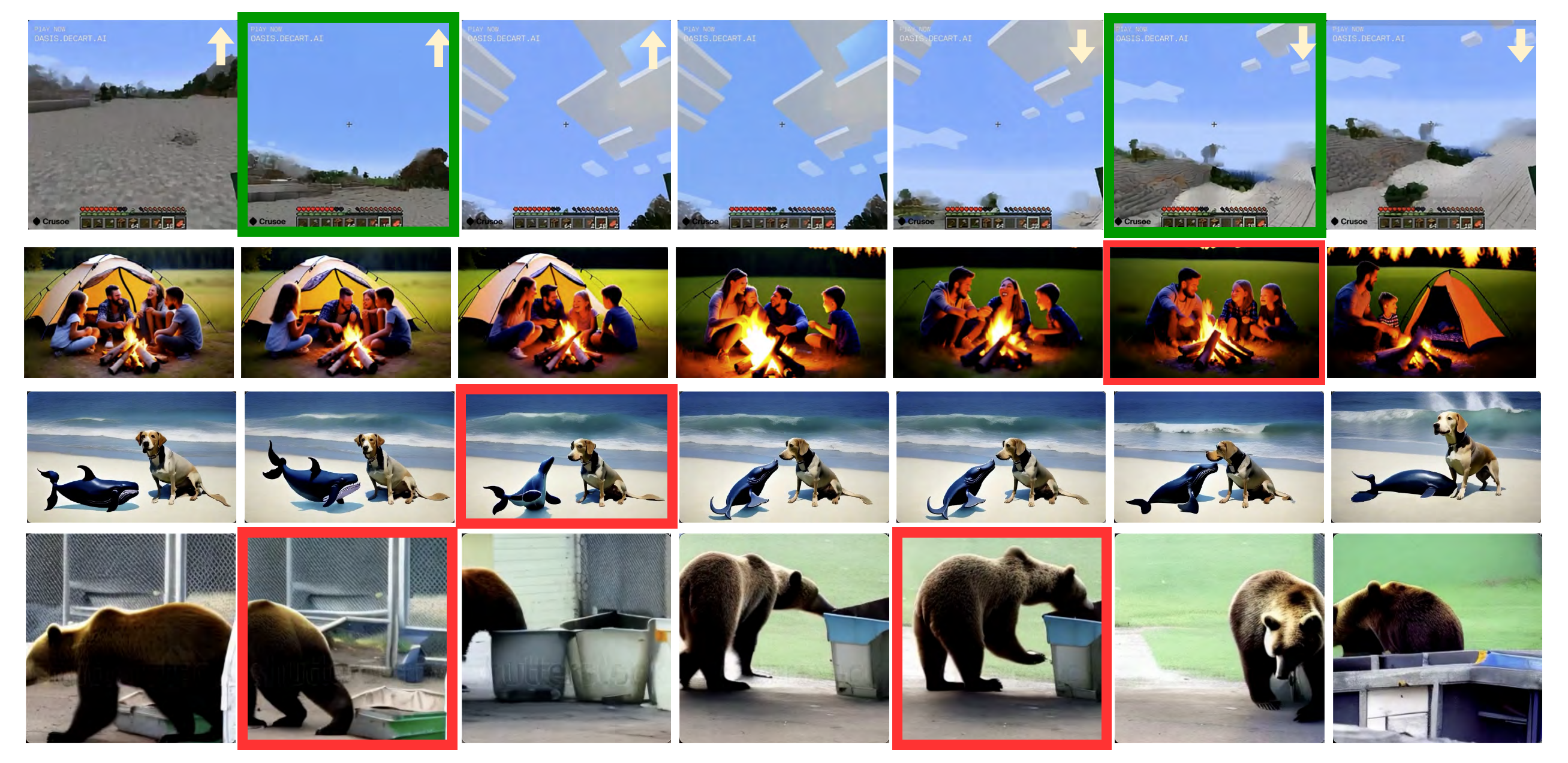}
  \end{center}
  \vspace{-0.9em}
  \caption{Example frames sampled from videos that are longer than $100$ frames.
  We using color of bounding boxes to differentiate {\color[HTML]{009900} History Forgetting}
  and {\color[HTML]{FF6666} Temporal Degradation}. } 
  \label{fig:ar}
\end{figure}

However, all \ac{arvdm}s exhibit two limitations not typically seen in \acp{natvdm}: 
(i) \textit{history forgetting}, where the model loses track of previously generated
content (e.g., subjects or backgrounds), as illustrated in the first row of Fig.~\ref{fig:ar};
and (ii) \textit{temporal degradation}, where video quality progressively
deteriorates along the temporal axis. This degradation is reflected in the last three
rows of Fig.~\ref{fig:ar}. These issues significantly limit the utility of \ac{arvdm}s
for world modeling and realistic video generation. Yet their origins, mathematical
formulations, and evaluations remain underexplored, hindering the development of effective
mitigation strategies. While early studies decompose \ac{natvdm} errors into noise
initialization, score estimation, and discretization errors~\citep{chen2023improved},
it remains unclear whether history forgetting and temporal degradation fall within
this framework or necessitate novel characterizations of various  error terms.



In this paper, we present a unified framework, \texttt{Meta-ARVDM}, specifically designed for \ac{arvdm}s. This framework facilitates a principled analysis, demonstrating that phenomena such as history forgetting and temporal degradation are intrinsic to the auto-regressive characteristics of these models, rather than mere byproducts of particular implementation choices. \texttt{Meta-ARVDM} encompasses all \ac{arvdm}s that meet our proposed auto-regressive requirement (Requirement~\ref{req:ar_step}), which is largely satisfied by most existing models.

Through a comprehensive mathematical error analysis of \texttt{Meta-ARVDM}, we show that history forgetting is captured by the conditional mutual information between the output and the full history given the input—a novel term absent from prior analyses. We prove that the \emph{emergence} of this term is inevitable in \ac{arvdm}s, however, its magnitude can be  reduced in a monotonic fashion by injecting more historical context into the denoising process. This formalization implies that standard distributional metrics (e.g., \ac{fvd}, \ac{fid}) and motion-based metrics (e.g., smoothness, motion magnitude) fail to characterize history forgetting. To remedy this problem, we propose a visual ``needle-in-a-haystack'' task across varying history lengths that both reveals and quantifies this phenomenon, verifying that it can mitigate the history forgetting problem in a monotonic fashion. Following this new task setting, we empirically show that some existing methods with good performances on traditional measures fail to reduce history forgetting.



Finally, we demonstrate that temporal degradation arises from the cumulative impact of noise initialization, score estimation, and discretization errors. Our theoretical analysis is robustly validated by experimental results, which, for example, accurately forecast the performance of various schedulers. The relationship between temporal degradation and these three error components suggests that metrics sensitive to them can effectively quantify temporal degradation. Furthermore, our evaluations reveal a previously unreported correlation between history forgetting and temporal degradation.

In summary, our contributions are two-fold. 
\begin{itemize}
    \item We derive the error analysis of \ac{arvdm}s within a general framework, \texttt{Meta-ARVDM}. Our analysis quantitatively characterizes history forgetting as a novel conditional mutual information term, and temporal degradation as the cumulative accumulation of errors propagated from preceding frames. These results provide formal theoretical justifications for the common beliefs in empirical studies. Specifically, for history forgetting, we prove that this error is statistically inevitable but can be monotonically mitigated by incorporating past frames. For temporal degradation, our analysis explains why later-generated video segments exhibit lower quality than earlier ones, even when they share the same length.

    \item Our theoretical results yield new empirical insights into history forgetting and temporal degradation. For history forgetting, our results imply that commonly used distributional metrics (e.g., FVD, FID) and motion-based metrics (e.g., smoothness) fail to capture this error. In contrast, our proposed visual “needle-in-a-haystack” task effectively quantifies the extent of history forgetting. For temporal degradation, we show that standard distributional metrics can evaluate it reliably, and our theoretical quantification accurately predicts the performance of different step schedulers without generating videos. Leveraging these improved evaluations, we further uncover a correlation between history forgetting and temporal degradation—a relationship not previously reported in the literature.
\end{itemize}
\section{Related Work}
\label{sec:related_works}

\textbf{Video Diffusion Models}  Given the remarkable successes of diffusion models in image generation, several early works in \ac{t2v} have proposed treating the entire video as a vector to learn joint score functions~\citep{ho2022video, blattmann2023stable, ma2024latte, chen2024videocrafter2, guo2023animatediff, singer2022make, yuan2024instructvideo}. To effectively capture the correlations between frames, these studies have developed a temporal attention module that applies attention mechanisms across the temporal dimension. The overall architecture integrates both temporal and spatial attention layers~\citep{wang2023modelscope, blattmann2023align, wang2024lavie, xing2025dynamicrafter}.
Furthermore, to leverage the fine-grained spatial-temporal relationships between pixels, some researchers have adopted 3D attention modules as foundational components of denoising networks~\citep{lin2024open, zheng2024open, yang2024cogvideox}. Beyond the use of 3D attention, additional works have investigated the combination of Mamba and attention mechanisms~\citep{gao2024matten, mo2024scaling}.
In addition to these T2V approaches, video generation can also be conditioned on various factors, such as images~\citep{zhang2023i2vgen, chen2023seine, ren2024consisti2v}, poses~\citep{karras2023dreampose, ma2024follow}, motion~\citep{chen2023motion, wang2024motionctrl}, and sound~\citep{liu2023generative}.


\noindent \textbf{Auto-Regressive Video Models}  A body of research has emerged that utilizes the \ac{ar} framework to generate long videos. These studies employ two main approaches: adapting diffusion models to the AR framework and tokenizing frames for next-token prediction. We will first discuss examples of the former approach.
FIFO-Diffusion~\citep{kim2024fifo} is a training-free method that leverages pretrained models to denoise frames at various noise levels. To address the training-inference gap, this method uses latent partitioning to reduce discrepancies between noise levels. Other notable works, including AR-Diffusion~\citep{wu2023ar}, Rolling Diffusion Models~\citep{ruhe2024rolling}, Diffusion-forcing~\citep{chen2024diffusion}, and Pyramidal Flow Matching~\citep{jin2024pyramidal}, train networks specifically to denoise frames at different noise levels. In contrast, ART-V~\citep{weng2024art}, StreamingT2V~\citep{henschel2024streamingt2v}, GameNGen~\citep{valevski2024diffusion}, Self-forcing~\citep{huang2025self}, and GameGen-X~\citep{che2024gamegen} focus on denoising frames that share the same noise level. All the above methods rely on denoised or noisy frames together with the KV cache as context. Another line of work focuses on selective compression of memory to improve generation quality, including Mixture of Contexts~\citep{cai2025mixture} and FramePack~\citep{zhang2025frame}.

The second approach involves tokenizing frames and training networks for next-token prediction. Examples include LlamaGen~\citep{sun2024autoregressive}, Emu3~\citep{wang2024emu3}, and Loong~\citep{wang2024loong}, which generate visual tokens based on their spatial-temporal ranking. Additionally, WorldMem~\citep{WorldMem} seeks to implement a memory mechanism by incorporating historical frames via the attention mechanism, demonstrating effectiveness in video lengths of around 10 seconds within Minecraft scenarios. Context-as-Memory~\citep{ContextAsMemory} further enhances this by using historical context frames as memory, employing simple concatenation and a memory retrieval module to select relevant frames based on camera field of view (FOV) overlap. Lastly, MemoryPack~\citep{MemoryPack} builds on this memory mechanism, extending it to a hierarchical structure to ensure both short-term consistency and long-term retention.
Our work specifically analyzes the methods belonging to the first approach.

\section{Preliminaries: Video Diffusion Models}\label{sec:prelim}
A \ac{vdm} generates video frames by reversing a forward diffusion process~\citep{ho2022video,ho2022imagen}. We represent a video $X_{1:w}$  as a sequence of frames along the temporal axis $(X_1, \ldots, X_w) \in \mathbb{R}^{w \times d}$, where $w$ is the number of frames and $d$ is the dimensionality of each frame (e.g., $d = 512 \times 512$ for $512 \times 512$-pixel frames). Starting from the clean video $X_{1:w}^0 = X_{1:w}$, the diffusion process gradually adds Gaussian noise, as described by the following process.
\begin{align}
    \rmd X_{1:w}^{t}=f(X_{1:w}^{t},t)\, \rmd t+g(t)\, \rmd B^{t} \text{ for } t\in[0,T],\label{eq:forward}
\end{align}
where $t$ denotes the time index of the diffusion process, $X_{1:w}^{t}$ denotes the noisy version of video frames at diffusion time $t$, and the  functions $f:\bbR^{w\times d}\times \bbR\rightarrow\bbR^{w\times d}$ and $g:\bbR\rightarrow\bbR$ govern the drift and diffusion coefficients, and $B^{t}$ is  $(w\times d)$-dimensional Brownian motion with identity covariance. Throughout, we use subscripts for frame indices and superscripts for diffusion time steps. To avoid confusion, we refer to $t$ as the diffusion time (i.e., noise level) and $w$ as the frame index. 

As discussed in~\cite{song2020score}, two common instantiations of Eqn.~\eqref{eq:forward} 
are the variance-exploding SDE (SMLD)~\citep{song2019generative} with $f=0$, $g(t)=\beta(t)$, and the variance-preserving SDE (DDPM)~\citep{ho2020denoising} with $f(X,t) = -0.5 \beta(t) X$, $g(t) = \sqrt{\beta(t)}$, where $\beta(t)$ is a scalar noise schedule. In both cases, $X_{1:w}^{T}$ converges in distribution to a Gaussian distribution as $T \to \infty$.
To generate samples, we reverse the diffusion from $t = T$ to $t = 0$, which yields the reverse-time SDE~\citep{anderson1982reverse}:
\begin{align}
  \rmd X_{1:w}^{t}& = \big( f(X_{1:w}^{t},t) - g(t)^{2}\, \nabla \log  P_{t}(X_{1:w}^{t} ) \big)\,\rmd t + g(t)\,\rmd \tilB^{t} \text{ for }t\in[0,T],\label{eq:rev_1}
\end{align}
where $\tilB^{t}$ is \emph{reverse-time} Brownian motion,   $P_t$ is the distribution of $X_{1:w}^{t}$,  $\nabla\log P_{t}(X_{1:w}^{t})$ is known as the {\em  score function}, which is typically learned using neural networks~\citep{ho2022video}. By applying the time reversal $t \mapsto T - t$, we obtain the equivalent forward-time representation of the reverse process as
\begin{align}
    &\rmd \tilX_{1:w}^{t} = \big( - f(\tilX_{1:w}^{t},T - t) + g(T- t)^{2}\nabla\log \!P_{T-t}(\tilX_{1:w}^{t})\big)\,\rmd t +g(T-t)\,\rmd B^{t}. \label{eq:rev}
\end{align} 
We will write $\tilX^t$ (resp.\ $\tilB^t$) and $X^t$ (resp.\ $B^t$) to denote the reverse-time and original stochastic processes (Brownian motion), respectively. We will focus on this form of the reverse-time process for the theoretical analysis of our work.

\begin{figure*}[t]
\centering
 \includegraphics[width=1.0\textwidth]{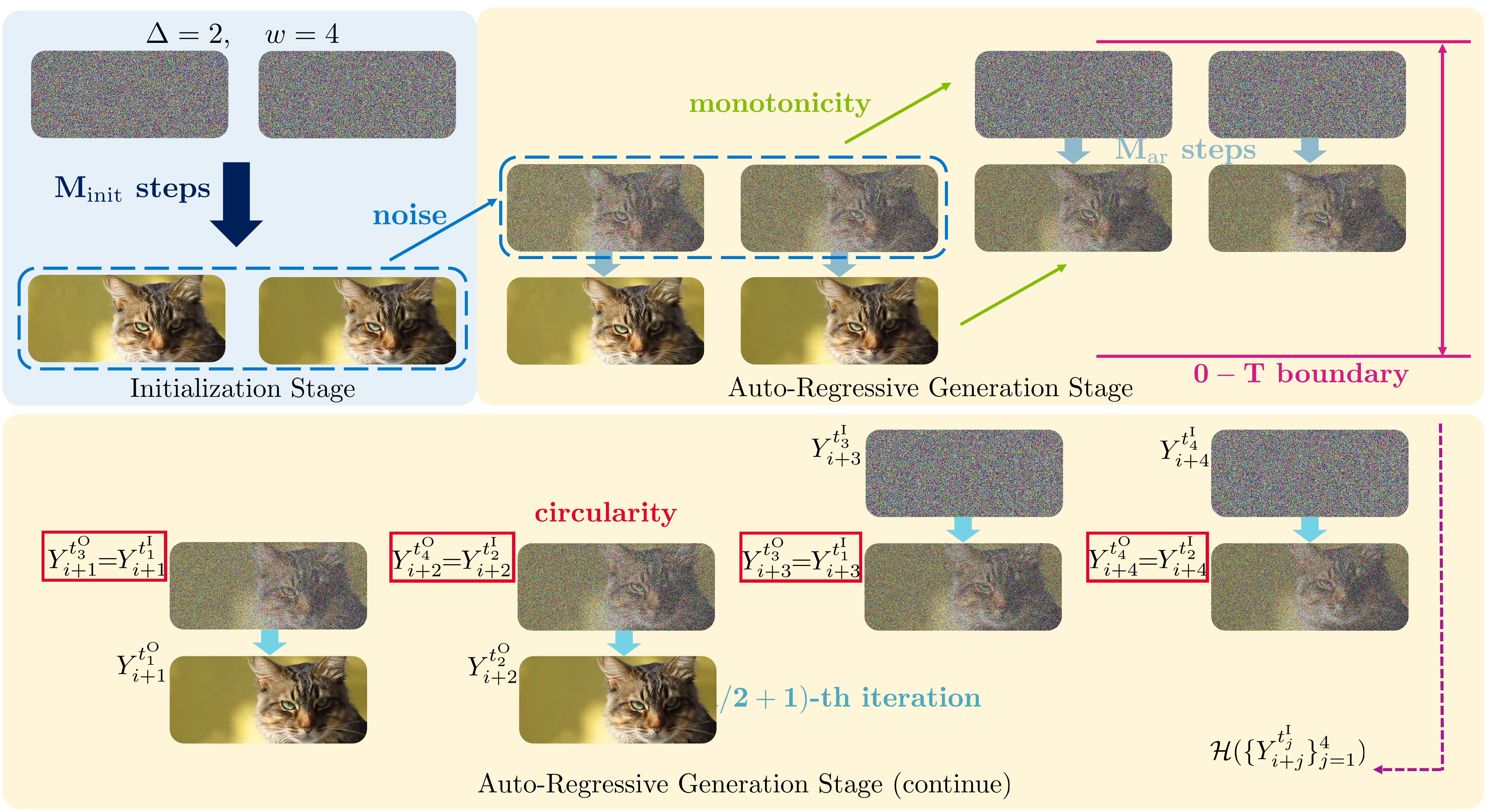}
 
 \vspace{-0.5em}
\caption{The \texttt{Meta-ARVDM} framework begins with an initialization stage (depicted on the left), which denoises the input using $M_{\init}$ steps. Noise is then added to the output of this stage to produce the starting point for the \ac{ar} generation stage (shown on the right).
The figure highlights key requirements for plausible implementation: \textcolor{gr}{monotonicity}, \textcolor{re}{circularity}, and the \textcolor{dg}{$0$--$T$ boundary} condition. Here, $\calH(\cdot)$ denotes all frames available prior to the execution of the $(i/2 + 1)$-th iteration.}
\label{fig:framework}
\end{figure*}
 \begin{algorithm}[t]
	\caption{\texttt{Meta-ARVDM}}\label{algo:framework}
        \textbf{Input:} \ac{vdm} $s_{\theta}$, \ac{ar} step-size $\Delta\in\bbN$, effective window size $w\in\bbN$, length of the video $N\in\bbN$, input noise levels $\calL^{\rmI}=(t_{1}^{\rmI},\cdots,t_{w}^{\rmI})$, output noise levels $\calL^{\rmO}=(t_{1}^{\rmO},\cdots,t_{w}^{\rmO})$, reference frames sets $\{\calR_{i}\}_{i\in[N]}$\\
	\textbf{Procedure:}
	\begin{algorithmic}[1]
        \STATE $(Y_{1}^{t_{1}^{\rmI}},\cdots,Y_{w-\Delta}^{t_{w-\Delta}^{\rmI}})$ = \texttt{Initialization}$(s_{\theta},\calL^{\rmI})$. {\color{blue} // Initialization Stage}\label{line:init}
        \FOR{$k=1,\ldots,\lceil N/\Delta\rceil$}
            \STATE \textcolor{white}{r}
            
            \vspace{-2.6em}
            \begin{align*}
            &Y_{(k-1)\Delta+1:k\Delta}^{0}, (Y_{k\Delta+1}^{t_{1}^{\rmI}},\cdots,Y_{k\Delta+w-\Delta}^{t_{w-\Delta}^{\rmI}})\ \ \text{\color{blue} // Auto-Regressive Generation Stage} \nonumber\\
            &\qquad=\texttt{Auto-Regressive Step}\big((Y_{(k-1)\Delta+j}^{t_{j}^{\rmI}})_{j=1}^{w-\Delta},s_{\theta},w,\Delta,\calL^{\rmI},\calL^{\rmO},Y_{\calR_{(k-1)\Delta+1}}^{0}\big)
            \end{align*}

            \vspace{-1em}
        \ENDFOR
        \STATE Output $Y_{1:N}^{0}$.
	\end{algorithmic}
\end{algorithm}

\noindent\emph{Notation:} For a comprehensive list of all mathematical symbols used in this paper, please refer to Appendix~\ref{app:notation}.

\section{\texttt{Meta-ARVDM}: A Unified Framework of AR-VDMs}
\label{sec:framework}

\ac{arvdm}s generate video frames sequentially according to their temporal positions. In contrast to the approach suggested in Eqn.~\eqref{eq:rev}, which denoises all frames simultaneously at the same noise level $t$, \ac{arvdm}s enable the denoising of frames at varying noise levels. Importantly, the generation of each frame depends on the frames that have been generated previously. We will present \ac{arvdm}s in a cohesive framework, utilizing outpainting and FIFO-Diffusion~\citep{kim2024fifo} as illustrative examples. 


Algorithm~\ref{algo:framework} provides an overview of our framework, \texttt{Meta-ARVDM}. This meta-algorithm operates in two stages. The first stage, \texttt{Initialization} (Algorithm\ref{algo:init}), utilizes a \ac{vdm} to generate $i_{0}$ frames, with all frames being denoised at the same noise level. The second stage, \texttt{Auto-Regressive Step} (detailed in Algorithm~\ref{algo:ar_step}), denoises $w$ frames at varying noise levels. 
During each \ac{ar} step, the algorithm takes as input reference frames in $\calR$, $w - \Delta$ past noisy frames, and $\Delta$ Gaussian noise instances. It outputs $\Delta$ clean frames alongside $w - \Delta$ noisy frames for subsequent iterations. The input and output noise levels are denoted as $\calL^{\rm I} = (t_{1}^{\rm I}, \ldots, t_{w}^{\rm I})$ for input levels and $\calL^{\rm O} = (t_{1}^{\rm O}, \ldots, t_{w}^{\rm O})$ for output levels. 
To initialize the \ac{ar} step, we introduce different noise levels to the $i_{0}$ frames generated in the initial stage. We will now formally define this framework to facilitate further analysis. While the following notations may appear complex, they are essential for ensuring mathematical rigor.\footnote{Readers not interested in the technical
details may skip this section and proceed directly to Requirement~\ref{req:ar_step},
which specifies the conditions of \ac{arvdm} satisfying the proposed framework.} In the following, we will explain the two stages   via two examples: outpainting and FIFO-Diffusion. Other \ac{arvdm}s covered by the framework are explained in Appendix~\ref{app:other_methods}.

The initialization stage adopts a \ac{vdm} to generate $i_{0}$ frames via the
Euler--Maruyama scheme.
\begin{align}
     & \rmd \tilX_{1:w}^{t}=\big( - f( X_{1:w}^{\tilt_{n}^{\init}},T - \tilt_{n}^{\init}) + g(T-\tilt_{n}^{\init})^{2}s_{\theta}( \tilX_{1:w}^{\tilt_{n}^{\init}},T -\tilt_{n}^{\init})\big)\rmd t +g(T-\tilt_{n}^{\init})\rmd B^{t}\label{eq:init}
\end{align}
for $t\in[\tilt_{n}^{\init},\tilt_{n+1}^{\init}]$ with $n=0,\ldots,M_{\init}-1$,
where the partition $0 = t_{0}^{\init}\leq \cdots \leq t_{M_{\init}}^{\init}= T$
divides $[0, T]$ into $M_{\init}$ intervals, and
$\tilt_{n}^{\init}= T - t_{M_{\init} - n}^{\init}$ denotes the reversed noise
levels.
\begin{algorithm}
    [t] \small
    \caption{\texttt{Initialization}}
    \label{algo:init} \textbf{Input:} \ac{vdm} $s_{\theta}$, size of the
    effective window $w\in\bbN$, input noise levels $\calL^{\rmI}$\\ \textbf{Procedure:}
    \begin{algorithmic}
        [1] \STATE Generate $i_{0}\geq w$ independent Gaussian vectors $Y_{1:i_0}
        ^{T}$.
        \STATE Denoise $Y_{1:i_0}^{T}$ to $Y_{1:i_0}^{0}$ with Eqn.~\eqref{eq:init}.
        \STATE Add noises to $Y_{1:i_0}^{0}$ to derive
        $\{Y_{i}^{t_{i}^{\rmI}}\}_{i=1}^{w-\Delta}$.
        \STATE Return $\{Y_{i}^{t_{i}^{\rmI}}\}_{i=1}^{w-\Delta}$.
    \end{algorithmic}
\end{algorithm}
The score function $\nabla \log P$ is estimated by a neural network $s_{\theta}$
with parameters $\theta$. This process is illustrated in Figure~\ref{fig:framework}
(left). In outpainting, this initialization corresponds to generating the first
block of videos. In FIFO-Diffusion, this corresponds to generating the initial
$w$ frames (comprising several blocks) and injecting Gaussian noise at varying
levels.

The \emph{auto-regressive generation stage} in methods such as FIFO-Diffusion and StreamingT2V simultaneously denoises frames at various noise levels, often using additional reference frames. To address this, we generalize the ground-truth reverse equation (Eqn.~\eqref{eq:rev_1}) along two dimensions. 
First, to accommodate multiple noise levels, we introduce the extended time vector $\bart(t) = (t + \delta_{1}, \ldots, t + \delta_{w})$, where each $\delta_{i} \in \bbR$ represents the offset of the $i$-th frame's noise level relative to the shared evolution variable $t$. For clarity, we will refer to $\bart(t)$ simply as $t$ when the context is unambiguous.
Second, we condition the initial distribution of $X_{1:w}^{0}$ in Eqn.~\eqref{eq:forward} on the reference clean frames $X_{\calR}^{0}$, where $\calR \subseteq \bbN$ denotes the set of indices of the reference frames. The extended reverse-time SDE can then be written as
\begin{align}
     & \rmd \tilX_{1:w}^{\bart}=\big(-f(\tilX_{1:w}^{\bart},\barT-\bart) +g(\barT-\bart)^{2}\nabla\log P_{X}^{\barT-\bart}(\tilX_{1:w}^{\bart}|X_{\calR}^{0})\big)\rmd t +g(\barT-\bart)\rmd B_{1:w}^{\bart}\label{eq:x_reverse}
\end{align}
for $t\in[T-t^{\rmI},T-t^{\rmO}]$, where 
$\barT=(T,\ldots,T)$, and $w$ is the \ac{ar} \emph{window size}. The frame set
$X_{1:w}^{\bart}$ represents frames at varied noise levels, i.e., $X_{1:w}^{\bart}
=(X_{1}^{t+\delta_{1}},\ldots,X_{w}^{t+\delta_{w}})=\{X_{i}^{t+\delta_{i}}\}_{i=1}
^{w}$ and $P_{X}^{\bart}$ is the distribution of $X_{1:w}^{\bart}$. The functions $f:\bbR
^{w\times d}\times \bbR^{w}\rightarrow\bbR^{w\times d}$ and $g:\bbR^{w}\rightarrow
\bbR$ determine the evolution of the noisy frames.

Assuming all frames evolve at the same rate for
variable rates, we fix input and output noise levels respectively as
$\calL^{\rmI}=(t_{1}^{\rmI},\ldots,t_{w}^{\rmI})$ and
$\calL^{\rmO}=(t_{1}^{\rmO},\ldots,t_{w}^{\rmO})$, and define
$\delta_{i}=t_{i}^{\rmI}-t_{1}^{\rmI}=t_{i}^{\rmO}-t_{1}^{\rmO}$ (see Appendix~\ref{app:notation} for detailed notation). 
Then each \ac{ar} step approximates Eqn.~\eqref{eq:x_reverse} via the following Euler--Maruyama
scheme
\begin{align}
     & \rmd \tilY_{1:w}^{\bart}=\big(-f(\tilY_{1:w}^{\bar{\tilt}_{n}^{\ar}},\barT-\bar{\tilt}_{n}^{\ar})\label{eq:multi_reverse_score}+g(\barT-\bar{\tilt}_{n}^{\ar})^{2}s_{\theta}(\tilY_{1:w}^{\bar{\tilt}_{n}^{\ar}},\barT-\bar{\tilt}_{n}^{\ar},Y_{\calR}^{0})\big) \,\rmd t \!+\! g(\barT-\bar{\tilt}_{n}^{\ar})\,\rmd B_{1:w}^{\bart}
\end{align}
for $t\in[\tilt_{n}^{\ar},\tilt_{n+1}^{\ar}]$ with $n=0,\ldots,M_{\ar}-1$. Similar
to Eqn.~\eqref{eq:init}, $\tilt_{n}^{\ar}=T-t_{M_{\ar}-n}^{\ar}$ is the reverse
noise level, and $\bar{\tilt}_{n}^{\ar}=\bart(\tilt_{n}^{\ar})$ is the reverse noise
levels of all frames induced by $\tilt_{n}^{\ar}$. The full denoising schedule
is discretized over $[t_{1}^{\rmO}, t_{1}^{\rmI}]$ into $M_{\ar}$ intervals. For
ease of notation, we denote the denoising neural network in both initialization
and \ac{ar} stages as $s_{\theta}$. The score estimate $s_{\theta}$ takes the
current noisy frames $\tilY_{1:w}^{\bart(\tilt_{n}^{\ar})}$, current noise level
$T-\tilt_{n}^{\ar}$, and the reference frames $Y_{\calR}^{0}$ as inputs. This process
is implemented in the \texttt{Auto-Regressive Step} module.

\begin{algorithm}
    [t] \small
    \caption{\texttt{Auto-Regressive Step}}
    \label{algo:ar_step} \textbf{Input:} \ac{vdm} $s_{\theta}$, size of the
    effective window $w\in\bbN$, \ac{ar} step-size $\Delta\in\bbN$, input noise levels
    $\calL^{\rmI}$, output noise levels $\calL^{\rmO}$, input noisy video $\{Y_{i}
    ^{t_{i}^{\rmI}}\}_{i=1}^{w-\Delta}$, reference frames $Y_{\calR}^{0}$\\ \textbf{Procedure:}
    \begin{algorithmic}
        [1] \STATE Generate $\Delta$ independent Gaussian vectors $\{Y_{i}^{t_{i}^{\rmI}}
        \}_{i=w-\Delta+1}^{w}$. \label{line:noise_init} \STATE Concatenate
        $\{Y_{i}^{t_{i}^{\rmI}}\}_{i=1}^{w-\Delta}$ with the generated noise to
        form $\{Y_{i}^{t_{i}^{\rmI}}\}_{i=1}^{w}$. \STATE Denoise $\{Y_{i}^{t_{i}^{\rmI}}
        \}_{i=1}^{w}$ to $\{Y_{i}^{t_{i}^{\rmO}}\}_{i=1}^{w}$ with the reference
        frames $Y_{\calR}^{0}$ as Eqn.~\eqref{eq:multi_reverse_score}.\label{line:denoise}
        \STATE Return $\!\{Y_{i}^{t_{i}^{\rmO}}\}_{i=1}^{w}\!=\!\big(\{Y_{i}^{t_{i}^{\rmO}}
        \}_{i=1}^{\Delta},\!\{Y_{i}^{t_{i}^{\rmO}}\}_{i=\Delta+1}^{w}\big)$
    \end{algorithmic}
\end{algorithm}

The \ac{ar} generation stage consists of iterative calls to \texttt{Auto-Regressive
Step} (Algorithm~\ref{algo:ar_step}), as shown on the right side of Figure~\ref{fig:framework}.
For outpainting that utilizes $12$ already generated frames to denoise $4$ frames,
the effective window size is $4$. For FIFO-Diffusion, the latent partitioning enables
the \ac{vdm} network to denoise $n \cdot f$ frames concurrently, giving an
effective window of $w = n \cdot f$. Each \ac{ar} step jointly denoises frames
from noise levels $\calL^{\rmI}= (t_{1}^{\rmI}, \ldots, t_{w}^{\rmI})$ to
$\calL^{\rmO}= (t_{1}^{\rmO}, \ldots, t_{w}^{\rmO})$. To ensure coherent \ac{ar}
generation, we impose specific requirements on $\calL^{\rmI}$, $\calL^{\rmO}$, and
the reference set $\calR$.
\begin{requirement}
    {(\ac{ar} Requirements)}\label{req:ar_step}
    \begin{itemize}[parsep=0pt, topsep=0pt, leftmargin=10pt]
        \item (Monotonicity) The noise levels are monotone, i.e.,
            $t_{i-1}^{\rmI}\leq t_{i}^{\rmI}$, $t_{i-1}^{\rmO}\leq t_{i}^{\rmO}$,
            and $t_{i}^{\rmO}<t_{i}^{\rmI}$ for $i\in[w]$.

        \item ($0-T$ Boundary) The boundaries of $\calL^{\rmI}$ and $\calL^{\rmO}$
            are $T$ and $0$, respectively, i.e., $t_{i}^{\rmI}=T$ for $i\geq w-\Delta
            +1$, and $t_{i}^{\rmO}=0$ for $i\leq \Delta$.

        \item (Circularity) The output noise levels are the same as the input noise
            levels up to a position shift, i.e.,
            $t_{i}^{\rmI}=t_{\Delta+i}^{\rmO}$ for all $i\in[w-\Delta]$.

        \item (Constant Pace) The difference between the input and output noise levels
            are constant, i.e.,
            $t_{i}^{\rmO}-t_{i}^{\rmI}=t_{j}^{\rmO}-t_{j}^{\rmI}$ for all
            $i,j\in[w]$.

        \item (Causality) When denoising frames indexed from $i+1$ to $i+w$, the
            reference frames set is a subset of $[i]$, i.e., $\calR_{i}\subseteq[
            i]$ for all $i\in\bbN$.
    \end{itemize}
\end{requirement}

Figure~\ref{fig:framework} illustrates these conditions for $w=4,\Delta=2$.
\emph{Monotonicity} ensures that earlier frames (with smaller index $i$) are assigned
lower noise levels $t_{i}^{\rmI}$ and $t_{i}^{\rmO}$ during denoising. The \emph{$0
-T$
boundary} requirement ensures that the input to \texttt{Auto-Regressive Step}
contains $\Delta$ new frames, i.e., there are $\Delta$ Gaussian vectors, and
that the output contains $\Delta$ clean frames. \emph{Circularity} ensures that
the output of one step can serve as input to the next, enabling seamless iteration.
The \emph{constant pace} constraint is introduced to simplify theoretical analysis
and is relaxed at the end of this section. The \emph{causality} condition is essential
for practical generation, ensuring that current frames are not conditioned on
future ones. Both outpainting and FIFO-Diffusion satisfy these conditions. For
outpainting, we have $t_{i}^{\rmI}= T$ and $t_{i}^{\rmO}= 0$ for all $i \in [ w]$,
with $\Delta = w$, denoising $w$ noise vectors into clean frames per step. For
FIFO-Diffusion, we have $\Delta=1$, and
$(0,t_{1}^{\rmI},\ldots,t_{w-1}^{\rmI})=(t_{1}^{\rmO},\ldots,t_{w-1}^{\rmO},T)$
are set to the noise levels of a pre-chosen scheduler. With all conditions met,
each iteration of \texttt{Auto-Regressive Step} takes $\Delta$ new noise vectors
and the previous outputs as input, producing $\Delta$ clean frames via Eqn.~\eqref{eq:multi_reverse_score}. 
\looseness=-1

We can also consider an extension of the constant pace and scalar function $g(\cdot)$  from Eqn.~\eqref{eq:x_reverse}. Specifically, we can extend the constant pace setting to accommodate various paces  by introducing a new definition, namely that of the  extended time $   \bart(t) = (\alpha_{1}t + \delta_{1}, \alpha_{2}t + \delta_{2}, \ldots, \alpha_{w}t + \delta_{w}) $. In this expression, the coefficients $  \alpha_{i} $  for $  i \in [w] $  represent the different paces applied to each frame. To extend the scale function $  g(\bart) $  into matrix form $  G(X_{1:2}, \bart) $, we will utilize the same approach as presented in \citep{song2020score}. In fact, we have,
\begin{align*}
    \rmd X_{1:w}^{\bart} & =\Big(f(X_{1:w}^{\bart},\bart)-\nabla\cdot [G(X_{1:w}^{\bart},\bart)G(X_{1:w}^{\bart},\bart)^{\top}]                                                                                      \\
                         & \qquad\quad -G(X_{1:w}^{\bart},\bart)G(X_{1:w}^{\bart},\bart)^{\top}\nabla\log P_{X}^{\bart}(X_{1:w}^{\bart}|X_{\calR}^{0})\Big)\rmd t +G(X_{1:w}^{\bart},\bart)\rmd \tilB_{1:w}^{\bart}.
\end{align*}


\section{Error Analysis of AR-VDMs through
\texorpdfstring{\texttt{Meta-ARVDM}}{Meta-ARVDM}}
\label{sec:analysis}

All algorithms encompassed by \texttt{Meta-ARVDM} utilize a shared analytical framework to analyze their errors. Building on previous theoretical analyses of diffusion processes~\citep{chen2022sampling,chen2023improved,chen2023score}, we adopt a simplified model for clarity, specifically choosing $f(\tilY_{1:w}^{t}, t) = -0.5 \cdot \tilY_{1:w}^{t})$ and $g(t) = 1$. This corresponds to a DDPM model characterized by a constant $\beta(t) = 1$, as discussed in Section~\ref{sec:prelim}. Our findings can be easily generalized to accommodate other configurations. For our analysis, we impose three mild assumptions: the boundedness of pixel values, Lipschitz continuity of the score function, and a bounded score estimate error. These assumptions are standard in diffusion theory.

\begin{assumption}
    [Boundness of Pixel Values]\label{assump:bound} For the frames of size
    $l=\max\{\Delta,i_{0}\}$, the $\ell_{2}$-norm of $X_{i+1:i+l}^{0}$ is almost
    surely bounded by a constant $B > 0$, i.e., $\|X_{i+1:i+l}^{0}\|_{2}\leq B$
    a.s. for any $i \in \bbN$.
\end{assumption}
In latent diffusion models~\citep{chen2024videocrafter2,guo2023animatediff}, frames
are denoised in  the latent space, where pixel values are quantized and therefore
bounded. For diffusion models in the pixel space, the pixel values are bounded by $255$. Thus, this assumption is satisfied by most \ac{vdm}s. 

\begin{assumption}
    [Lipschitz continuity of Score Function]\label{assump:lips} For any
    $t\in[0,T]$, both $\nabla \log P_{X}^{t}$ and $\nabla\log P_{X}^{\bart(t)}$ are
    $L$-Lipschitz with respect to the $\ell_{2}$-norm.
\end{assumption}
This assumption controls the discretization error by ensuring that the score function
does not change too rapidly. It is standard in numerical analysis and diffusion
theory~\citep{gautschi2011numerical,suli2003introduction,chen2023improved}.

\begin{assumption}[Score Estimate Error]\label{assump:err}
    The average score estimation error evaluated at the discretization steps is upper bounded by $\epsilon_{\est}^{2}>0$ for both the initialization and \ac{ar} stages, i.e., the following hold  for any $k\in\bbN$:
    \begin{align*}
        &T^{-1}\sum_{n=1}^{M_{\init}}(t_{n}^{\init}-t_{n-1}^{\init})\bbE\Big[\big\|\nabla\log P_{X}^{t_{n}^{\init}}(X_{1:i_0}^{t_{n}^{\init}})-s_{\theta}(X_{1:i_0}^{t_{n}^{\init}},t_{n}^{\init})\big\|^{2}\Big]\leq \epsilon_{\est}^{2}\qquad\mbox{and}\\
        & (t_{1}^{\rmI}-t_{1}^{\rmO})^{-1}\sum_{n=1}^{M_{\ar}}(t_{n}^{\ar}-t_{n-1}^{\ar})\bbE\Big[\big\|\nabla\log P_{X}^{\bart_{n}^{\ar}}(X_{k\Delta+1:k\Delta+w}^{\bart_{n}^{\ar}}\,|\,X_{\calR_{k\Delta+1}}^{0}) \notag\\
        &\qquad-s_{\theta}(X_{k\Delta+1:k\Delta+w}^{\bart_{n}^{\ar}},\bart_{n}^{\ar},X_{\calR_{k\Delta+1}}^{0})\big\|^{2}\Big]\leq \epsilon_{\est}^{2}.
    \end{align*}
\end{assumption}
These conditions quantify the discrepancies between the true and estimated score functions along
the discretized diffusion trajectory, stemming from the finite training of the
denoising networks. Similar assumptions are widely used in theoretical analyses of
diffusion models~\citep{chen2022sampling,chen2024probability}. We note that the estimation error of the diffusion models is averaged along the whole trajectory of the diffusion process. To state our results concisely, we write $x \lesssim y$ to mean that $x \leq C y$ for some absolute constant $C > 0$ and write $\kl(X\,\|\,Y)$ for the KL divergence between the distributions of $X$ and $Y$.

\begin{theorem}
    \label{thm:main} Under regularity assumptions (Assumptions~\ref{assump:bound},
    \ref{assump:lips}, and \ref{assump:err}),
    the KL-divergence between the distributions of the video generated by \texttt{Meta-ARVDM}
    and the nominal video is
    \begin{align}
        \kl(X_{1:K\Delta}^{0}\,\|\,Y_{1:K\Delta}^{0})= \rmI\rmE + \sum_{k=1}^{K}\rmA\rmR\rmE_{k}. 
        \label{res:joint}
    \end{align}
    Here the error of the initialization stage $\rmI\rmE$ and the error of $k$-th
    \ac{ar} step $\rmA\rmR\rmE_{k}$ are respectively bounded as follows.
    \begin{align}
        \rmI\rmE & \lesssim \mathrm{NIE}+ \mathrm{SEE}+ \mathrm{DE}\quad\mbox{and}\quad \rmA\rmR\rmE_{k}\lesssim \mathrm{NIE}_{\rmA\rmR}+ \mathrm{SEE}_{\rmA\rmR}+ \mathrm{DE}_{\rmA\rmR}+ \mathrm{HF}_{k}
    \end{align}
    where the Noise Initialization Errors (NIEs) are $\mathrm{NIE}:= (d i_{0}+ B^{2})\exp(-T
    )$ and $\mathrm{NIE}_{\rmA\rmR}:=(\Delta d+ B^{2})\exp(-T)$, the Score
    Estimation Errors (SEEs) are $\mathrm{SEE}:= T\epsilon_{\est}^{2}$ and
    $\mathrm{SEE}_{\rmA\rmR}:= (t_{1}^{\rmI}-t_{1}^{\rmO})\epsilon_{\est}^{2}$, the
    Discretization Errors (DEs) are $\mathrm{DE}:= wdL^{2}\sum_{n=1}^{M_{\init}}(t_{n}^{\init}
    -t_{n-1}^{\init})^{2}$ and $\mathrm{DE}_{\rmA\rmR}:= wdL^{2}\sum_{n=1}^{M_{\ar}}
    (t_{n}^{\ar}-t_{n-1}^{\ar})^{2}$, and finally, the history forgetting error is
    \begin{align}
        \mathrm{HF}_{k}:= I\big( \texttt{Output}_{k}; \texttt{Past}_{k}\big|\texttt{Input}_{k}\big).
    \end{align}
    Here, the $\texttt{Output}_{k}$, $\texttt{Past}_{k}$, and $\texttt{Input}_{k}$ are
    defined as $
    \texttt{Output}_{k}= \{X_{k\Delta+j}^{t_{j}^{\rmO}}\}_{j=1}^{w}, \texttt{Past
    }_{k}=\calH(\{X_{k\Delta+j}^{t_{j}^{\rmO}}\}_{j=1}^{w})\backslash\{X_{\calR_{k\Delta+1}}
    ^{0}\}$, and
    $\texttt{Input}_{k}= \{X_{\calR_{k\Delta+1}}^{0}\}\cup \{X_{k\Delta+j}^{t_{j}^{\rmI}}
    \}_{j=1}^{w}.
    $
    The history $\calH(\cdot)$ is formally defined in Appendix~\ref{app:main_proof},
    and an example is shown in Figure~\ref{fig:framework}. If $\calR=\emptyset$,
    the KL-divergence between the distributions of the $K$-th video clip generated
    by \texttt{Meta-ARVDM} and the nominal video is
    \begin{align}
         & \kl\big( X_{K\Delta+1:(K+1)\Delta}^{0}\big\| Y_{K\Delta+1:(K+1)\Delta}^{0}\big)\lesssim \rmI\rmE + K\big[ \mathrm{NIE}_{\mathrm{AR}}+ \mathrm{SEE}_{\mathrm{AR}}+ \mathrm{DE}_{\mathrm{AR}}\big].\label{res:recursive}
    \end{align}
\end{theorem}
The proof is provided in Appendix~\ref{app:main_proof}. This theorem
characterizes the KL-divergence of the errors for both the generated long videos (Eqn.~\eqref{res:joint})
and the short clips (Eqn.~\eqref{res:recursive}). The error for long videos comprises
contributions from both the initialization stage (Algorithm~\ref{algo:init}) and
the \ac{ar} step (Algorithm~\ref{algo:ar_step}). Each stage introduces three types
of error: noise initialization, score estimation, and discretization. These error
sources also arise in image diffusion models~\citep{chen2022sampling,chen2023score},
and are further discussed in Appendix~\ref{app:err_discussion}. In the remainder
of this section, we focus on how these errors  specifically relate to \ac{arvdm}s.



\noindent \textbf{Effect of history forgetting.} In Eqn.~\eqref{res:joint}, the history forgetting term for the $k$-th \ac{ar} step is defined by the conditional mutual information between its output and all previous frames, conditioned on the input and reference frames at that step. This concept is intuitive: the output can only draw upon information from the past through its input. This phenomenon leads to a specific type of inconsistency in the generated videos. For instance, as illustrated in the first row of Figure~\ref{fig:ar}, when the camera pans up to the sky and then returns downward, the \ac{arvdm} seems to \emph{forget} earlier frames, resulting in a completely different scene. It is important to note that the concept of \emph{history forgetting} is closely linked to the Information Bottleneck principle~\citep{tishby2000information}.

Despite the apparent visual evidence of history forgetting demonstrated in Figure~\ref{fig:ar}, a quantitative evaluation of this phenomenon remains challenging. Intuitively, commonly used video distribution metrics from previous studies, such as FID and FVD~\citep{harvey2022flexible}, are not well-suited for this purpose. These metrics assess distributional discrepancies (e.g., through the KL-divergence) and aggregate all error contributions noted on the right-hand side of Eqn.~\eqref{res:joint}, which obscures the specific impact of history forgetting. Additionally, motion-related metrics, such as smoothness and temporal flickering, employed in auto-regressive video generation~\citep{henschel2024streamingt2v}, concentrate on local frame characteristics and fail to capture the correlation between potentially distant generated frames and past frames—the key aspect of history forgetting. In contrast, our experimental section directly measures the dependence between $\texttt{Output}_{k}$ and $\texttt{Past}_{k}$ as defined in the context of history forgetting, employing ``needle-in-a-haystack'' problems. This approach effectively isolates and reflects the impact of forgetting.

\noindent \textbf{Effect of temporal degradation.}  Eqn.~\eqref{res:recursive} defines the error associated with the $(K+1)$-st short video clip of length $\Delta$ generated by an \ac{arvdm}. This error accumulates contributions from the initialization stage and all $K$ preceding \ac{ar} steps, including noise initialization, score estimation, and discretization errors. Unlike Eqn.~\eqref{res:joint}, which assesses the consistency of the entire video, Eqn.~\eqref{res:recursive} focuses specifically on the quality of a local segment, omitting a history forgetting term that accounts for inconsistencies across multiple \ac{ar} steps. This is exemplified in the last three cases of Figure~\ref{fig:ar}, where later generated frames display unrealistic configurations. To quantitatively assess temporal degradation, our findings indicate that standard video distribution metrics applied to short clips, specifically $Y_{K\Delta+1:(K+1)\Delta}^{0}$, effectively capture this effect.

We have shown that the intuitive concept of \emph{history forgetting} is evident in the \emph{upper bound} of the error. However, it remains unclear whether this phenomenon arises from our analysis or is \emph{intrinsic} to \ac{arvdm}s—that is, whether it is also reflected in a \emph{lower bound}. To address this question, we establish a lower bound under minimal assumptions, demonstrating that history forgetting is not just an artifact of our analysis, but rather an inherent limitation of the model. We will now demonstrate the inevitability of history forgetting, even in the presence of unlimited computational resources.


This information-theoretic limitation arises because the score function $s_{\theta}$
only conditions on the current input, without full access to past frames. We reduce
the problem to the setting in Figure~\ref{fig:lb} in Appendix~\ref{app:lb_fig},
where the goal is to learn the joint distribution $P_{X, Y, Z}$. Here, the random variables $X$, $Y$
and $Z$ represent $\texttt{Past}$, $\texttt{Input}$, and $\texttt{Output}$, respectively.
However, \ac{arvdm}s use only $\texttt{Input}$ to predict $\texttt{Output}$—i.e.,
partial observations. We assume access only to samples from the marginals:
$\{(X_{i}, Y_{i})\}_{i=1}^{N}\sim P_{X, Y}$ and
$\{(Y_{i}, Z_{i})\}_{i=N+1}^{2N}\sim P_{Y, Z}$, reflecting that $s_{\theta}$ is trained
on segments of long videos. The goal is to estimate $P$ using the dataset
$\calD_{N}(P) = \{(X_{i}, Y_{i})\}_{i=1}^{N}\cup \{(Y_{i}, Z_{i})\}_{i=N+1}^{2N}$.
The resulting estimate $\hat{P}$ captures both training and inference. For
simplicity, we assume $X, Y, Z \in \{0,1\}$.

\begin{theorem}
    \label{thm:low_bd} For $s\in[0,1]$ and any random variables $X,Y,Z$ (representing
    $\texttt{Past}$, $\texttt{Input}$ and $\texttt{Output}$, respectively), we
    define the conditional mutual information-constrained distribution set as
    $\calS(s)=\{ P=P_{X,Y,Z}\in\calP(\Omega^{3}) \mid I(X;Z|Y)\leq s\}$. Then for
    any $N\in\bbN$, $P\in\calS(s)$, and any estimate $\hatP$ of the distribution
    $P$ derived from $\calD_{N}(P)$, we have that
    \begin{align*}
        \inf_{\hatP\in\sigma(\calD_{N}(P))}\sup_{P\in\calS(s)}\Pr\big(\kl(P\,\|\,\hatP)\geq s^{2}/2 \big)\geq 0.5.
    \end{align*}
\end{theorem}

The proof is provided in Appendix~\ref{app:low_proof}. This theorem 
implies that
$\Pr(\kl(P\,\|\,\hatP)\geq 0.5 I^{2}(X;Z|Y))\geq \Pr(\kl(P\,\|\,\hatP)\geq 0.5 s^{2}
) \geq 0.5$. Thus, the conditional mutual information is an information-theoretic
barrier of \ac{arvdm}. We do not impose constraints on the availability of
computational resources. This implies history forgetting cannot be avoided even with
unlimited computational resources.

While the \emph{presence} of history forgetting is inevitable, its \emph{magnitude}
can be reduced. Intuitively, the more information the input carries about the
past, the smaller this term becomes. This intuition is formalized in the
following result.
\begin{proposition}
    \label{prop:bless} For any random variables $X,Y,Z$ and functions $f,g,h$ such
    that $g(x) = h(f(x))$ for all $x$, we have that
    $I(X;Z| Y, f(X))\leq I(X;Z| Y,g(X)).$
\end{proposition}

The proof is in Appendix~\ref{app:bless}. A special case of the result is that $I
(X;Z| Y, f(X))\leq I(X;Z| Y)$, where $h(f(x))=0$ is a trivial function. This result shows that adding more information from the past results in the history forgetting term  to be \emph{monotonically} non-increasing. The most direct construction of $f(\cdot)$
is to let the denoising networks take $X$ or a part of $X$ as the input. However,
there can be redundant information in all the past frames that cannot mitigate the
history forgetting.
\begin{fact}
    \label{fact:redun} For any three random variables $X,Y,Z$, if $f(X)$ and $Z$
    are conditionally independent given $Y$, then $I(X;Z| Y, f(X))= I(X;Z| Y)$.
\end{fact}
Thus, for efficient inference, it is also important to compress $X$, i.e., discarding
conditionally independent components. We will explore this method in the next
section. 
\section{Empirical Verifications}
\label{sec:improve}
The primary objectives of our experiments are twofold: (1) to validate the theoretical
conclusions established in Section~\ref{sec:analysis}, particularly regarding the
monotonic relationship between history length and history forgetting, as well as
the correlation between history forgetting and temporal degradation; and (2) to
provide new empirical insights for the characteristics of \ac{arvdm}s. While we compare
our approach against various baselines, these comparisons serve to contextualize
our findings and support our theoretical claims rather than to compete for state-of-the-art
performance on generation quality metrics.

\subsection{Experiments on History Forgetting}

\subsubsection{Experimental Setup}
We conduct experiments on two controlled environments: DMLab~\citep{DMLab_BeattieLTWWKLGV16} and Minecraft~\citep{MineRL_GussHTWCVS19}. For DMLab, we follow the TECO~\citep{yan2023temporally} training split and select evaluation cases from the corresponding validation split. For Minecraft, we render scenarios using MineRL while preserving the original frame rate. Our models are trained using the Adam optimizer $(\beta_1, \beta_2) = (0.9, 0.999)$ with no weight decay. Gradients are clipped to a maximum norm of 1.0, with additional noise clipping at a norm of 6.0. For Minecraft experiments, we adopt the VAE from \textit{stable-diffusion-3-medium} to enable latent diffusion. Across all experiments, we use fused SNR reweighting with a cumulative SNR decay of 0.96 and train with v-prediction as the target. Our denoiser is a 3D UNet with 
\begin{wrapfigure}{r}{0.45\textwidth}
\centering
\includegraphics[width=0.45\textwidth]{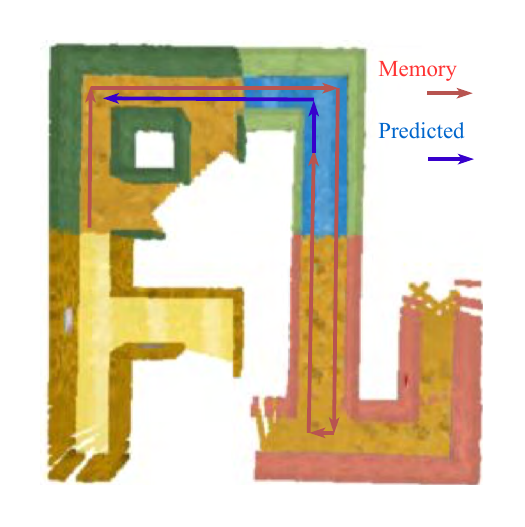}
\caption{An example of the history retrieval task in DMLab. The agent starts at the initial position and must generate frames showing the return to previously visited areas, testing the model's ability to retrieve historical context.}
\label{fig:dm_lab}
\vspace{-2em}
\end{wrapfigure}
ResNet blocks alongside spatial and temporal attention modules, comprising 4 blocks each in the downsampling and upsampling paths.

The direct calculation of history forgetting—defined as the conditional mutual
information between the output and the past frames—is often infeasible for high-dimensional
problems due to computational and sample complexity~\citep{belghazi2018mutual,paninski2003estimation}.
To overcome this, we empirically evaluate history forgetting by measuring the inconsistency
between the output and the past frames. Specifically, we design a visual ``needle-in-the-haystack''
task, where \ac{arvdm}s are tasked with predicting several future frames that
are determined by the past, serving as ground truth. History forgetting is
quantified by the difference between the predicted frames and the ground truth.
Figure~\ref{fig:dm_lab} illustrates an example of the history retrieval task in DMLab, where the model must navigate back to previously visited locations to generate frames consistent with the past context.

\textbf{Evaluation Metric.} To access ground truth, we use controlled
environments—DMLab~\citep{DMLab_BeattieLTWWKLGV16} and Minecraft~\citep{MineRL_GussHTWCVS19}—where
frames are determined by actions. In DMLab, which features simple scenes with floors,
walls, and wall decorations, we use the \ac{srr} metric, defined as the ratio of
trials in which floor and wall colors and wall decorations are correctly
retrieved. For the more complex and diverse Minecraft scenes, we adopt SSIM~\citep{DBLP:journals/tip/WangBSS04}.
It measures the similarity of two images via the expectation, variance, and
covariance of pixels. Retrieval is considered failed if \ac{srr} is $0$ or SSIM
is below $0.4$.
\begin{figure}[t]
    \centering
    \begin{subfigure}{\textwidth}
        \includegraphics[width=\linewidth]{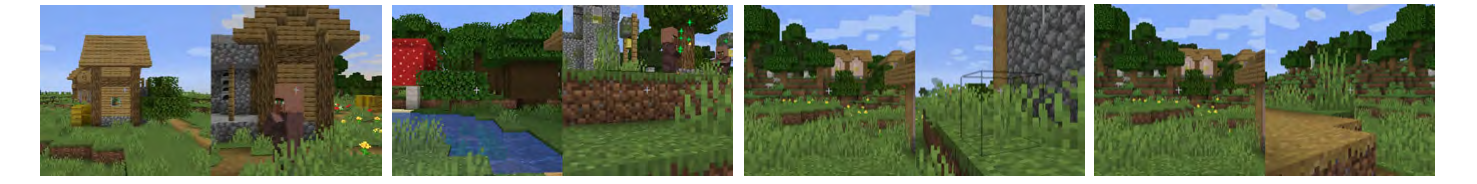}
        \caption{SSIM < 0.4: The visual similarity between frames is generally weak, making it difficult to discern a clear relationship between them.}
    \end{subfigure}
    \begin{subfigure}{\textwidth}
        \includegraphics[width=\linewidth]{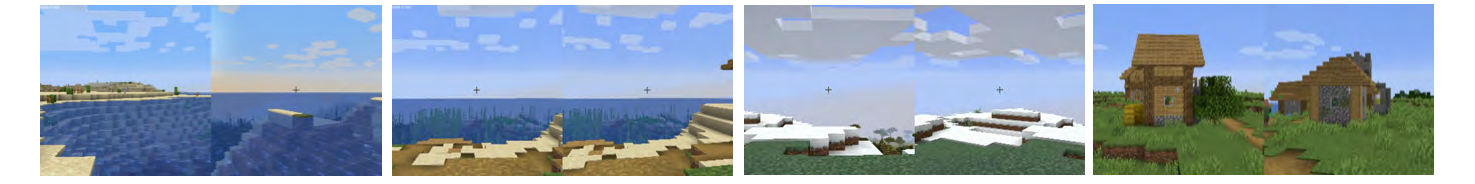}
        \caption{$0.4 \leq \text{SSIM} < 0.9$: The two frames typically share an overall resemblance in scene composition but exhibit noticeable differences in details, such as slight variations in camera angle or object placement.}
    \end{subfigure}
    \begin{subfigure}{\textwidth}
        \includegraphics[width=\linewidth]{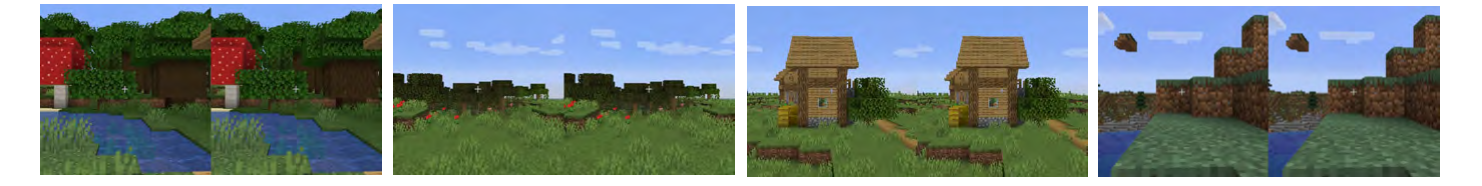}
        \caption{SSIM $\ge$ 0.9: The frames are nearly identical, with only minor pixel-level differences that are imperceptible to the human eye.}
    \end{subfigure}
    \caption{Visual interpretation of SSIM ranges in Minecraft. Example frame pairs demonstrate the three thresholds used for evaluating history retrieval performance.}
    \label{fig:minecraft_ssim}
\end{figure}

To provide intuition for these SSIM thresholds, Figure~\ref{fig:minecraft_ssim} illustrates example frame pairs across different SSIM ranges in Minecraft, sampled from approximately 1,000 video trajectories. As shown in the figure, SSIM < 0.4 indicates weak visual similarity where the relationship between frames is difficult to discern; $0.4 \leq \text{SSIM} < 0.9$ represents overall resemblance with noticeable detail differences (e.g., slight camera angle variations or object placement); and SSIM $\ge$ 0.9 indicates nearly identical frames with only imperceptible pixel-level differences.

We will employ these metrics to assess how injection of past information impacts history
forgetting, fixing the sliding window and stride to $w=\Delta=16$ across all experiments.

\begin{figure}[t]
    \centering
\includegraphics[trim={0.6cm, 0.5cm, 0.6cm, 0.1cm}, clip, width=1.0\textwidth]{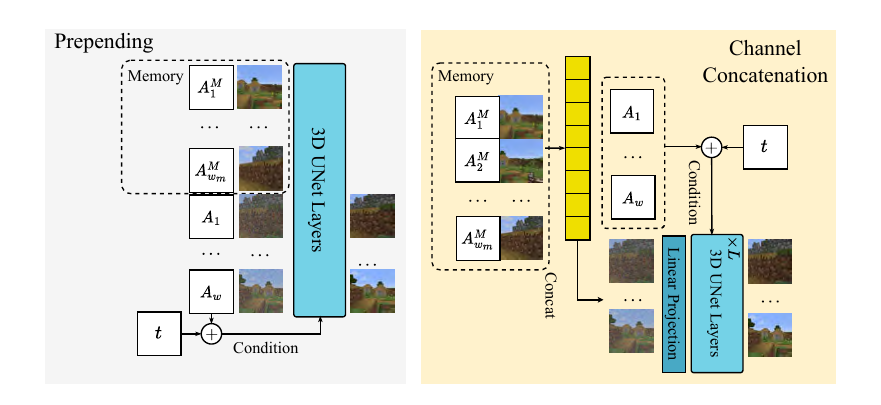}
    \caption{Network structures for adding information of previous frames into each \ac{ar} step. Here $w$ is the number of past frames provided for the denoising network. The superscript $M$ refers to the past frames and actions (memory).}
    \label{fig:network}
\end{figure}

\textbf{Network Architectures.} We explore two structures for integrating past frame information
into current \ac{ar} generation on a shared U-Net backbone, as illustrated in Figure~\ref{fig:network}:

\textit{Prepending (temporal concatenation)}: We directly concatenate the past $w$ frames before the current frames along the temporal axis; past actions are layer-normalized and injected as control signals. This feeds explicit, retrievable history into the backbone.

\textit{Channel concatenation}: We concatenate the past frames and their action embeddings along the channel dimension at the network input and use an initial convolutional fusion. Unlike \citet{che2024gamegen} (which extends image diffusion to video and relies on cross-modality attention), we explicitly fuse actions with frames via a lightweight Transformer encoder to form the input tokens, and inject history at every denoised frame since VDMs denoise multiple frames per step.

\begin{figure}[t]
\centering
\includegraphics[trim={0.5cm, 1cm, 0.5cm, 0.1cm}, clip,width=0.9\textwidth]{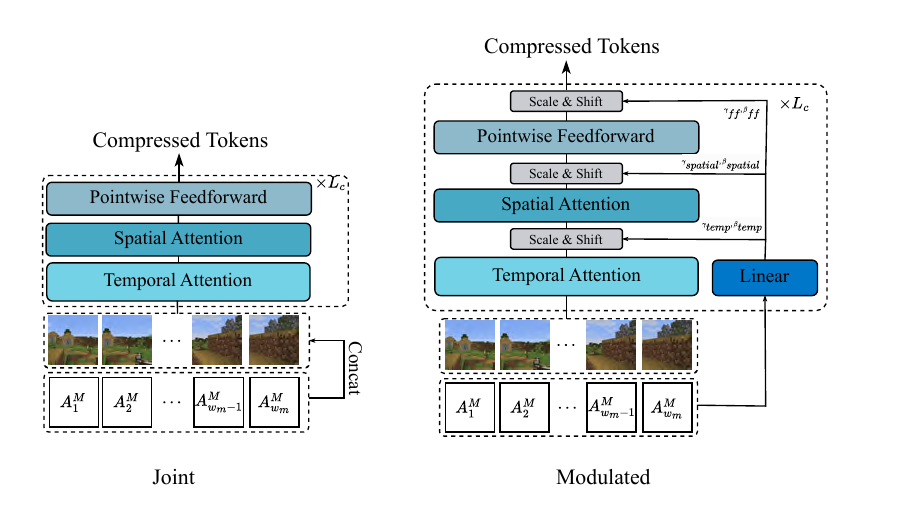}
\caption{Network structures for compressing past frames and actions. Left: joint compression processes concatenated frames and actions through feedforward and attention modules. Right: modulated compression uses actions to modulate frame representations. Both retain only the last several tokens as the final compressed memory.}
\label{fig:comp}
\end{figure}

\textbf{History Compression (Optional).} Orthogonally, to balance retrieval quality and efficiency, we optionally compress the ``past frames + actions'' into a compact memory with budget $B\in\{8,16,32,48\}$ using a small Transformer (Figure~\ref{fig:comp}). We adopt two methods to compress the past frames and actions. The joint method, which is shown in the left part of Figure~\ref{fig:comp}, processes the concatenation of the frame and actions jointly. The network consists of the feedforward module and spatial and temporal attention modules. The output is also the concatenation of compressed frames and actions. The modulated method, which is shown in the right part of Figure~\ref{fig:comp}, utilizes the actions to modulate the frames. In both methods, we only retain the last several frames and tokens as the final compressed memory. In the experiment, the prepending and channel concatenation structures respectively adopts the joint compression and modulated compression. The reason is that the prepending structure also requires compressed actions to transform the frames to different feature spaces, and the channel concatenation needs to merge the frame and action information.

\begin{table}[t]
    \centering
    \caption{History-Aware Retrieval Metrics: SRR $\uparrow$ for DMLab, and SSIM
    $\uparrow$ for Minecraft. Given a reference timestamp, the notation ``$i\text{-}
    j$'' denotes the task of retrieving the scenes that appeared in the past
    $i\text{-}j$ frames relative to the timestamp. ``History'' indicates the length
    of the provided historical context, while ``Budget'' refers to the network
    compression constraint.}
    \resizebox{1.0\textwidth}{!}{
    \begin{tabular}{c c c | c c c c | c c c c}
        \toprule \multirow{2}{*}{\textbf{Method}} & \multirow{2}{*}{\textbf{History}} & \multirow{2}{*}{\textbf{Budget}} & \multicolumn{4}{c|}{DMLab (SRR $\uparrow$) for $i-j$} & \multicolumn{4}{c}{Minecraft (SSIM $\uparrow$) for $i-j$} \\
                                                  &                                   &                                  & 0-16                                                  & 16-48                                                    & 48-112 & Avg  & 0-16 & 16-48 & 48-112 & Avg  \\
        \midrule \multirow{7}{*}{Prepending}      & 16                                & -                                & 0.52                                                  & 0.03                                                     & 0.00   & 0.18 & 0.75 & 0.39  & 0.37   & 0.50 \\
                                                  & 48                                & -                                & 0.44                                                  & 0.32                                                     & 0.00   & 0.25 & 0.74 & 0.62  & 0.35   & 0.57 \\
                                                  & 112                               & -                                & 0.48                                                  & 0.28                                                     & 0.23   & 0.33 & 0.72 & 0.60  & 0.55   & 0.62 \\
                                                  & 48                                & 8                                & 0.28                                                  & 0.22                                                     & 0.00   & 0.17 & 0.63 & 0.53  & 0.38   & 0.51 \\
                                                  & 48                                & 16                               & 0.36                                                  & 0.25                                                     & 0.00   & 0.20 & 0.67 & 0.55  & 0.37   & 0.53 \\
                                                  & 48                                & 32                               & 0.44                                                  & 0.28                                                     & 0.00   & 0.24 & 0.70 & 0.57  & 0.38   & 0.55 \\
                                                  & 48                                & 48                               & 0.40                                                  & 0.34                                                     & 0.00   & 0.25 & 0.72 & 0.61  & 0.35   & 0.56 \\
        \midrule \multirow{7}{*}{Channel Concat}  & 16                                & -                                & 0.24                                                  & 0.03                                                     & 0.00   & 0.09 & 0.61 & 0.38  & 0.38   & 0.46 \\
                                                  & 48                                & -                                & 0.16                                                  & 0.19                                                     & 0.00   & 0.12 & 0.58 & 0.57  & 0.37   & 0.51 \\
                                                  & 112                               & -                                & 0.12                                                  & 0.13                                                     & 0.08   & 0.11 & 0.59 & 0.56  & 0.51   & 0.55 \\
                                                  & 48                                & 8                                & 0.16                                                  & 0.16                                                     & 0.00   & 0.11 & 0.54 & 0.53  & 0.36   & 0.48 \\
                                                  & 48                                & 16                               & 0.20                                                  & 0.19                                                     & 0.00   & 0.13 & 0.56 & 0.54  & 0.35   & 0.48 \\
                                                  & 48                                & 32                               & 0.24                                                  & 0.22                                                     & 0.00   & 0.15 & 0.57 & 0.56  & 0.36   & 0.50 \\
                                                  & 48                                & 48                               & 0.28                                                  & 0.22                                                     & 0.00   & 0.17 & 0.57 & 0.57  & 0.37   & 0.50 \\
        \bottomrule
    \end{tabular}} \label{tab:dmlab_minecraft_retrieval}
    \vspace{-0.8em}
\end{table}

\subsubsection{Experiment Results.} 

\paragraph{Verification of Theorem~\ref{thm:main}, Proposition~\ref{prop:bless}, and Fact~\ref{fact:redun}.} As discussed in Section~\ref{sec:analysis}, we quantify history forgetting using the conditional mutual information between the generated frames and all past frames, conditioned on the inputs. Proposition~\ref{prop:bless} and Fact~\ref{fact:redun} provide two key properties. Proposition~\ref{prop:bless} states that conditioning on more informative variables cannot worsen performance, whereas Fact~\ref{fact:redun} shows that conditioning on redundant variables—i.e., variables independent of the future frames given the input—cannot improve it. We empirically validate these intuitions. To verify Proposition~\ref{prop:bless}, we condition the frame-generation process on increasingly large sets of past frames, where each larger set strictly contains the smaller ones. If the monotonicity property holds, history-forgetting error should decrease as more frames are given. To verify Fact~\ref{fact:redun}, we apply different levels of compression to the conditioned frames. If redundant information does not help, then light compression (which mainly removes redundancy) should yield performance comparable to no compression, while heavy compression (which removes useful information) should significantly degrade it.

\begin{figure}
    \begin{center}
        \includegraphics[width=0.80\textwidth]{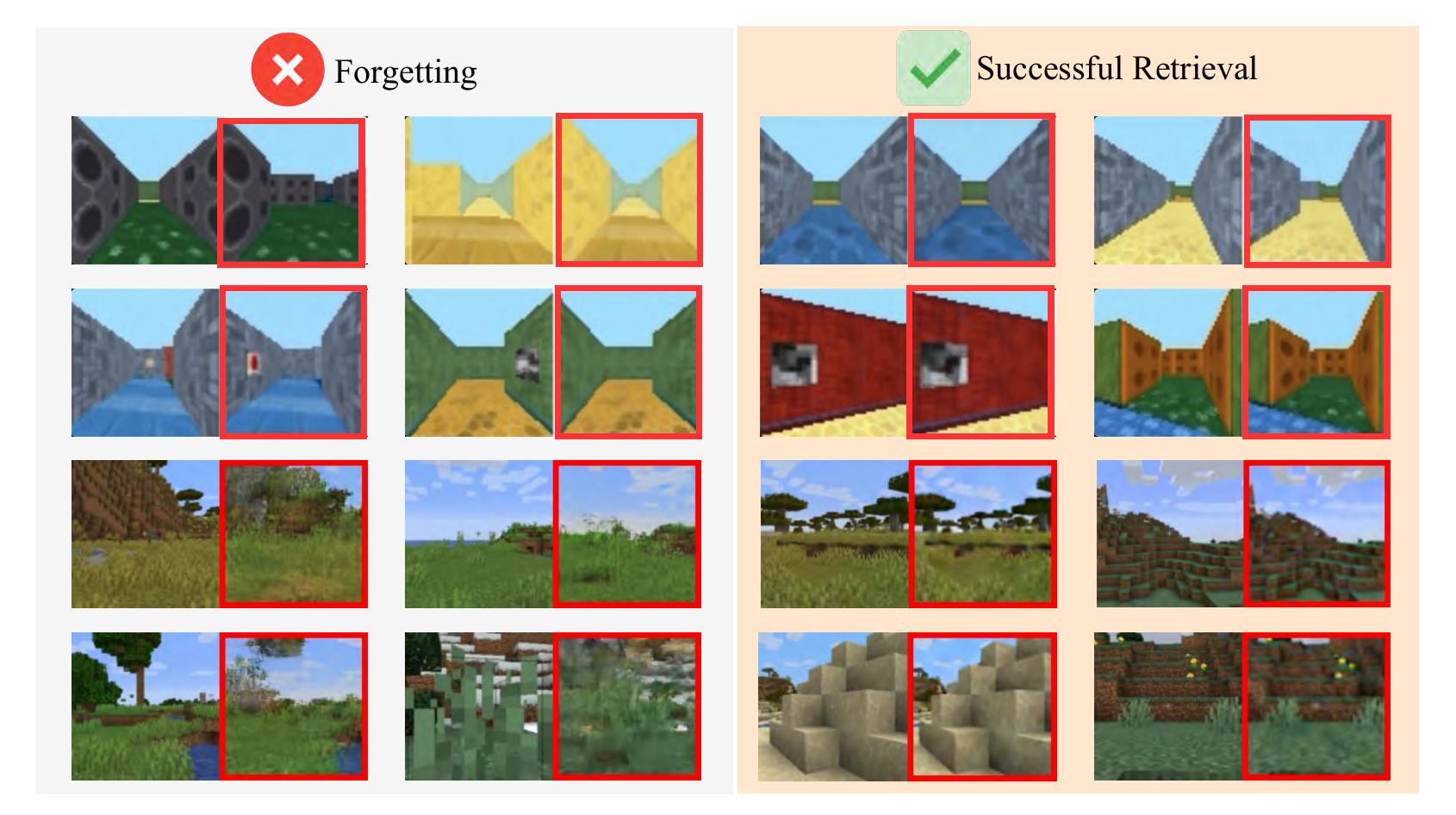}
    \end{center}
    \caption{Examples of successful history retrieval in DMLab and Minecraft.
    Rows show cases where the model returns to previously visited areas or reproduces earlier scene layouts, indicating low history-forgetting error.}
    \label{fig:demo_mini}
\end{figure}

The results of \ac{srr} for DMLab and SSIM for Minecraft in
Tables~\ref{tab:dmlab_minecraft_retrieval} effectively demonstrate the impact of
history budget for compression ($8, 16, 32, 48$), history length ($16, 48, 112$),
and network structures. Figure~\ref{fig:demo_mini} presents representative successful retrievals, and Appendix~\ref{app:demo} provides additional successful and failed cases. Across all architectures, history forgetting errors consistently decrease with increasing history length and history budget,
aligning with Proposition~\ref{prop:bless}. In this context, the functions $f$
and $g$ serve as the submodules responsible for integrating historical
information. Moreover, the compression results in Tables~\ref{tab:dmlab_minecraft_retrieval}
reveal significant redundancy in the past frames and actions. Notably, when the
history length is compressed from 48 to 8, the performance does not degrade proportionally,
verifying Fact~\ref{fact:redun}. This indicates that conditioning the model on redundant
information offers limited benefits.

\begin{table}[t]
    \centering
    \caption{Common Evaluation Metrics on DMLab and Minecraft across both structures. We report FVD and
    FID for both environments, while motion metrics (TF for Temporal Flickering and
    MS for Motion Smoothness) are reported only for Minecraft due to DMLab's
    insufficient frame rate for reliable motion evaluation. History length = w (16/48/112); Budget = number of preserved history latents after compression (8/16/32/48).}
    \resizebox{1.0\textwidth}{!}{
    \begin{tabular}{c c c | c c | c c c c}
        \toprule \multirow{2}{*}{ \textbf{Method} } & \multirow{2}{*}{ \textbf{History} } & \multirow{2}{*}{\hspace{5pt} \textbf{Budget} \hspace{5pt}} & \multicolumn{2}{c|}{\hspace{5pt} \textbf{DMLab} \hspace{5pt}} & \multicolumn{4}{c}{\hspace{5pt} \textbf{Minecraft} \hspace{5pt}} \\
                                                  &                                   &                                  & \hspace{5pt} \textbf{\ac{fvd}} \hspace{5pt}                   & \hspace{5pt} \textbf{\ac{fid}} \hspace{5pt}                     & \hspace{5pt} \textbf{\ac{fvd}} \hspace{5pt} & \hspace{5pt} \textbf{\ac{fid}} \hspace{5pt} & \hspace{5pt} \textbf{TF} \hspace{5pt} & \hspace{5pt} \textbf{MS} \hspace{5pt} \\
        \midrule \multirow{7}{*}{Prepending}      & 16                                & -                                & 302.59                              & 80.27                                 & 224.30            & 75.74             & 0.750       & 0.853       \\
                                                  & 48                                & -                                & 308.92                              & 78.44                                 & 222.78            & 73.96             & 0.754       & 0.853       \\
                                                  & 112                               & -                                & 308.81                              & 80.45                                 & 226.49            & 75.78             & 0.751       & 0.867       \\
                                                  & 48                                & 8                                & 310.59                              & 79.50                                 & 222.30            & 76.36             & 0.754       & 0.873       \\
                                                  & 48                                & 16                               & 307.20                              & 78.35                                 & 226.08            & 74.43             & 0.738       & 0.863       \\
                                                  & 48                                & 32                               & 311.94                              & 73.65                                 & 223.45            & 75.35             & 0.760       & 0.863       \\
                                                  & 48                                & 48                               & 308.73                              & 74.37                                 & 223.13            & 78.50             & 0.737       & 0.845       \\
        \midrule \multirow{7}{*}{Channel Concat}  & 16                                & -                                & 294.01                              & 71.07                                 & 209.12            & 71.49             & 0.757       & 0.861       \\
                                                  & 48                                & -                                & 300.56                              & 73.40                                 & 212.92            & 68.43             & 0.769       & 0.858       \\
                                                  & 112                               & -                                & 298.85                              & 74.00                                 & 206.17            & 70.51             & 0.762       & 0.862       \\
                                                  & 48                                & 8                                & 304.95                              & 72.84                                 & 208.16            & 68.44             & 0.761       & 0.855       \\
                                                  & 48                                & 16                               & 302.86                              & 74.78                                 & 207.02            & 68.58             & 0.773       & 0.859       \\
                                                  & 48                                & 32                               & 296.06                              & 75.40                                 & 214.01            & 66.04             & 0.770       & 0.860       \\
                                                  & 48                                & 48                               & 298.56                              & 74.02                                 & 209.55            & 71.40             & 0.771       & 0.847       \\
        \bottomrule
    \end{tabular}
    } \label{tab:common_metrics_dmlab_mc}
\end{table}

\paragraph{Proper evaluations of history forgetting.} The results in Tables~\ref{tab:dmlab_minecraft_retrieval} show that our ``needle-in-the-haystack'' design accurately captures history forgetting. As discussed in Section~\ref{sec:analysis}, our theory further implies that standard video-distribution metrics (e.g., \ac{fvd} and \ac{fid})~\citep{henschel2024streamingt2v,harvey2022flexible} and motion-related metrics (e.g., temporal flickering and motion smoothness)~\citep{huang2024vbench} are ineffective for measuring this phenomenon. Table~\ref{tab:common_metrics_dmlab_mc} presents results for both structures
across DMLab and Minecraft. Due to DMLab's low FPS, we assess temporal
flickering and motion smoothness only for Minecraft. These metrics show minor fluctuations
across both structures and settings without any clear link to history forgetting, consistent with our theoretical analysis. Notably, despite the sizable retrieval gap between prepending and channel concatenation, their FVD and FID scores remain comparable, further confirming these metrics' limitations in capturing history forgetting. \looseness=-1

A natural follow-up question is whether existing methods that improve distribution and motion
scores can also reduce history forgetting. To investigate this, we conducted a comprehensive comparison between our soft compression approach and various baselines, including frame sampling strategies, training-free autoregressive methods, and reimplemented state-of-the-art techniques. For fair comparison, we incorporate the action-conditioned benchmark from GameNGen~\citep{valevski2024diffusion} to enable autoregressive long video generation. We also reimplement representative baselines such as StreamingT2V~\citep{henschel2024streamingt2v} and adopt various frame sampling strategies inspired by prior work~\citep{harvey2022flexible}. All re-implemented baselines, which lack mechanisms to leverage long-horizon history, are restricted to using only the last 16 frames as input.

Table~\ref{tab:compression_baseline} presents the quantitative comparison on DMLab and Minecraft under a fixed network compression budget of 16 and history length of 48. The results reveal several important findings. First, our soft compression (prepending memory with memory compression) consistently outperforms all baselines across both short-term (0-16) and longer temporal horizons (16-32), achieving the best overall performance on both environments. Second, training-free methods (FIFO and Outpainting) show limited effectiveness in reducing history forgetting, with particularly poor performance on longer horizons. Third, while reimplemented baselines like StreamingT2V and GameNGen demonstrate reasonable short-term performance, they struggle with mid-range and long-range history retrieval.

\begin{table}[t]
    \centering
    \caption{Comparison of different history compression methods on DMLab and Minecraft. Our soft compression approach (bottom row) consistently outperforms frame sampling strategies, training-free methods, and re-implemented baselines, especially over longer temporal horizons (16-32 frames).}
    \resizebox{0.9\textwidth}{!}{
    \begin{tabular}{l|c c c|c c c}
        \toprule \multirow{2}{*}{\textbf{Method}}                           & \multicolumn{3}{c|}{\textbf{DMLab (SRR $\uparrow$)}} & \multicolumn{3}{c}{\textbf{Minecraft (SSIM $\uparrow$)}} \\
                                                                            & \textbf{0-16}                                        & \textbf{16-32}                                          & \textbf{Overall} & \textbf{0-16} & \textbf{16-32} & \textbf{Overall} \\
        \midrule \multicolumn{7}{l}{\textbf{Frame Sampling}}                 \\
        \midrule \makecell[l]{Prepending Memory\\ + Last 16 Frames}         & 0.520                                                & 0.000                                                   & 0.260            & 0.610         & 0.380          & 0.495            \\
        \makecell[l]{Prepending Memory\\+ Last 32 Frames with 1 Frame Skip} & 0.260                                                & 0.154                                                   & 0.207            & 0.510         & 0.434          & 0.472            \\
        \midrule \multicolumn{7}{l}{\textbf{Training-free AR Methods}}       \\
        \midrule FIFO                                                       & 0.123                                                & 0.030                                                   & 0.077            & 0.453         & 0.356          & 0.405            \\
        Outpainting                                                         & 0.100                                                & 0.021                                                   & 0.061            & 0.430         & 0.360          & 0.395            \\
        \midrule \multicolumn{7}{l}{\textbf{Re-implemented Baselines}}       \\
        \midrule GameNGen                                                   & 0.214                                                & 0.000                                                   & 0.107            & 0.471         & 0.321          & 0.396            \\
        StreamingT2V                                                        & 0.263                                                & 0.040                                                   & 0.152            & 0.480         & 0.312          & 0.396            \\
        \midrule \multicolumn{7}{l}{\textbf{Soft Compression}}               \\
        \midrule \makecell[l]{Prepending Memory\\+ Memory Compression}      & 0.360                                                & 0.250                                                   & 0.305            & 0.561         & 0.543          & 0.552            \\
        \bottomrule
    \end{tabular}
    } \label{tab:compression_baseline}
\end{table}

Specifically, StreamingT2V~\citep{henschel2024streamingt2v}, which includes several first frames (long-term information) and near frames (short-term information), is designed to improve motion degree and warping error. However, as shown in Table~\ref{tab:compression_baseline}, it cannot effectively retrieve mid-range history, achieving only 0.040 SRR on DMLab's 16-32 frame range. Similarly, GameNGen~\citep{valevski2024diffusion} emphasizes short-term inter-frame consistency but sacrifices long-term consistency, completely failing on longer horizons (0.000 SRR for 16-32 frames in DMLab). Frame sampling strategies~\citep{harvey2022flexible} merely trade off history forgetting errors across different horizons rather than fundamentally addressing the problem. These results suggest that existing approaches are not designed to effectively tackle history forgetting, and discovering optimal strategies for compressing historical information remains an open challenge.

\subsection{Experiments on Temporal Degradation}

\begin{figure}
    \begin{center}
        \includegraphics[width=0.7\textwidth]{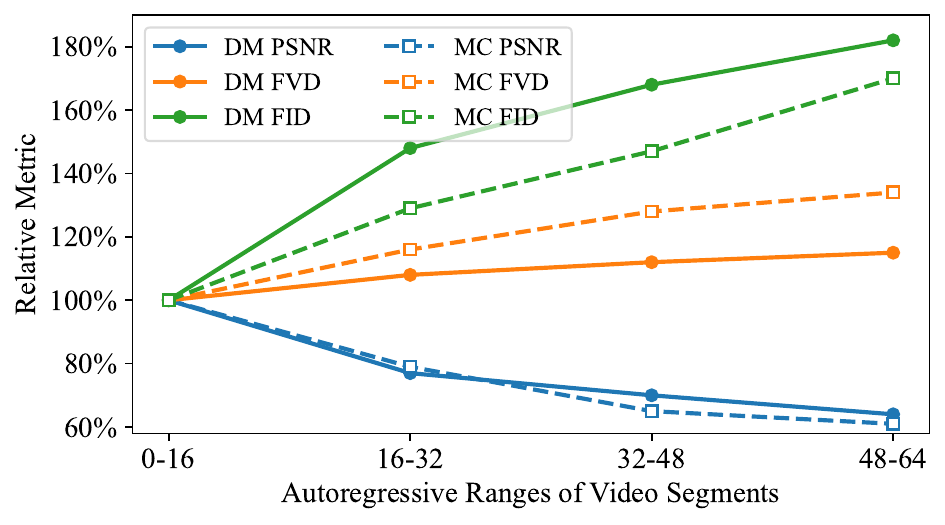}
        \caption{Temporal degradation results for DMLab and Minecraft.
        All values are normalized with respect to frames 0–16. }
        \label{fig:temp_deg_norm}
    \end{center}
\end{figure}

\paragraph{Proper evaluation of temporal degradation.} Our theoretical results in Section~\ref{sec:analysis} imply that any video-distribution metrics are suitable to evaluate temporal degradation. In the following, we will verify this by evaluating \ac{fvd}, \ac{fid}, and PSNR. Similar to the
concept of history forgetting, this evaluation is conducted within DMLab and
Minecraft environments. Figure \ref{fig:temp_deg_norm} indicates normalized different metrics for
temporal degradation. All the metrics of the first $16$ frames are normalized to
$100\%$, which corresponds to the initial portion of the generated video and is
expected to exhibit the highest visual quality. The results show that
\ac{fvd}, \ac{fid}, and PSNR show clear trends to degrade along the temporal
axis. It qualitatively corroborates our theoretical findings. 
\begin{figure}
    \centering
    \includegraphics[width=0.7\textwidth]{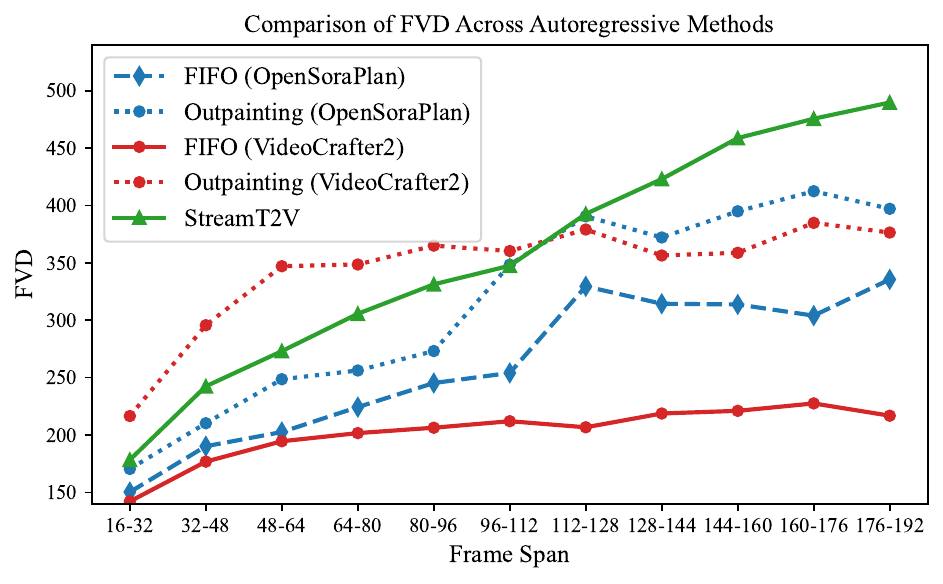}
    \caption{FVDs of \emph{short} clips generated by different models and methods.
    The FVD of short clips generated by all the methods and models grows with
    increasing \ac{ar} steps, which we coin temporal degradation.}
    \label{fig:fvd}
\end{figure}
Note that assessing temporal degradation does not necessitate
closed-ended environments. To demonstrate the generality of temporal degradation beyond these controlled settings, we also evaluate it on general text-to-video generation tasks using pretrained models. Figure~\ref{fig:fvd} shows FVD measurements of short clips generated by different pretrained models across multiple autoregressive steps, confirming that temporal degradation is a widespread phenomenon affecting various video generation architectures.

\begin{figure}
 \begin{subfigure}{0.45\textwidth}
            \centering
            \includegraphics[width=0.99\textwidth]{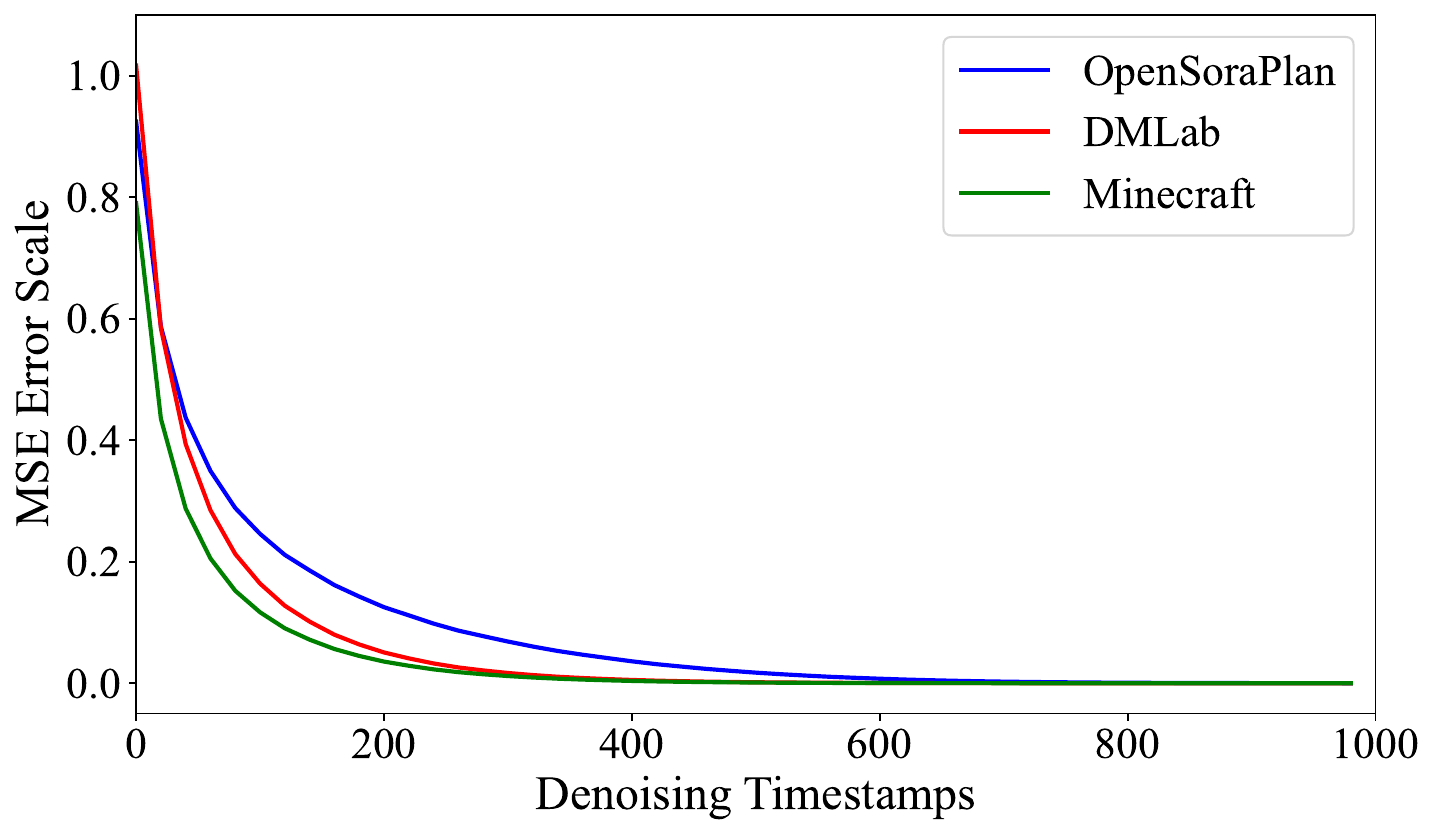}
            \caption{Pretraining errors of different time steps for OpenSoraPlan, DMLab, and Minecraft.}
             \label{fig:pretrained_loss_curve}
\end{subfigure}
\hfill
 \begin{subfigure}{0.52\textwidth}
            \centering
            \includegraphics[width=0.99\textwidth]{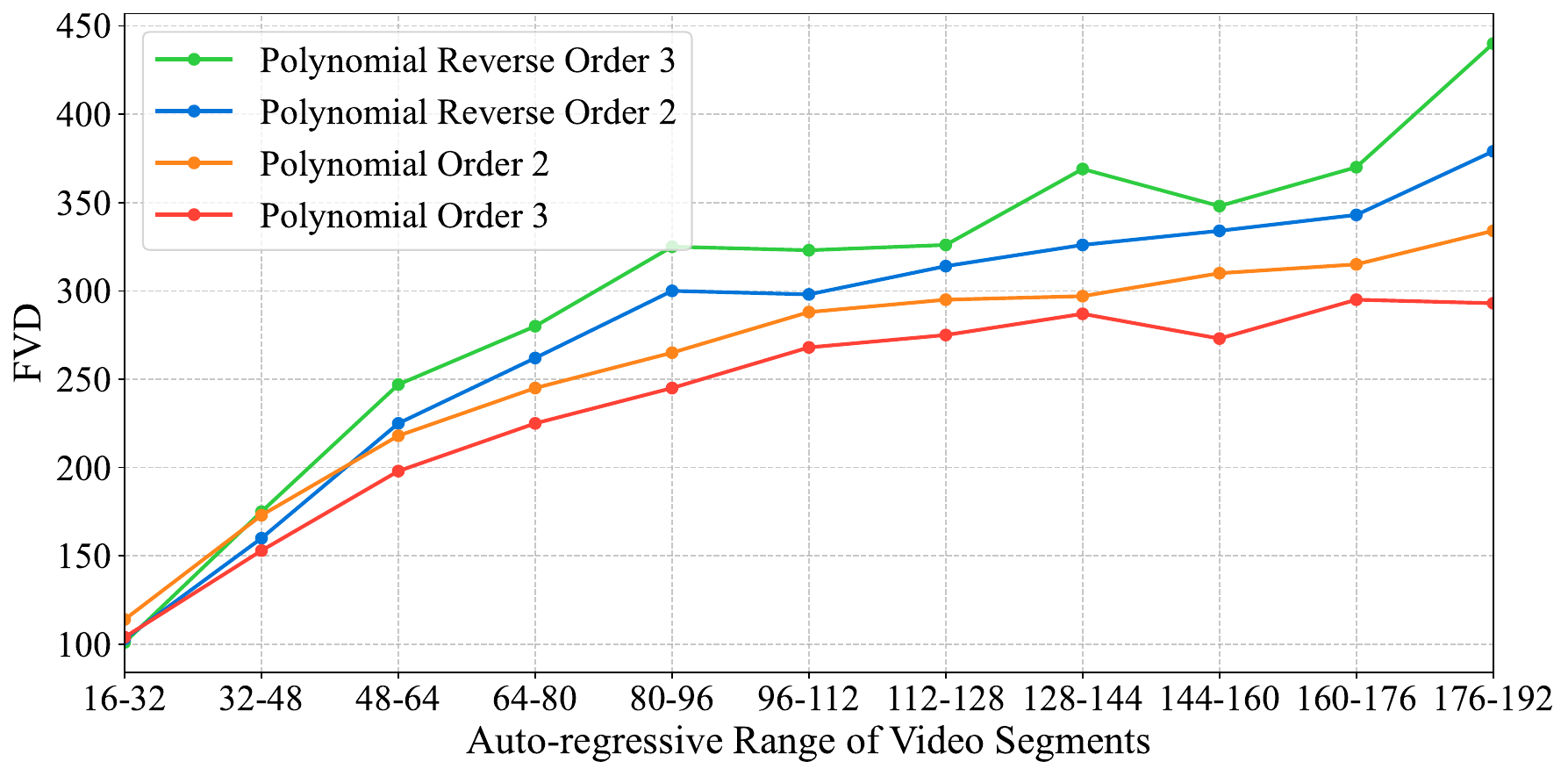}
            \caption{Comparison of FVD across different timestep sets for OpenSoraPlan. }
              \label{fig:td_diff_timesteps}
\end{subfigure}
\caption{Impact of timestep schedules on training error and temporal degradation. Left: pretraining loss evolution for different timestep sets across OpenSoraPlan, DMLab, and Minecraft. Right: FVD comparison of the same schedules on OpenSoraPlan, highlighting how denser early steps curb degradation.}
              
\end{figure}

\paragraph{Quantitative verification of Eqn.~\eqref{res:recursive}.} We highlight Eqn.~\eqref{res:recursive} can also quantitatively predict the rate of temporal degradation.
We measure the temporal degradation of four DDPM time-step sets concerning \ac{fvd}:
(a,1) $\unif([0,999^{1/2}],N)^{2}$, (a,2) $999 -\unif([0,999^{1/2}],N)^{2}$, (b,1)
$\unif([0,999^{1/3}],N)^{3}$, (b,2) $999 - \unif([0,999^{1/3}],N)^{3}$. Here, $\unif([0,999^{1/2}], N)^{2}$ denotes the following sampling procedure: we first draw $N$ points uniformly from the interval $[0, 999^{1/2}]$, and then define the selected steps as the squared values of these $N$ points. The notation $999 - \unif([0,999^{1/2}], N)^{2}$ means that the selected steps are obtained by subtracting each element of $\unif([0,999^{1/2}], N)^{2}$ from $999$. Figure~\ref{fig:td_diff_timesteps}
shows that ($\ast$,1) is better than ($\ast$,2) for both (a) and (b), and that
the performance gap in (b,1)/(b,2) is larger. These all corroborate Eqn.~\eqref{res:recursive}.
$\text{NIE}$ and $\text{DE}$ are identical for ($\ast$,1) and ($\ast$,2).
By Assumption~\ref{assump:err}, the weights of steps in $\text{SEE}$ are proportional to the intervals
between steps. Since ($\ast$,1) has denser steps near 0 (i.e., smaller intervals
and weights), our bound predicts slower temporal degradation for ($\ast$,2). (b,1)/(b,2)
have a larger weight difference, implying a larger performance gap. Similar results
hold for Minecraft and pretrained T2V models (see Appendix~\ref{app:additional_experiments} for detailed results).

\noindent

\begin{figure}[t]
    \centering
        \centering
        \begin{subfigure}{0.49\textwidth}
            \centering
            \includegraphics[width=0.99\textwidth]{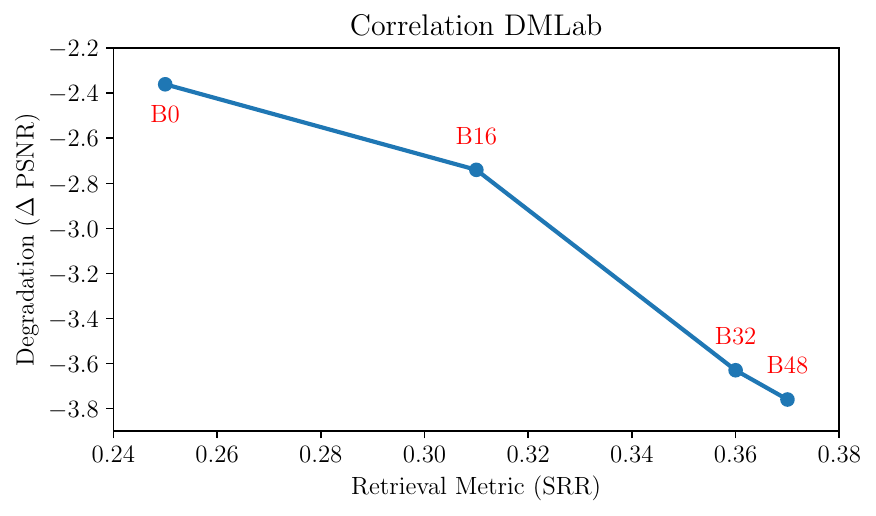}
            \caption{DMLab}
        \end{subfigure}
        \begin{subfigure}{0.49\textwidth}
            \centering
            \includegraphics[width=0.99\textwidth]{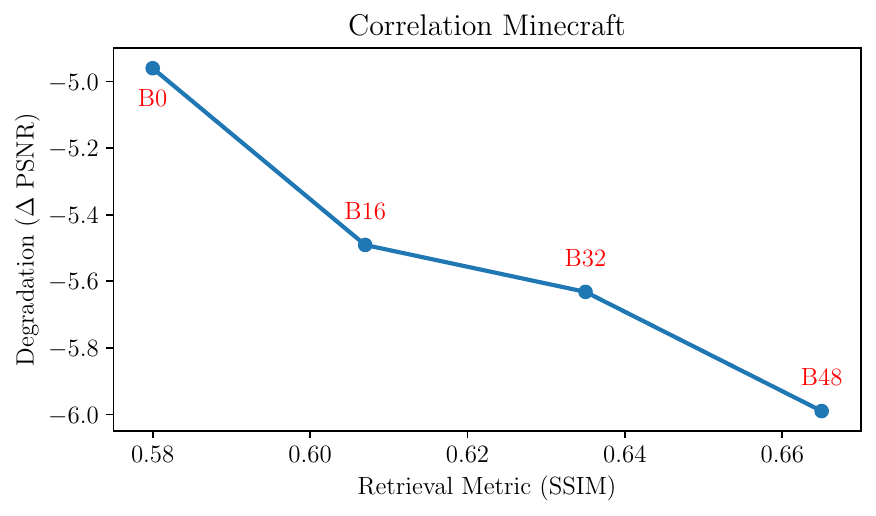}
            \caption{Minecraft}
        \end{subfigure}
        \caption{Correlation between history forgetting and temporal degradation in DMLab and Minecraft. Here $B_{i}$ denotes the network compression budget. Both environments exhibit a consistent negative correlation: higher retrieval accuracy (reduced forgetting) leads to faster quality degradation.}
        \label{fig:recall_vs_psnr}
\end{figure}

\textbf{Correlation between history forgetting and temporal degradation.}
Intuitively, history forgetting—the failure to retrieve relevant information from
prior context—and temporal degradation—the accumulation of noise over time—are negatively
correlated. A lower forgetting error indicates a stronger ability to retrieve
historical information, but this may also introduce more historical noise into current
outputs, resulting in more severe temporal degradation. Figure~\ref{fig:recall_vs_psnr}
illustrates this trade-off by plotting temporal degradation (measured as maximum
PSNR decay, see Table~\ref{tab:error-accumulation}) against the average \ac{srr}
for compressed prepending. As \ac{srr} increases (indicating reduced
history forgetting), PSNR decay also increases (indicating faster quality loss).
This negative correlation is consistently observed in both DMLab and Minecraft environments, as shown in Figure~\ref{fig:recall_vs_psnr}.
We leave the theoretical investigation of this phenomenon to future work.

\begin{table}[t]
    \centering
    \caption{Temporal degradation quantified by PSNR across different frame segments in DMLab and Minecraft. PSNR values consistently decrease over time, indicating progressive quality loss. Higher compression budgets generally result in faster degradation. Error bars are computed from 5 independent runs with different random seeds.}
    \resizebox{1.0\textwidth}{!}{
    \begin{tabular}{c | c c c c | c c c c}
        \toprule \multirow{2}{*}{\textbf{Budget}} & \multicolumn{4}{c|}{\textbf{DMLab}} & \multicolumn{4}{c}{\textbf{Minecraft}} \\
                                                  & \textbf{0--16}                      & \textbf{16--32}                       & \textbf{32--48} & \textbf{48--64} & \textbf{0--16}   & \textbf{16--32}  & \textbf{32--48}  & \textbf{48--64}  \\
        \midrule 8                                & 10.24 $\pm$ 0.11                    & 9.47 $\pm$ 0.13                       & 8.83 $\pm$ 0.14 & 7.88 $\pm$ 0.19 & 15.25 $\pm$ 0.15 & 13.43 $\pm$ 0.17 & 11.39 $\pm$ 0.20 & 10.29 $\pm$ 0.23 \\
        16                                        & 10.14 $\pm$ 0.10                    & 8.81 $\pm$ 0.14                       & 8.24 $\pm$ 0.17 & 7.39 $\pm$ 0.21 & 15.47 $\pm$ 0.13 & 12.92 $\pm$ 0.19 & 11.10 $\pm$ 0.23 & 9.98 $\pm$ 0.24  \\
        32                                        & 10.46 $\pm$ 0.14                    & 8.19 $\pm$ 0.13                       & 7.82 $\pm$ 0.18 & 6.83 $\pm$ 0.20 & 15.36 $\pm$ 0.16 & 12.64 $\pm$ 0.20 & 11.01 $\pm$ 0.22 & 9.73 $\pm$ 0.26  \\
        48                                        & 10.52 $\pm$ 0.13                    & 8.00 $\pm$ 0.15                       & 7.58 $\pm$ 0.18 & 6.76 $\pm$ 0.23 & 15.46 $\pm$ 0.18 & 12.31 $\pm$ 0.19 & 10.57 $\pm$ 0.26 & 9.47 $\pm$ 0.27  \\
        \bottomrule
    \end{tabular}
    } \label{tab:error-accumulation}
\end{table}

\section{Conclusion}
Based on the proposed unified framework \ac{arvdm}, we demonstrated that history
forgetting corresponds to the conditional mutual information between the output
and past frames, and its emergence is inherently inevitable. The temporal
degradation, in turn, accumulates as the total error from all preceding \ac{ar} steps.
Moreover, we observed a negative correlation between history forgetting and temporal
degradation. From a theoretical perspective, a limitation lies in the gap between
the upper and lower bounds for history forgetting, which could be narrowed by
replacing Pinsker's inequality with tighter alternatives. From an experimental standpoint,
our results do not directly indicate the most effective structure for mitigating
both history forgetting and temporal degradation. Additionally, understanding why certain attention-based mechanisms struggle to integrate historical information falls outside the scope of our theoretical framework. We hypothesize that these difficulties stem from the same underlying issue that causes weak conditioning from auxiliary modalities (e.g., text or image)~\citep{esser2024scaling}. Verifying this hypothesis and exploring optimal architectures for history integration are left for future investigations.


\vskip 0.2in
\bibliography{sample}
\clearpage
\appendix
\section{Notation}\label{app:notation}

This appendix provides a comprehensive list of mathematical notation used throughout the paper. Symbols are organized by category for ease of reference.

\subsection{Basic Notation}
\vspace{-1em}
\begin{table}[H]
\centering
\small
\begin{tabular}{@{}p{0.2\textwidth}p{0.6\textwidth}p{0.15\textwidth}@{}}
\toprule
\textbf{Symbol} & \textbf{Description} & \textbf{Defined in} \\
\midrule
$X_{1:w}$ & Clean video frames sequence $(X_1, \ldots, X_w) \in \mathbb{R}^{w \times d}$ & §\ref{sec:prelim} \\
$X_{1:w}^0$ & Initial clean video (same as $X_{1:w}$) & §\ref{sec:prelim} \\
$X_{1:w}^t$ & Noisy video frames at diffusion time $t$ & §\ref{sec:prelim} \\
$\tilde{X}_{1:w}^t$ & Reverse-time process for video frames & §\ref{sec:prelim} \\
$Y_{1:w}$ & Noisy video frames sequence (generated by algorithm) & Algorithm~\ref{algo:framework} \\
$Y_{1:w}^t$ & Noisy video at noise level $t$ (generated by algorithm) & Algorithm~\ref{algo:framework} \\
$w \in \mathbb{N}$ & Number of frames / effective window size & §\ref{sec:prelim} \\
$d$ & Dimensionality of each frame (e.g., $512 \times 512$ for $512 \times 512$-pixel frames) & §\ref{sec:prelim} \\
$t$ & Diffusion time / noise level & §\ref{sec:prelim} \\
$T$ & Maximum noise level of the diffusion process & §\ref{sec:prelim} \\
$B^t$ & $(w \times d)$-dimensional Brownian motion with identity covariance & §\ref{sec:prelim} \\
$\tilde{B}^t$ & Reverse-time Brownian motion & §\ref{sec:prelim} \\
$f(\cdot, t)$ & Drift coefficient function: $f:\mathbb{R}^{w\times d}\times \mathbb{R}\rightarrow\mathbb{R}^{w\times d}$ & §\ref{sec:prelim} \\
$g(t)$ & Diffusion coefficient function: $g:\mathbb{R}\rightarrow\mathbb{R}$ & §\ref{sec:prelim} \\
$\beta(t)$ & Scalar noise schedule (used in DDPM and SMLD) & §\ref{sec:prelim} \\
$P_t$ & Distribution of $X_{1:w}^t$ at diffusion time $t$ & §\ref{sec:prelim} \\
$\nabla\log P_t$ & Score function (gradient of log-density) & §\ref{sec:prelim} \\
\bottomrule
\end{tabular}
\end{table}
\subsection{AR-VDM Framework}

\begin{table}[H]
\centering
\small
\begin{tabular}{@{}p{0.2\textwidth}p{0.6\textwidth}p{0.15\textwidth}@{}}
\toprule
\textbf{Symbol} & \textbf{Description} & \textbf{Defined in} \\
\midrule
$\Delta \in \mathbb{N}$ & Auto-regressive step size (number of new frames generated per step) & Algorithm~\ref{algo:framework} \\
$\calL^{\rmI}$ & Input noise levels $(t_1^{\rmI}, \ldots, t_w^{\rmI})$ & Algorithm~\ref{algo:framework} \\
$\calL^{\rmO}$ & Output noise levels $(t_1^{\rmO}, \ldots, t_w^{\rmO})$ & Algorithm~\ref{algo:framework} \\
$t_i^{\rmI}$ & Input noise level for frame $i$ & Algorithm~\ref{algo:framework} \\
$t_i^{\rmO}$ & Output noise level for frame $i$ & Algorithm~\ref{algo:framework} \\
$\delta_i$ & Noise level offset for frame $i$: $\delta_i = t_i^{\rmI} - t_1^{\rmI} = t_i^{\rmO} - t_1^{\rmO}$ & §5.2 \\
$\bar{t}(t)$ & Multi-frame noise level mapping: $\bar{t}(t) = (t+\delta_1, \ldots, t+\delta_w)$ & §5.2 \\
$\bar{T}$ & Vector of maximum noise levels: $\bar{T} = (T, \ldots, T) \in \mathbb{R}^w$ & §5.2 \\
$X_{1:w}^{\bar{t}}$ & Frames at varied noise levels: $X_{1:w}^{\bar{t}} = (X_1^{t+\delta_1}, \ldots, X_w^{t+\delta_w})$ & §5.2 \\
$\tilde{Y}_{1:w}^{\bar{t}}$ & Reverse-time noisy frames at varied noise levels (AR stage) & §5.2 \\
$P_X^{\bar{t}}$ & Distribution of frames at varied noise levels & §5.2 \\
$\calR$ & Reference frames index set & Algorithm~\ref{algo:framework} \\
$\calR_i$ & Reference frames index set when denoising frames $i+1$ to $i+w$: $\calR_i \subseteq [i]$ & Requirement~\ref{req:ar_step} \\
$s_\theta$ & Score function / denoising neural network with parameters $\theta$ & §\ref{sec:prelim} \\
$N$ & Length of the video to be generated & Algorithm~\ref{algo:framework} \\
$K$ & Number of auto-regressive steps: $K = \lceil N/\Delta \rceil$ & §5.3 \\
\bottomrule
\end{tabular}
\end{table}

\subsection{Denoising Steps}
\vspace{-1em}
\begin{table}[H]
\centering
\small
\begin{tabular}{@{}p{0.2\textwidth}p{0.6\textwidth}p{0.15\textwidth}@{}}
\toprule
\textbf{Symbol} & \textbf{Description} & \textbf{Defined in} \\
\midrule
$M_{\init}$ & Number of discretization steps in the initialization stage & §5.1 \\
$M_{\ar}$ & Number of discretization steps in the auto-regressive stage & §5.2 \\
$t_n^{\init}$ & Noise level at step $n$ in the initialization stage: $0 = t_0^{\init} \leq \ldots \leq t_{M_{\init}}^{\init} = T$ & §5.1 \\
$t_n^{\ar}$ & Noise level at step $n$ in the AR stage: $t_1^{\rmO} = t_0^{\ar} \leq \ldots \leq t_{M_{\ar}}^{\ar} = t_1^{\rmI}$ & §5.2 \\
$\tilde{t}_n^{\init}$ & Reverse noise level: $\tilde{t}_n^{\init} = T - t_{M_{\init}-n}^{\init}$ & §5.1 \\
$\tilde{t}_n^{\ar}$ & Reverse noise level: $\tilde{t}_n^{\ar} = T - t_{M_{\ar}-n}^{\ar}$ & §5.2 \\
$\bar{\tilde{t}}_n^{\init}$ & Multi-frame reverse noise levels induced by $\tilde{t}_n^{\init}$: $\bar{t}(\tilde{t}_n^{\init})$ & §5.1 \\
$\bar{\tilde{t}}_n^{\ar}$ & Multi-frame reverse noise levels induced by $\tilde{t}_n^{\ar}$: $\bar{t}(\tilde{t}_n^{\ar})$ & §5.2 \\
\bottomrule
\end{tabular}
\end{table}

\subsection{Theoretical Analysis}

\begin{table}[H]
\centering
\small
\begin{tabular}{@{}p{0.2\textwidth}p{0.6\textwidth}p{0.15\textwidth}@{}}
\toprule
\textbf{Symbol} & \textbf{Description} & \textbf{Defined in} \\
\midrule
$\calH(\cdot)$ & History frames function: all already generated frames before implementing \texttt{Auto-Regressive Step} & Appendix~\ref{app:joint} \\
$\calG(\cdot)$ & Generated frames function: all generated frames after implementing \texttt{Auto-Regressive Step} & Appendix~\ref{app:joint} \\
$\calS^{\rmI}$ & Set of starting indices of each input noise level & Appendix~\ref{app:joint} \\
$s^{\rmI}(t_i^{\rmI})$ & Smallest index whose input noise level equals $t_i^{\rmI}$ & Appendix~\ref{app:joint} \\
$i_0$ & Number of initially generated clean frames & §5.1 \\
$L$ & Lipschitz constant of the score function & §5.3 \\
$B$ & Bound on the norm of video frames: $\|X_{1:w}^0\| \leq B$ & §5.3 \\
\bottomrule
\end{tabular}
\end{table}

\subsection{Error Terms}

\begin{table}[H]
\centering
\small
\begin{tabular}{@{}p{0.2\textwidth}p{0.6\textwidth}p{0.15\textwidth}@{}}
\toprule
\textbf{Symbol} & \textbf{Description} & \textbf{Defined in} \\
\midrule
$\rmI\rmE$ & Initialization error & §5.3 \\
$\rmA\rmR\rmE_k$ & Auto-regressive error at step $k$ & §5.3 \\
$\mathrm{NIE}$ & Noise initialization error in initialization stage: $(d i_0 + B^2)\exp(-T)$ & §5.3 \\
$\mathrm{NIE}_{\rmA\rmR}$ & Noise initialization error in AR stage: $(\Delta d + B^2)\exp(-T)$ & §5.3 \\
$\mathrm{SEE}$ & Score estimation error in initialization stage: $T\epsilon_{\est}^2$ & §5.3 \\
$\mathrm{SEE}_{\rmA\rmR}$ & Score estimation error in AR stage: $(t_1^{\rmI} - t_1^{\rmO})\epsilon_{\est}^2$ & §5.3 \\
$\mathrm{DE}$ & Discretization error in initialization stage: $wdL^2\sum_{n=1}^{M_{\init}}(t_n^{\init} - t_{n-1}^{\init})^2$ & §5.3 \\
$\mathrm{DE}_{\rmA\rmR}$ & Discretization error in AR stage: $wdL^2\sum_{n=1}^{M_{\ar}}(t_n^{\ar} - t_{n-1}^{\ar})^2$ & §5.3 \\
$\mathrm{HF}_k$ & History forgetting at step $k$: $I(\texttt{Output}_k; \texttt{Past}_k | \texttt{Input}_k)$ & §5.3 \\
$\texttt{Output}_k$ & Output frames from $k$-th AR step: $\{X_{k\Delta+j}^{t_j^{\rmO}}\}_{j=1}^w$ & §5.3 \\
$\texttt{Past}_k$ & Past frames: $\calH(\{X_{k\Delta+j}^{t_j^{\rmO}}\}_{j=1}^w)\backslash\{X_{\calR_{k\Delta+1}}^0\}$ & §5.3 \\
$\texttt{Input}_k$ & Input to $k$-th AR step: $\{X_{\calR_{k\Delta+1}}^0\}\cup\{X_{k\Delta+j}^{t_j^{\rmI}}\}_{j=1}^w$ & §5.3 \\
$\epsilon_{\est}$ & Score estimation error bound & §5.3 \\
\bottomrule
\end{tabular}
\end{table}

\subsection{Divergence and Distance Measures}

\begin{table}[H]
\centering
\small
\begin{tabular}{@{}p{0.2\textwidth}p{0.6\textwidth}p{0.15\textwidth}@{}}
\toprule
\textbf{Symbol} & \textbf{Description} & \textbf{Defined in} \\
\midrule
$\kl(\cdot\|\cdot)$ & Kullback-Leibler divergence & §5.3 \\
$I(\cdot;\cdot|\cdot)$ & Conditional mutual information (used in history forgetting $\mathrm{HF}_k$) & §5.3 \\
$\mathbb{E}[\cdot]$ & Expectation operator & §\ref{sec:prelim} \\
$\|\cdot\|$ & $\ell_2$-norm & §\ref{sec:prelim} \\
\bottomrule
\end{tabular}
\end{table}



\section{Proof of Theorem~\ref{thm:main}}\label{app:main_proof}
 We provide the proof of Theorem~\ref{thm:main} in this section. We respectively prove Eqn.~\eqref{res:joint} and \eqref{res:recursive} in Sections~\ref{app:joint} and \ref{app:recursive}.

 \begin{figure}[t]
\centering
 \includegraphics[trim={3.0cm, 0.5cm, 3.0cm, 0.1cm}, clip, width=1.0\textwidth]{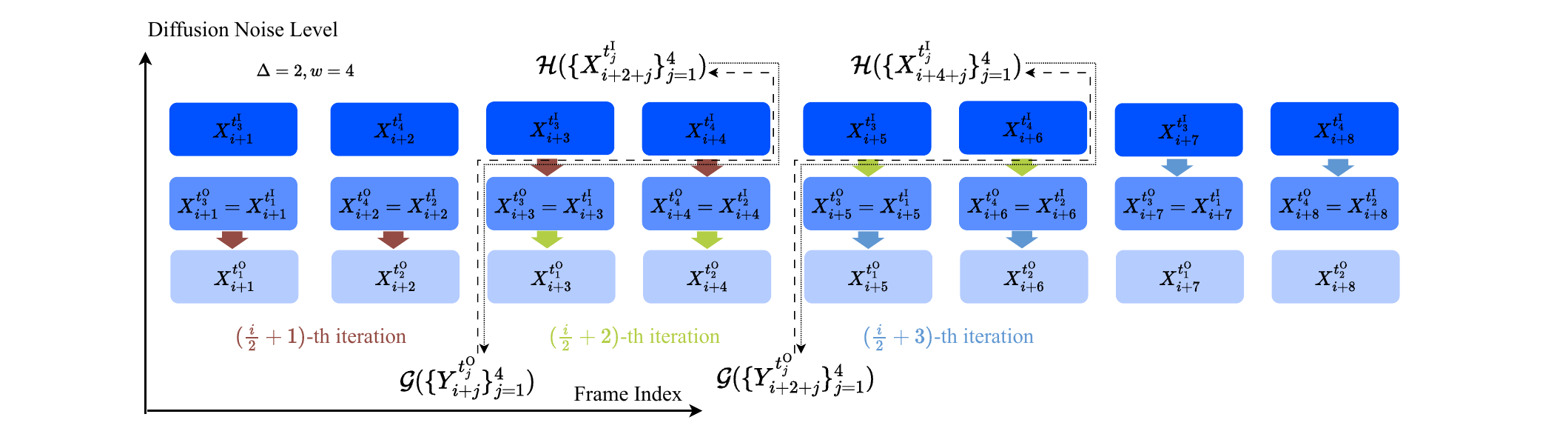}
\caption{The examples of $\calG(\cdot)$ and $\calH(\cdot)$.}
\label{fig:theory_fig}
\end{figure}
 \subsection{Proof of Eqn.~(\ref{res:joint})}\label{app:joint}

    Our proof consists of five main procedures.
    \begin{itemize}
        \item Decomposition of the concatenation of all the noisy and clean frames.
        \item Decomposition of the KL-divergence of all the frames into the KL-divergence of each auto-regressive generation step.
        \item Bounding the denoising error in the KL-divergence decomposition.
        \item Bounding the initialization error in the KL-divergence decomposition.
        \item Concluding the proof.
    \end{itemize}

    \textbf{Step 1: Decomposition of the concatenation of all the noisy and clean frames.}

    During each implementation of \texttt{Auto-Regressive Step}, we denoise the random vectors $\{Y_{i+j}^{t_{j}^{\rmI}}\}_{j=1}^{w}$. We define all the already generated random vectors, including inputs and outputs of \texttt{Auto-Regressive Step}, \emph{by this implementation} as
    \begin{align}
        \calH\Big(\big\{Y_{i+j}^{t_{j}^{\rmI}}\big\}_{j=1}^{w}\Big)=\{X_{1:i}^{0}\}\cup \big\{X_{1:i+j-1}^{t_{j}^{\rmI}}\big\}_{j\in\calS^{\rmI}} \text{ for }i\geq \Delta, \text{ and }\calH\Big(\big\{Y_{i+j}^{t_{j}^{\rmI}}\big\}_{j=1}^{w}\Big)=\emptyset\text{ for }i<\Delta,\label{eq:h}
    \end{align}
    where the set $\calS^{\rmI}\subseteq \calL^{\rmI}$ is the set of the starting indexes of each noise input noise level. It is defined as follows.
    \begin{align*}
        \calS^{\rmI}=\big\{s^{\rmI}\big(t_{j}^{\rmI}\big)\,|\, j\in [w]\big\}, \text{ and }s^{\rmI}\big(t_{i}^{\rmI}\big)=\min\big\{j\,|\, t_{j}^{\rmI}=t_{i}^{\rmI}, j\in[w]\big\},
    \end{align*}
    where $s^{\rmI}(t_{i}^{\rmI})$ is the smallest index whose input noise level is equal to $t_{i}^{\rmI}$. The examples of this definition of $\calH(\cdot )$ are provided in Figure~\ref{fig:theory_fig}. After each implementation of \texttt{Auto-Regressive Step}, we output $\{Y_{i+j}^{t_{j}^{\rmO}}\}_{j=1}^{w}$. We define all the generated random vectors, including inputs and outputs of \texttt{Auto-Regressive Step}, \emph{after this implementation} as
    \begin{align}
        \calG\Big(\big\{Y_{i+j}^{t_{j}^{\rmO}}\big\}_{j=1}^{w}\Big)= \calH\Big(\big\{Y_{i+j}^{t_{j}^{\rmI}}\big\}_{j=1}^{w}\Big)\cup\big\{Y_{i+j}^{t_{j}^{\rmI}}\big\}_{j=1}^{w}\cup \big\{Y_{i+j}^{t_{j}^{\rmO}}\big\}_{j=1}^{w}.\label{eq:g}
    \end{align}
    With these definitions, we can prove the following relationship between them.
    \begin{proposition}\label{prop:h_g}
        For any $i\geq\Delta$, we have that
        \begin{align*}
            \calG\Big(\big\{Y_{i-\Delta+j}^{t_{j}^{\rmO}}\big\}_{j=1}^{w}\Big)=\calH\Big(\big\{Y_{i+j}^{t_{j}^{\rmI}}\big\}_{j=1}^{w}\Big)\cup \big\{Y_{i+j}^{t_{j}^{\rmI}}\big\}_{j=1}^{w-\Delta}.
        \end{align*}
    \end{proposition}
    The proof of this proposition is provided in Appendix~\ref{app:h_g}.

    \textbf{Step 2: Decomposition of the KL-divergence of all the frames into the KL-divergence of each auto-regressive generation step. }

    Then we would like to decompose the upper bound of the KL-divergence between videos. In fact, we have that
    \begin{align}
        &\kl(X_{1:K\Delta}^{0}\,\|\,Y_{1:K\Delta}^{0})\nonumber\\
        &\quad\leq \kl\bigg(\calG\Big(\big\{X_{(K-1)\Delta+j}^{t_{j}^{\rmO}}\big\}_{j=1}^{w}\Big)\,\bigg\|\,\calG\Big(\big\{Y_{(K-1)\Delta+j}^{t_{j}^{\rmO}}\big\}_{j=1}^{w}\Big)\bigg)\nonumber\\
        &\quad = \kl\bigg(\calG\Big(\big\{X_{0+j}^{t_{j}^{\rmO}}\big\}_{j=1}^{w}\Big)\,\bigg\|\,\calG\Big(\big\{Y_{0+j}^{t_{j}^{\rmO}}\big\}_{j=1}^{w}\Big)\bigg)\nonumber\\
        &\quad\quad+\sum_{k=1}^{K-2}\kl\bigg(\{X_{k\Delta+j}^{t_{j}^{\rmI}}\}_{j=w-\Delta+1}^{w}\Big|\calG\Big(\big\{X_{(k-1)\Delta+j}^{t_{j}^{\rmO}}\big\}_{j=1}^{w}\Big)\,\bigg\|\,\{Y_{k\Delta+j}^{t_{j}^{\rmI}}\}_{j=w-\Delta+1}^{w}\Big|\calG\Big(\big\{Y_{(k-1)\Delta+j}^{t_{j}^{\rmO}}\big\}_{j=1}^{w}\Big)\bigg)\nonumber\\
        &\quad\quad+\sum_{k=1}^{K-2}\kl\bigg(\{X_{k\Delta+j}^{t_{j}^{\rmO}}\}_{j=1}^{w}\Big|\calG\Big(\big\{X_{(k-1)\Delta+j}^{t_{j}^{\rmO}}\big\}_{j=1}^{w}\Big)\cup \{X_{k\Delta+j}^{t_{j}^{\rmI}}\}_{j=w-\Delta+1}^{w}\,\nonumber\\
        &\quad\qquad\qquad\qquad\qquad\qquad\qquad\qquad \bigg\|\,\{Y_{k\Delta+j}^{t_{j}^{\rmO}}\}_{j=1}^{w}\Big|\calG\Big(\big\{Y_{(k-1)\Delta+j}^{t_{j}^{\rmO}}\big\}_{j=1}^{w}\Big)\cup \{Y_{k\Delta+j}^{t_{j}^{\rmI}}\}_{j=w-\Delta+1}^{w}\bigg)\nonumber\\
        &\quad = \text{(I)}+\text{(II)}+\text{(III)},\label{ieq:kl_1}
    \end{align}
    where the definition of the KL-divergence between conditional distributions is provided in Appendix~\ref{app:notation}, the inequality results from the data processing inequality, and the first equality results from the chain-rule of KL-divergence and Proposition~\ref{prop:h_g}. To see this, we have that
    \begin{align*}
        &\calG\Big(\big\{Y_{(k-1)\Delta+j}^{t_{j}^{\rmO}}\big\}_{j=1}^{w}\Big)\cup \{Y_{k\Delta+j}^{t_{j}^{\rmI}}\}_{j=w-\Delta+1}^{w}\cup \{Y_{k\Delta+j}^{t_{j}^{\rmO}}\}_{j=1}^{w}\\
        &\quad = \calH\Big(\big\{Y_{k\Delta+j}^{t_{j}^{\rmO}}\big\}_{j=1}^{w}\Big)\cup\big\{Y_{k\Delta+j}^{t_{j}^{\rmI}}\big\}_{j=1}^{w-\Delta}\cup\{Y_{k\Delta+j}^{t_{j}^{\rmI}}\}_{j=w-\Delta+1}^{w}\cup \{Y_{k\Delta+j}^{t_{j}^{\rmO}}\}_{j=1}^{w}\\
        &\quad = \calH\Big(\big\{Y_{k\Delta+j}^{t_{j}^{\rmO}}\big\}_{j=1}^{w}\Big)\cup\big\{Y_{k\Delta+j}^{t_{j}^{\rmI}}\big\}_{j=1}^{w}\cup \{Y_{k\Delta+j}^{t_{j}^{\rmO}}\}_{j=1}^{w}\\
        &\quad = \calG\Big(\big\{Y_{k\Delta+j}^{t_{j}^{\rmO}}\big\}_{j=1}^{w}\Big),
    \end{align*}
    where the first equality results from Proposition~\ref{prop:h_g}, the second equality results from combining the middle two terms, and the last equality results from the definition of $\calG$ in Eqn.~\eqref{eq:g}. We note that in the right-hand side of Eqn.~\eqref{ieq:kl_1}, the term (I) corresponds to the error from the initialization stage of Algorithm~\ref{algo:framework}, the term (II) corresponds to the error from the noise initialization (Line~\ref{line:noise_init} of Algorithm~\ref{algo:ar_step}), and the term (III) corresponds to the error from the denoising step (Line~\ref{line:denoise} of Algorithm~\ref{algo:ar_step}). We would like to further decompose the term (II). We first transform each term in (II) as follows.
    \begin{align}
        & \kl\bigg(\{X_{k\Delta+j}^{t_{j}^{\rmO}}\}_{j=1}^{w}\Big|\calG\Big(\big\{X_{(k-1)\Delta+j}^{t_{j}^{\rmO}}\big\}_{j=1}^{w}\Big)\cup \{X_{k\Delta+j}^{t_{j}^{\rmI}}\}_{j=w-\Delta+1}^{w}\,\nonumber\\
        &\quad\qquad\qquad\qquad\qquad\qquad\qquad\qquad \bigg\|\,\{Y_{k\Delta+j}^{t_{j}^{\rmO}}\}_{j=1}^{w}\Big|\calG\Big(\big\{Y_{(k-1)\Delta+j}^{t_{j}^{\rmO}}\big\}_{j=1}^{w}\Big)\cup \{Y_{k\Delta+j}^{t_{j}^{\rmI}}\}_{j=w-\Delta+1}^{w}\bigg)\nonumber\\
        &=\kl\bigg(\{X_{k\Delta+j}^{t_{j}^{\rmO}}\}_{j=1}^{w}\Big|\calH\Big(\big\{X_{k\Delta+j}^{t_{j}^{\rmO}}\big\}_{j=1}^{w}\Big)\cup \{X_{k\Delta+j}^{t_{j}^{\rmI}}\}_{j=1}^{w}\,\nonumber\\
        &\quad\qquad\qquad\qquad\qquad\qquad\qquad\qquad \bigg\|\,\{Y_{k\Delta+j}^{t_{j}^{\rmO}}\}_{j=1}^{w}\Big|\calH\Big(\big\{Y_{k\Delta+j}^{t_{j}^{\rmO}}\big\}_{j=1}^{w}\Big)\cup \{Y_{k\Delta+j}^{t_{j}^{\rmI}}\}_{j=1}^{w}\bigg),\label{eq:2_eq}
    \end{align}
    where the equality follows from Proposition~\ref{prop:h_g}. From the denoising step in Eqn.~\eqref{eq:multi_reverse_score}, we have the following Markov chain.
    \begin{align}
        \calH\Big(\big\{Y_{k\Delta+j}^{t_{j}^{\rmO}}\big\}_{j=1}^{w}\Big)\backslash \big\{Y_{\calR_{k\Delta+1}^{0}}\big\}\mbox{---}\big\{Y_{\calR_{k\Delta+1}^{0}}\big\}\cup\big\{Y_{k\Delta+j}^{t_{j}^{\rmI}}\big\}_{j=1}^{w}\mbox{---}\big\{Y_{k\Delta+j}^{t_{j}^{\rmO}}\big\}_{j=1}^{w}.\label{eq:mc}
    \end{align}
    Thus, we have that
    \begin{align}
        &\kl\bigg(\{X_{k\Delta+j}^{t_{j}^{\rmO}}\}_{j=1}^{w}\Big|\calH\Big(\big\{X_{k\Delta+j}^{t_{j}^{\rmO}}\big\}_{j=1}^{w}\Big)\cup \{X_{k\Delta+j}^{t_{j}^{\rmI}}\}_{j=1}^{w}\,\nonumber\\*
        &\quad\qquad\qquad\qquad\qquad\qquad\qquad\qquad \bigg\|\,\{Y_{k\Delta+j}^{t_{j}^{\rmO}}\}_{j=1}^{w}\Big|\calH\Big(\big\{Y_{k\Delta+j}^{t_{j}^{\rmO}}\big\}_{j=1}^{w}\Big)\cup \{Y_{k\Delta+j}^{t_{j}^{\rmI}}\}_{j=1}^{w}\bigg)\nonumber\\
        &\quad = \kl\bigg(\{X_{k\Delta+j}^{t_{j}^{\rmO}}\}_{j=1}^{w}\Big|\calH\Big(\big\{X_{k\Delta+j}^{t_{j}^{\rmO}}\big\}_{j=1}^{w}\Big)\cup \{X_{k\Delta+j}^{t_{j}^{\rmI}}\}_{j=1}^{w}\,\nonumber\\*
        &\quad\qquad\qquad\qquad\qquad\qquad\qquad\qquad \bigg\|\,\{Y_{k\Delta+j}^{t_{j}^{\rmO}}\}_{j=1}^{w}\Big|\big\{Y_{\calR_{k\Delta+1}^{0}}\big\}\cup \{Y_{k\Delta+j}^{t_{j}^{\rmI}}\}_{j=1}^{w}\bigg)\nonumber\\
        &\quad = I\bigg( \{X_{k\Delta+j}^{t_{j}^{\rmO}}\}_{j=1}^{w} ; \calH\Big(\big\{X_{k\Delta+j}^{t_{j}^{\rmO}}\big\}_{j=1}^{w}\Big)\backslash\big\{X_{\calR_{k\Delta+1}^{0}}\big\}\bigg|\big\{X_{\calR_{k\Delta+1}^{0}}\big\}\cup \{X_{k\Delta+j}^{t_{j}^{\rmI}}\}_{j=1}^{w}\bigg)\nonumber\\
        &\quad\quad\!\! + \kl\bigg(\{X_{k\Delta+j}^{t_{j}^{\rmO}}\}_{j=1}^{w}\Big|\big\{X_{\calR_{k\Delta+1}^{0}}\big\}\cup \{X_{k\Delta+j}^{t_{j}^{\rmI}}\}_{j=1}^{w}\bigg\|\,\{Y_{k\Delta+j}^{t_{j}^{\rmO}}\}_{j=1}^{w}\Big|\big\{Y_{\calR_{k\Delta+1}^{0}}\big\}\cup \{Y_{k\Delta+j}^{t_{j}^{\rmI}}\}_{j=1}^{w}\bigg),\label{eq:2_eq_1}
    \end{align}
    where the first equality results from the Markov chain in \eqref{eq:mc}, and the second equality results from the definition of the conditional mutual information.

    \textbf{Step 3: Bounding the denoising error in the KL-divergence decomposition.}

    Eqn.~\eqref{eq:2_eq} and \eqref{eq:2_eq_1} show that to bound the denoising error term (III) in Eqn.~\eqref{ieq:kl_1}, we only need to bound the second term in the right-hand side of Eqn.~\eqref{eq:2_eq_1}. We first define the multiple noise-level version of Eqn.~\eqref{eq:x_reverse} and apply the change of variable $t\rightarrow T-t$ as follows.
    \begin{align}
        \rmd \tilX_{1:w}^{\bart}&=\bigg(\frac{1}{2}\tilX_{1:w}^{\bart}+\nabla\log P_{X}^{\barT-\bart}(\tilX_{1:w}^{\bart}\,|\,X_{\calR}^{0})\bigg) \rmd t+\rmd B_{1:w}^{\bart} \text{ for }T-t_{1}^{\rmI}\leq t\leq T-t_{1}^{\rmO}.\label{eq:x_rev}
    \end{align}
    Recall that we define the joint time $\bart$ as $\bart(t)=(t,t+t_{2}^{\rmI}-t_{1}^{\rmI},\cdots,t+t_{w}^{\rmI}-t_{1}^{\rmI} )=(t_1,\cdots,t_w)$. We define the probability of $\tilX_{1:w}^{\bart}$ as $\tilP_{X}^{\bart}$. Then we have that $\tilP_{X}^{\bart}=P_{X}^{\barT-\bart}$. We also applied the change of variable to Eqn.~\eqref{eq:multi_reverse_score}.
    \begin{align}
        \rmd \tilY_{1:w}^{\bart}&=\bigg(\frac{1}{2}\tilY_{1:w}^{\bar{\tilt}_{n}^{\ar}}+s_{\theta}(\tilY_{1:w}^{\bar{\tilt}_{n}^{\ar}},\barT-\bar{\tilt}_{n}^{\ar},Y_{\calR}^{0})\bigg)\rmd t+\rmd B_{1:w}^{\bart} \text{ for }\tilt_{n}^{\ar}\leq t \leq \tilt_{n+1}^{\ar}.\label{eq:y_rev}
    \end{align}
    In the following proof of this step, we will omit the symbols $X_{\calR}^{0}$ and $Y_{\calR}^{0}$ for ease of notation, since all the proof in this step is conditioned on them. We define the probability of $\tilY_{1:w}^{\bart}$ as $\tilP_{Y}^{\bart}$ and have that $\tilP_{Y}^{\bart}=P_{Y}^{\barT-\bart}$. For notation simplicity, we define $\bart^{\rmI}=\barT-\calL^{\rmO}$ and $\bart^{\rmO}=\barT-\calL^{\rmI}$. Conditioned on the value at $\bart^{\prime}$, the conditional distribution of $\tilX_{1:w}^{\bart}$ and $\tilY_{1:w}^{\bart}$ are denoted as $\tilP_{X}^{\bart|\bart^{\prime}}$ and $\tilP_{Y}^{\bart|\bart^{\prime}}$, respectively. For any $\tilt_{n}^{\ar}\leq t \leq \tilt_{n+1}^{\ar}$,  Lemma~\ref{lem:diff} shows that
    \begin{align*}
        &\frac{\rmd}{\rmd t}\kl\big(\tilP_{X}^{\bart|\bart^{\prime}}(\cdot|a)\,\big\|\,\tilP_{Y}^{\bart|\bart^{\prime}}(\cdot|a)\big)\nonumber\\
        &\quad= -\bbE_{\tilP_{X}^{\bart|\bart^{\prime}}(\cdot|a)\cdot\tilP_{X}^{\bart^{\prime}}(a)}\bigg[\bigg\|\nabla\log\frac{\tilP_{X}^{\bart|\bart^{\prime}}(\tilX_{1:w}^{\bart}|a)}{\tilP_{Y}^{\bart|\bart^{\prime}}(\tilX_{1:w}^{\bart}|a)}\bigg\|^{2}\bigg]\nonumber\\
        &\quad\qquad+\bbE_{\tilP_{X}^{\bart|\bart^{\prime}}(\cdot|a)\cdot\tilP_{X}^{\bart^{\prime}}(a)}\bigg[\bigg\langle \nabla\log P_{X}^{\barT-\bart}(\tilX_{1:w}^{\bart}\,|\,X_{\calR}^{0})-s_{\theta}(\tilX_{1:w}^{\bar{\tilt}_{n}^{\ar}},\barT-\bar{\tilt}_{n}^{\ar},X_{\calR}^{0}),\nabla\log\frac{\tilP_{X}^{\bart|\bart^{\prime}}(\tilX_{1:w}^{\bart}|a)}{\tilP_{Y}^{\bart|\bart^{\prime}}(\tilX_{1:w}^{\bart}|a)}\bigg\rangle\bigg]\nonumber\\
        &\quad\leq\frac{1}{2}\bbE_{\tilP_{X}^{\bart|\bart^{\prime}}(\cdot|a)\cdot\tilP_{X}^{\bart^{\prime}}(a)}\Big[\big\|\nabla\log P_{X}^{\barT-\bart}(\tilX_{1:w}^{\bart}\,|\,X_{\calR}^{0})-s_{\theta}(\tilX_{1:w}^{\bar{\tilt}_{n}^{\ar}},\barT-\bar{\tilt}_{n}^{\ar},X_{\calR}^{0})\big\|^{2}\Big],
    \end{align*}
    where the inequality results from Cauchy inequality. Then we have that
    \begin{align*}
        &\kl\big(\tilP_{X}^{\bart^{\rmO}|\bart^{\rmI}}(\cdot|a)\,\big\|\,\tilP_{Y}^{\bart^{\rmO}|\bart^{\rmI}}(\cdot|a)\big)\\
        &\quad =\kl\big(\tilP_{X}^{\bart^{\rmO}|\bart^{\rmI}}(\cdot|a)\,\big\|\,\tilP_{Y}^{\bart^{\rmO}|\bart^{\rmI}}(\cdot|a)\big)-\kl\big(\tilP_{X}^{\bart^{\rmI}|\bart^{\rmI}}(\cdot|a)\,\big\|\,\tilP_{Y}^{\bart^{\rmI}|\bart^{\rmI}}(\cdot|a)\big)\nonumber\\
        &\quad \leq \frac{1}{2}\sum_{n=1}^{M_{\ar}}\int_{\tilt_{n-1}^{\ar}}^{\tilt_{n}^{\ar}}\bbE_{\tilP_{X}^{\bart|\bart^{\rmI}}(\cdot|a)\cdot\tilP_{X}^{\bart^{\rmI}}(a)}\Big[\big\|\nabla\log P_{X}^{\barT-\bart}(\tilX_{1:w}^{\bart}\,|\,X_{\calR}^{0})-s_{\theta}(\tilX_{1:w}^{\bart},\barT-\bar{\tilt}_{n}^{\ar},X_{\calR}^{0})\big\|^{2}\Big]\rmd t.
    \end{align*}
    Thus, for the second term on the right-hand side of Eqn.~\eqref{eq:2_eq_1}, we have that
    \begin{align}
        &\kl\bigg(\{X_{k\Delta+j}^{t_{j}^{\rmO}}\}_{j=1}^{w}\Big|\big\{X_{\calR_{k\Delta+1}^{0}}\big\}\cup \{X_{k\Delta+j}^{t_{j}^{\rmI}}\}_{j=1}^{w}\bigg\|\,\{Y_{k\Delta+j}^{t_{j}^{\rmO}}\}_{j=1}^{w}\Big|\big\{Y_{\calR_{k\Delta+1}^{0}}\big\}\cup \{Y_{k\Delta+j}^{t_{j}^{\rmI}}\}_{j=1}^{w}\bigg)\nonumber\\
        &\quad\leq \frac{1}{2}\sum_{n=1}^{M_{\ar}}\int_{t_{n-1}^{\ar}}^{t_{n}^{\ar}}\bbE\Big[\big\|\nabla\log P_{X}^{\bart}(X_{k\Delta+1:k\Delta+w}^{\bart}\,|\,X_{\calR_{k\Delta+1}}^{0})-s_{\theta}(X_{k\Delta+1:k\Delta+w}^{\bart_{n}^{\ar}},\bart_{n}^{\ar},X_{\calR_{k\Delta+1}}^{0})\big\|^{2}\Big]\rmd t.\label{ieq:dkl}
    \end{align}
    Then we would like to upper bound each term in the right-hand side of Eqn.~\eqref{ieq:dkl}. In fact, we have that
    \begin{align}
        &\int_{t_{n-1}^{\ar}}^{t_{n}^{\ar}}\bbE\Big[\big\|\nabla\log P_{X}^{\bart}(X_{k\Delta+1:k\Delta+w}^{\bart}\,|\,X_{\calR_{k\Delta+1}}^{0})-s_{\theta}(X_{k\Delta+1:k\Delta+w}^{\bart_{n}^{\ar}},\bart_{n}^{\ar},X_{\calR_{k\Delta+1}}^{0})\big\|^{2}\Big]\rmd t\nonumber\\
        &\quad\leq 2\int_{t_{n-1}^{\ar}}^{t_{n}^{\ar}}\bbE\Big[\big\|\nabla\log P_{X}^{\bart}(X_{k\Delta+1:k\Delta+w}^{\bart}\,|\,X_{\calR_{k\Delta+1}}^{0})-\nabla\log P_{X}^{\bart_{n}^{\ar}}(X_{k\Delta+1:k\Delta+w}^{\bart_{n}^{\ar}}\,|\,X_{\calR_{k\Delta+1}}^{0})\big\|^{2}\Big]\rmd t\nonumber\\
        &\quad\qquad +2\int_{t_{n-1}^{\ar}}^{t_{n}^{\ar}}\bbE\Big[\big\|\nabla\log P_{X}^{\bart_{n}^{\ar}}(X_{k\Delta+1:k\Delta+w}^{\bart_{n}^{\ar}}\,|\,X_{\calR_{k\Delta+1}}^{0})-s_{\theta}(X_{k\Delta+1:k\Delta+w}^{\bart_{n}^{\ar}},\bart_{n}^{\ar},X_{\calR_{k\Delta+1}}^{0})\big\|^{2}\Big]\rmd t\nonumber\\
        &\quad =2\int_{t_{n-1}^{\ar}}^{t_{n}^{\ar}}\bbE\Big[\big\|\nabla\log P_{X}^{\bart}(X_{k\Delta+1:k\Delta+w}^{\bart}\,|\,X_{\calR_{k\Delta+1}}^{0})-\nabla\log P_{X}^{\bart_{n}^{\ar}}(X_{k\Delta+1:k\Delta+w}^{\bart_{n}^{\ar}}\,|\,X_{\calR_{k\Delta+1}}^{0})\big\|^{2}\Big]\rmd t\nonumber\\
        &\quad\qquad +2(t_{n}^{\ar}-t_{n-1}^{\ar})\bbE\Big[\big\|\nabla\log P_{X}^{\bart_{n}^{\ar}}(X_{k\Delta+1:k\Delta+w}^{\bart_{n}^{\ar}}\,|\,X_{\calR_{k\Delta+1}}^{0})-s_{\theta}(X_{k\Delta+1:k\Delta+w}^{\bart_{n}^{\ar}},\bart_{n}^{\ar},X_{\calR_{k\Delta+1}}^{0})\big\|^{2}\Big],\label{ieq:est_err}
    \end{align}
    where the inequality results from the property of the $\ell_2$-norm. Then 
    we use the following proposition to upper bound the first term in the right-hand side of Eqn.~\eqref{ieq:est_err}.
    \begin{proposition}\label{prop:lips}
        Let $P^{t}$ denote the marginal distribution function of the following stochastic process at time $t$.
        \begin{align*}
            \rmd X^{t}&=-\frac{1}{2}X^{t}\rmd t+\rmd B^{t},\quad X^{0}\sim P_{X}.
        \end{align*}
        If $\nabla\log P^t$ is $L$-Lipschitz for $t\in[t_{0},t_{1}]$ with $t_{1}-t_{0}\leq 1$. Then for any $t_{0}\leq u<v\leq t_{1}$, we have that
        \begin{align*}
            \bbE\Big[\big\|\nabla\log P^{u}(X^u)-\nabla\log P^v(X^v)\big\|^{2}\Big]\lesssim dL^2(v-u).
        \end{align*}
    \end{proposition}
    The proof is provided in Appendix~\ref{app:lips}. This proposition implies that
    \begin{align}
        &\int_{t_{n-1}^{\ar}}^{t_{n}^{\ar}}\bbE\Big[\big\|\nabla\log P_{X}^{\bart}(X_{k\Delta+1:k\Delta+w}^{\bart}\,|\,X_{\calR_{k\Delta+1}}^{0})-\nabla\log P_{X}^{\bart_{n}^{\ar}}(X_{k\Delta+1:k\Delta+w}^{\bart_{n}^{\ar}}\,|\,X_{\calR_{k\Delta+1}}^{0})\big\|^{2}\Big]\rmd t\nonumber\\
        &\quad\lesssim \int_{t_{n-1}^{\ar}}^{t_{n}^{\ar}}wdL^{2}(t_{n}^{\ar}-t)\rmd t\nonumber\\
        &\quad\lesssim wdL^{2}(t_{n}^{\ar}-t_{n-1}^{\ar})^2.\label{ieq:dis_err}
    \end{align}
    Combining Eqn.~\eqref{ieq:dkl}, \eqref{ieq:est_err}, \eqref{ieq:dis_err}, we have that
    \begin{align}
        &\kl\bigg(\{X_{k\Delta+j}^{t_{j}^{\rmO}}\}_{j=1}^{w}\Big|\big\{X_{\calR_{k\Delta+1}^{0}}\big\}\cup \{X_{k\Delta+j}^{t_{j}^{\rmI}}\}_{j=1}^{w}\bigg\|\,\{Y_{k\Delta+j}^{t_{j}^{\rmO}}\}_{j=1}^{w}\Big|\big\{Y_{\calR_{k\Delta+1}^{0}}\big\}\cup \{Y_{k\Delta+j}^{t_{j}^{\rmI}}\}_{j=1}^{w}\bigg)\nonumber\\
        &\quad\lesssim \sum_{n=1}^{M_{\ar}}(t_{n}^{\ar}-t_{n-1}^{\ar})\bbE\Big[\big\|\nabla\log P_{X}^{\bart_{n}^{\ar}}(X_{k\Delta+1:k\Delta+w}^{\bart_{n}^{\ar}}\,|\,X_{\calR_{k\Delta+1}}^{0})-s_{\theta}(X_{k\Delta+1:k\Delta+w}^{\bart_{n}^{\ar}},\bart_{n}^{\ar},X_{\calR_{k\Delta+1}}^{0})\big\|^{2}\Big]\nonumber\\
        &\quad\qquad +wdL^{2}\sum_{n=1}^{M_{\ar}}(t_{n}^{\ar}-t_{n-1}^{\ar})^2.\label{ieq:dkl2}
    \end{align}

    \textbf{Step 4: Bounding the initialization error in the KL-divergence decomposition.}

    We note that there are two initialization error terms in Eqn.~\eqref{ieq:kl_1}, i.e., (I) and (II). Here (I) arises from Line~\ref{line:init} of Algorithm~\ref{algo:framework}, while (II) arises from Line~\ref{line:noise_init} of Algorithm~\ref{algo:ar_step}. We first derive the bound for term (II). In fact, we have that 
    \begin{align}
        &\kl\bigg(\{X_{k\Delta+j}^{t_{j}^{\rmI}}\}_{j=w-\Delta+1}^{w}\Big|\calG\Big(\big\{X_{(k-1)\Delta+j}^{t_{j}^{\rmO}}\big\}_{j=1}^{w}\Big)\,\bigg\|\,\{Y_{k\Delta+j}^{t_{j}^{\rmI}}\}_{j=w-\Delta+1}^{w}\Big|\calG\Big(\big\{Y_{(k-1)\Delta+j}^{t_{j}^{\rmO}}\big\}_{j=1}^{w}\Big)\bigg)\nonumber\\
        &\quad =\kl\bigg(\{X_{k\Delta+j}^{T}\}_{j=w-\Delta+1}^{w}\Big|\calG\Big(\big\{X_{(k-1)\Delta+j}^{t_{j}^{\rmO}}\big\}_{j=1}^{w}\Big)\,\bigg\|\,\calN(0,I)\bigg)\nonumber\\
        &\quad = \kl\bigg(\{X_{k\Delta+j}^{T}\}_{j=w-\Delta+1}^{w}\Big|\calG\Big(\big\{X_{(k-1)\Delta+j}^{t_{j}^{\rmO}}\big\}_{j=1}^{w}\Big)\cup \{X_{k\Delta+j}^{0}\}_{j=w-\Delta+1}^{w}\,\bigg\|\,\calN(0,I)\bigg)\nonumber\\
        &\quad\qquad - \kl\bigg(\{X_{k\Delta+j}^{T}\}_{j=w-\Delta+1}^{w}\Big|\calG\Big(\big\{X_{(k-1)\Delta+j}^{t_{j}^{\rmO}}\big\}_{j=1}^{w}\Big)\cup \{X_{k\Delta+j}^{0}\}_{j=w-\Delta+1}^{w}\nonumber\\
        &\quad\qquad\qquad\qquad\qquad\qquad\qquad\qquad\qquad \,\bigg\|\,\{X_{k\Delta+j}^{T}\}_{j=w-\Delta+1}^{w}\Big|\calG\Big(\big\{X_{(k-1)\Delta+j}^{t_{j}^{\rmO}}\big\}_{j=1}^{w}\Big)\bigg)\nonumber\\
        &\quad\leq \kl\bigg(\{X_{k\Delta+j}^{T}\}_{j=w-\Delta+1}^{w}\Big|\calG\Big(\big\{X_{(k-1)\Delta+j}^{t_{j}^{\rmO}}\big\}_{j=1}^{w}\Big)\cup \{X_{k\Delta+j}^{0}\}_{j=w-\Delta+1}^{w}\,\bigg\|\,\calN(0,I)\bigg)\nonumber\\
        &\quad=\kl\bigg(\{X_{k\Delta+j}^{T}\}_{j=w-\Delta+1}^{w}\Big| \{X_{k\Delta+j}^{0}\}_{j=w-\Delta+1}^{w}\,\bigg\|\,\calN(0,I)\bigg),\label{ieq:init_5}
    \end{align}
    where the first equality results from Line~\ref{line:noise_init} of Algorithm~\ref{algo:ar_step}, and the last equality results from the following Markov chain for the forward process of $X_{i}^{t}$
    \begin{align*}
        \{X_{k\Delta+j}^{T}\}_{j=w-\Delta+1}^{w}\mbox{---}\{X_{k\Delta+j}^{0}\}_{j=w-\Delta+1}^{w}\mbox{---}\calG\Big(\big\{X_{(k-1)\Delta+j}^{t_{j}^{\rmO}}\big\}_{j=1}^{w}\Big).
    \end{align*}
    With Assumption~\ref{assump:bound} and Lemma~\ref{lem:init}, we have that
    \begin{align}
        &\kl\bigg(\{X_{k\Delta+j}^{T}\}_{j=w-\Delta+1}^{w}\Big| \{X_{k\Delta+j}^{0}\}_{j=w-\Delta+1}^{w}\,\bigg\|\,\calN(0,I)\bigg)\leq (\Delta d+ B^2)\exp(-T).\label{ieq:init_6}
    \end{align}
    Combining these inequalities, we have that
    \begin{align}
        &\kl\bigg(\{X_{k\Delta+j}^{t_{j}^{\rmI}}\}_{j=w-\Delta+1}^{w}\Big|\calG\Big(\big\{X_{(k-1)\Delta+j}^{t_{j}^{\rmO}}\big\}_{j=1}^{w}\Big)\,\bigg\|\,\{Y_{k\Delta+j}^{t_{j}^{\rmI}}\}_{j=w-\Delta+1}^{w}\Big|\calG\Big(\big\{Y_{(k-1)\Delta+j}^{t_{j}^{\rmO}}\big\}_{j=1}^{w}\Big)\bigg)\nonumber\\
        &\quad\leq (d\Delta+ B^2)\exp(-T).\label{ieq:init_1}
    \end{align}
    We then upper bound the term (I). From the procedures of Algorithm~\ref{algo:init}, we have that
    \begin{align}
        & \kl\bigg(\calG\Big(\big\{X_{0+j}^{t_{j}^{\rmO}}\big\}_{j=1}^{w}\Big)\,\bigg\|\,\calG\Big(\big\{Y_{0+j}^{t_{j}^{\rmO}}\big\}_{j=1}^{w}\Big)\bigg)\nonumber\\
        &\quad =\kl\bigg(\big\{X_{i+j}^{t_{j}^{\rmI}}\big\}_{j=1}^{w}\cup \big\{X_{i+j}^{t_{j}^{\rmO}}\big\}_{j=1}^{w}\,\bigg\|\, \big\{Y_{i+j}^{t_{j}^{\rmI}}\big\}_{j=1}^{w}\cup \big\{Y_{i+j}^{t_{j}^{\rmO}}\big\}_{j=1}^{w}\bigg)\nonumber\\
        &\quad\leq \kl\big(X_{1:i_0}^{0}\,\big\|\,Y_{1:i_0}^{0}\big)\nonumber\\
        &\quad\leq \kl\big(X_{1:i_0}^{0}\big|X_{1:i_0}^{T}\,\big\|\,Y_{1:i_0}^{0}\big|Y_{1:i_0}^{T}\big)+\kl\big(X_{1:i_0}^{T}\,\big\|\,Y_{1:i_0}^{T}\big),\label{ieq:init_2}
    \end{align}
    where the equality follows from the definition of $\calG$ in Eqn.~\eqref{eq:g}, these two inequalities follows from the data processing inequality. For the first term in the right-hand side of Eqn.~\eqref{ieq:init_2}, we follow the same procedures in \textbf{Step 2} and derive that
    \begin{align}
        &\kl\big(X_{1:i_0}^{0}\big|X_{1:i_0}^{T}\,\big\|\,Y_{1:i_0}^{0}\big|Y_{1:i_0}^{T}\big)\nonumber\\
        &\quad\lesssim \sum_{n=1}^{M_{\ar}}(t_{n}^{\ar}-t_{n-1}^{\ar})\bbE\Big[\big\|\nabla\log P_{X}^{\bart_{n}^{\ar}}(X_{k\Delta+1:k\Delta+w}^{\bart_{n}^{\ar}}\,|\,X_{\calR_{k\Delta+1}}^{0})-s_{\theta}(X_{k\Delta+1:k\Delta+w}^{\bart_{n}^{\ar}},\bart_{n}^{\ar},X_{\calR_{k\Delta+1}}^{0})\big\|^{2}\Big]\nonumber\\
        &\quad\qquad +wdL^{2}\sum_{n=1}^{M_{\ar}}(t_{n}^{\ar}-t_{n-1}^{\ar})^2.\label{ieq:init_3}
    \end{align}
    For the second term in the right-hand side of Eqn.~\eqref{ieq:init_2}, Assumption~\eqref{assump:bound} and Lemma~\eqref{lem:init} show that
    \begin{align}
        \kl\big(X_{1:i_0}^{T}\,\big\|\,Y_{1:i_0}^{T}\big)\leq (d\cdot i_{0}+B^{2})\exp(-T).\label{ieq:init_4}
    \end{align}

    \textbf{Step 5: Concluding the proof.}

    Now we have derived all the components to upper bound the KL-divergence in Eqn.~\eqref{ieq:kl_1}. Combining Eqn.~\eqref{ieq:init_2}, \eqref{ieq:init_3} and \eqref{ieq:init_4}, we have that
    \begin{align}
        \text{(I)}&\lesssim (d\cdot i_{0}+B^{2})\exp(-T)\nonumber\\
        &\quad +\sum_{n=1}^{M_{\init}}(t_{n}^{\init}-t_{n-1}^{\init})\bbE\Big[\big\|\nabla\log P_{X}^{t_{n}^{\init}}(X_{k\Delta+1:k\Delta+w}^{t_{n}^{\init}})-s_{\theta}(X_{k\Delta+1:k\Delta+w}^{t_{n}^{\init}},t_{n}^{\init})\big\|^{2}\Big]\nonumber\\
        &\quad\qquad +wdL^{2}\sum_{n=1}^{M_{\init}}(t_{n}^{\init}-t_{n-1}^{\init})^2.\label{ieq:term_1}
    \end{align}
    Combining Eqn.~\eqref{ieq:init_5} and \eqref{ieq:init_6}, we have that
    \begin{align}
        \text{(II)}\leq K(\Delta d+ B^2)\exp(-T).\label{ieq:term_2}
    \end{align}
    Combining Eqn.~\eqref{eq:2_eq}, Eqn.~\eqref{eq:2_eq_1}, \eqref{ieq:dkl2}, we have that
    \begin{align}
        \text{(III)}
        & \leq \sum_{k=1}^{K-2}I\bigg( \{X_{k\Delta+j}^{t_{j}^{\rmO}}\}_{j=1}^{w} ; \calH\Big(\big\{X_{k\Delta+j}^{t_{j}^{\rmO}}\big\}_{j=1}^{w}\Big)\backslash\big\{X_{\calR_{k\Delta+1}}^{0}\big\}\bigg|\big\{X_{\calR_{k\Delta+1}}^{0}\big\}\cup \{X_{k\Delta+j}^{t_{j}^{\rmI}}\}_{j=1}^{w}\bigg)\nonumber\\
        &\quad+\sum_{k=1}^{K-2}\sum_{n=1}^{M_{\ar}}(t_{n}^{\ar}-t_{n-1}^{\ar})\bbE\Big[\big\|\nabla\log P_{X}^{\bart_{n}^{\ar}}(X_{k\Delta+1:k\Delta+w}^{\bart_{n}^{\ar}}\,|\,X_{\calR_{k\Delta+1}}^{0}) \notag\\
        &\qquad\qquad-s_{\theta}(X_{k\Delta+1:k\Delta+w}^{\bart_{n}^{\ar}},\bart_{n}^{\ar},X_{\calR_{k\Delta+1}}^{0})\big\|^{2}\Big]\nonumber\\
        &\quad +\sum_{k=1}^{K-2}wdL^{2}\sum_{n=1}^{M_{\ar}}(t_{n}^{\ar}-t_{n-1}^{\ar})^2.\label{ieq:term_3}
    \end{align}
    Combining Eqn.~\eqref{ieq:kl_1}, \eqref{ieq:term_1}, \eqref{ieq:term_2}, and \eqref{ieq:term_1}, we conclude the proof of Eqn.~\eqref{res:joint}.

\subsection{Proof of Eqn.~(\ref{res:recursive})}\label{app:recursive}
The proof of Eqn.~\eqref{res:recursive} largely shares the similar procedures with the proof of Eqn.~\eqref{res:joint}. Thus, we will reuse some intermediate results in the proof of Eqn.~\eqref{res:joint}. Here we consider the special case $\calR_{i}=\emptyset$, and the results of the general case $\calR_{i}\neq \emptyset$ can be derived by our analysis with complicated calculations.

The proof of Eqn.~(\ref{res:recursive}) consists of the following two steps.
\begin{itemize}
    \item Derive the recursive formula of the output error.
    \item Conclude the proof by aggregating the recursive formula.
\end{itemize}

\textbf{Step 1: Derive the recursive formula of the output error.}

To derive the KL-divergence between a generated clip and the nominal video clip, we upper bound the error as follows.
\begin{align}
    &\kl\big( X_{K\Delta+1:(K+1)\Delta}^{0}\big\| Y_{K\Delta+1:(K+1)\Delta}^{0} \big)\nonumber\\
    &\quad\leq \kl\Big(\big\{X_{K\Delta+i}^{t_{i}^{\rmO}}\big\}_{i=1}^{w}\Big\|\big\{Y_{K\Delta+i}^{t_{i}^{\rmO}}\big\}_{i=1}^{w}\Big)\nonumber\\
    &\quad\leq \kl\Big(\big\{X_{K\Delta+i}^{t_{i}^{\rmO}}\big\}_{i=1}^{w}\big| \big\{X_{K\Delta+i}^{t_{i}^{\rmI}}\big\}_{i=1}^{w}\Big\|\big\{Y_{K\Delta+i}^{t_{i}^{\rmO}}\big\}_{i=1}^{w} \big| \big\{Y_{K\Delta+i}^{t_{i}^{\rmI}}\big\}_{i=1}^{w}\Big) \nonumber\\
    &\qquad\quad+\kl \Big( \big\{X_{K\Delta+i}^{t_{i}^{\rmI}}\big\}_{i=1}^{w}\Big\| \big\{Y_{K\Delta+i}^{t_{i}^{\rmI}}\big\}_{i=1}^{w}\Big)\nonumber\\
    &\quad = \kl\Big(\big\{X_{K\Delta+i}^{t_{i}^{\rmO}}\big\}_{i=1}^{w}\big| \big\{X_{K\Delta+i}^{t_{i}^{\rmI}}\big\}_{i=1}^{w}\Big\|\big\{Y_{K\Delta+i}^{t_{i}^{\rmO}}\big\}_{i=1}^{w} \big| \big\{Y_{K\Delta+i}^{t_{i}^{\rmI}}\big\}_{i=1}^{w}\Big) \nonumber\\
    &\quad\qquad +\kl \Big( \big\{X_{K\Delta+i}^{t_{i}^{\rmI}}\big\}_{i=1}^{w-\Delta}\cup \big\{X_{K\Delta+i}^{t_{i}^{\rmI}}\big\}_{i=w-\Delta+1}^{w}\Big\| \big\{Y_{K\Delta+i}^{t_{i}^{\rmI}}\big\}_{i=1}^{w-\Delta}\cup \big\{Y_{K\Delta+i}^{t_{i}^{\rmI}}\big\}_{i=w-\Delta+1}^{w}\Big)\nonumber\\
    &\quad =\kl\Big(\big\{X_{K\Delta+i}^{t_{i}^{\rmO}}\big\}_{i=1}^{w}\big| \big\{X_{K\Delta+i}^{t_{i}^{\rmI}}\big\}_{i=1}^{w}\Big\|\big\{Y_{K\Delta+i}^{t_{i}^{\rmO}}\big\}_{i=1}^{w} \big| \big\{Y_{K\Delta+i}^{t_{i}^{\rmI}}\big\}_{i=1}^{w}\Big) \nonumber\\
    &\quad\qquad +\kl \Big( \big\{X_{K\Delta+i}^{t_{\Delta+i}^{\rmO}}\big\}_{i=1}^{w-\Delta}\cup \big\{X_{K\Delta+i}^{t_{i}^{\rmI}}\big\}_{i=w-\Delta+1}^{w}\Big\| \big\{Y_{K\Delta+i}^{t_{\Delta+i}^{\rmO}}\big\}_{i=1}^{w-\Delta}\cup \big\{Y_{K\Delta+i}^{t_{i}^{\rmI}}\big\}_{i=w-\Delta+1}^{w}\Big)\nonumber\\
    &\quad\leq \kl\Big(\big\{X_{K\Delta+i}^{t_{i}^{\rmO}}\big\}_{i=1}^{w}\big| \big\{X_{K\Delta+i}^{t_{i}^{\rmI}}\big\}_{i=1}^{w}\Big\|\big\{Y_{K\Delta+i}^{t_{i}^{\rmO}}\big\}_{i=1}^{w} \big| \big\{Y_{K\Delta+i}^{t_{i}^{\rmI}}\big\}_{i=1}^{w}\Big) \nonumber\\
    &\quad\qquad +\kl \Big( \big\{X_{(K-1)\Delta+i}^{t_{i}^{\rmO}}\big\}_{i=1}^{w}\Big\| \big\{Y_{(K-1)\Delta+i}^{t_{i}^{\rmO}}\big\}_{i=1}^{w}\Big)\nonumber\\
    &\quad\qquad +\kl \Big( \big\{X_{K\Delta+i}^{t_{i}^{\rmI}}\big\}_{i=w-\Delta+1}^{w}\Big|\big\{X_{K\Delta+i}^{t_{\Delta+i}^{\rmO}}\big\}_{i=1}^{w-\Delta}\Big\| \big\{Y_{K\Delta+i}^{t_{i}^{\rmI}}\big\}_{i=w-\Delta+1}^{w}\Big|\big\{Y_{K\Delta+i}^{t_{\Delta+i}^{\rmO}}\big\}_{i=1}^{w-\Delta}\Big),\label{ieq:f_up}
\end{align}
where the inequalities follows from the data processing inequality, and the second equality follows from the Circularity of Requirement~\ref{req:ar_step}. To derive the upper bound of the right-hand side of Eqn.~\eqref{ieq:f_up}, we can summing up the following recursive equation implied by Eqn.~\eqref{ieq:f_up}. In fact, we have that
\begin{align*}
    & \kl\Big(\big\{X_{k\Delta+i}^{t_{i}^{\rmO}}\big\}_{i=1}^{w}\Big\|\big\{Y_{k\Delta+i}^{t_{i}^{\rmO}}\big\}_{i=1}^{w}\Big)\nonumber\\
    &\quad\leq \kl\Big(\big\{X_{k\Delta+i}^{t_{i}^{\rmO}}\big\}_{i=1}^{w}\big| \big\{X_{k\Delta+i}^{t_{i}^{\rmI}}\big\}_{i=1}^{w}\Big\|\big\{Y_{k\Delta+i}^{t_{i}^{\rmO}}\big\}_{i=1}^{w} \big| \big\{Y_{k\Delta+i}^{t_{i}^{\rmI}}\big\}_{i=1}^{w}\Big) \nonumber\\
    &\quad\qquad +\kl \Big( \big\{X_{(k-1)\Delta+i}^{t_{i}^{\rmO}}\big\}_{i=1}^{w}\Big\| \big\{Y_{(k-1)\Delta+i}^{t_{i}^{\rmO}}\big\}_{i=1}^{w}\Big)\nonumber\\
    &\quad\qquad +\kl \Big( \big\{X_{k\Delta+i}^{t_{i}^{\rmI}}\big\}_{i=w-\Delta+1}^{w}\Big|\big\{X_{k\Delta+i}^{t_{\Delta+i}^{\rmO}}\big\}_{i=1}^{w-\Delta}\Big\| \big\{Y_{k\Delta+i}^{t_{i}^{\rmI}}\big\}_{i=w-\Delta+1}^{w}\Big|\big\{Y_{k\Delta+i}^{t_{\Delta+i}^{\rmO}}\big\}_{i=1}^{w-\Delta}\Big).
\end{align*}
This inequality indicates the recursive relationship of the error of the clips at different time steps.

\textbf{Step 2: Conclude the proof by aggregating the recursive formula.}

Summing the inequality about the recursive relationship from $k=1$ to $k=K$, we have that
\begin{align}
    & \kl\Big(\big\{X_{K\Delta+i}^{t_{i}^{\rmO}}\big\}_{i=1}^{w}\Big\|\big\{Y_{K\Delta+i}^{t_{i}^{\rmO}}\big\}_{i=1}^{w}\Big)\nonumber\\
    &\quad\leq \kl \Big( \big\{X_{i}^{t_{i}^{\rmO}}\big\}_{i=1}^{w}\Big\| \big\{Y_{i}^{t_{i}^{\rmO}}\big\}_{i=1}^{w}\Big)\nonumber\\
    &\quad\qquad +\sum_{k=1}^{K}\kl\Big(\big\{X_{k\Delta+i}^{t_{i}^{\rmO}}\big\}_{i=1}^{w}\big| \big\{X_{k\Delta+i}^{t_{i}^{\rmI}}\big\}_{i=1}^{w}\Big\|\big\{Y_{k\Delta+i}^{t_{i}^{\rmO}}\big\}_{i=1}^{w} \big| \big\{Y_{k\Delta+i}^{t_{i}^{\rmI}}\big\}_{i=1}^{w}\Big)\nonumber\\
    &\quad\qquad +\sum_{k=1}^{K}\kl \Big( \big\{X_{k\Delta+i}^{t_{i}^{\rmI}}\big\}_{i=w-\Delta+1}^{w}\Big|\big\{X_{k\Delta+i}^{t_{\Delta+i}^{\rmO}}\big\}_{i=1}^{w-\Delta}\Big\| \big\{Y_{k\Delta+i}^{t_{i}^{\rmI}}\big\}_{i=w-\Delta+1}^{w}\Big|\big\{Y_{k\Delta+i}^{t_{\Delta+i}^{\rmO}}\big\}_{i=1}^{w-\Delta}\Big).\label{ieq:recur_sum}
\end{align}
Then we upper bound the second and third terms in the right-hand side of Eqn.~\eqref{ieq:recur_sum}. Similar to Eqn.~\eqref{ieq:term_3}, we have that
\begin{align*}
    &\kl\Big(\big\{X_{k\Delta+i}^{t_{i}^{\rmO}}\big\}_{i=1}^{w}\big| \big\{X_{k\Delta+i}^{t_{i}^{\rmI}}\big\}_{i=1}^{w}\Big\|\big\{Y_{k\Delta+i}^{t_{i}^{\rmO}}\big\}_{i=1}^{w} \big| \big\{Y_{k\Delta+i}^{t_{i}^{\rmI}}\big\}_{i=1}^{w}\Big)\nonumber\\
    &\quad \leq \sum_{n=1}^{M_{\ar}}(t_{n}^{\ar}-t_{n-1}^{\ar})\bbE\Big[\big\|\nabla\log P_{X}^{\bart_{n}^{\ar}}(X_{k\Delta+1:k\Delta+w}^{\bart_{n}^{\ar}}\,|\,X_{\calR_{k\Delta+1}}^{0})-s_{\theta}(X_{k\Delta+1:k\Delta+w}^{\bart_{n}^{\ar}},\bart_{n}^{\ar},X_{\calR_{k\Delta+1}}^{0})\big\|^{2}\Big]\nonumber\\
    &\quad\qquad +wdL^{2}\sum_{n=1}^{M_{\ar}}(t_{n}^{\ar}-t_{n-1}^{\ar})^2\nonumber\\
        &\kl\Big( \big\{X_{k\Delta+i}^{t_{i}^{\rmI}}\big\}_{i=w-\Delta+1}^{w}\Big|\big\{X_{k\Delta+i}^{t_{\Delta+i}^{\rmO}}\big\}_{i=1}^{w-\Delta}\Big\| \big\{Y_{k\Delta+i}^{t_{i}^{\rmI}}\big\}_{i=w-\Delta+1}^{w}\Big|\big\{Y_{k\Delta+i}^{t_{\Delta+i}^{\rmO}}\big\}_{i=1}^{w-\Delta}\Big)\nonumber\\
        &\quad \leq (d\cdot i_{0}+B^{2})\exp(-T).
\end{align*}
Thus, we conclude the proof by combining these inequalities.

\section{Proof of Theorem~\ref{thm:low_bd}}\label{app:low_proof}
In the following, we would like to prove 
\begin{align*}
    \inf_{\hatP}\sup_{P\in\calS(s)}P\bigg(\tv(P,\hatP)\geq\frac{s}{2} \bigg)\geq\frac{1}{2}.
\end{align*}
Then Pinsker inequality implies that
\begin{align*}
    \inf_{\hatP}\sup_{P\in\calS(s)}P\bigg(\kl(P\,\|\,\hatP)\geq \frac{s^2}{2} \bigg)&\geq\inf_{\hatP}\sup_{P\in\calS(s)}P\bigg(2\big[\tv(P,\hatP)\big]^{2}\geq \frac{s^2}{2}\bigg)\\
    &=\inf_{\hatP}\sup_{P\in\calS(s)}P\bigg(\tv(P,\hatP)\geq\frac{s}{2} \bigg)\\
    &\geq \frac{1}{2}.
\end{align*}
We would like to follow the procedures of the proof of impossibility results in the non-parametric statistics~\citep{wainwright2019high}. We start the proof by constructing two distributions $P_0$ and $P_1$ as follows.
\begin{align}
    P_{0}(X=x,Y=y,Z=z)&=\frac{1}{8} \text{ for all }x,y,z\in\{0,1\},\label{eq:p0}\\
    P_{1}(X=x,Y=y,Z=z)&=P_{1}(X=x,Z=z)P_{1}(Y=y)=\frac{1}{2}P_{1}(X=x,Z=z)P_{1}(Y=y),\label{eq:p1}\\
    P_{1}(X=x,Z=1-x)&=\frac{\epsilon}{2},\quad P_{1}(X=x,Z=x)=\frac{1-\epsilon}{2}\nonumber,
\end{align}
where $\epsilon\in[0,1/2]$ is a hyperparameter. For $P_{0}$, $X,Y$ and $Z$ are independent Bernoulli($1/2$) random variables. For $P_{1}$, $Y$ is a independent Bernoulli($1/2$) with $X,Z$. The marginal distributions of $X$ and $Z$ are Bernoulli($1/2$), but these two variables are connected by a flip channel with flip probability $\epsilon$. We first verify that these two distributions are in $\calS(s)$. In fact, we have that
\begin{align*}
    I_{0}(X;Z|Y)=0,\quad I_{1}(X;Z|Y)=I_{1}(X;Z)=1-H(\epsilon),
\end{align*}
where $I_i$ is the mutual information with respect to the distribution $P_{i}$, and $H(x)=-x\log x-(1-x)\log(1-x)$ is the entropy of Bernoulli distribution. Obvisiously, $P_{0}\in\calS(s)$. We note that $H(x)\in[0,1]$ and is monotone on $[0,1/2]$. Thus, if $\epsilon\in[H^{-1}(1-s),1/2]$, $P_{1}\in\calS(s)$. In the following, we set $\epsilon=H^{-1}(1-s)$, then $I_{1}(X;Z|Y)=s$. A property of the constructed two distributions is that
\begin{align}
    P_{0}(X,Y)=P_{1}(X,Y),\quad P_{0}(Y,Z)=P_{1}(Y,Z).\label{eq:equal}
\end{align}
We also show that the total variation between them can be bounded as follows.
\begin{proposition}\label{prop:dist_bd}
    For the distributions defined in Eqn.~\eqref{eq:p0} and \eqref{eq:p1} with $\epsilon\in[0,1/2]$, we have that
    \begin{align*}
        \tv(P_{0},P_{1})\geq \frac{1}{2}\kl(P_{1}\,\|\,P_{0})=\frac{1}{2}I_{1}(X;Z|Y).
    \end{align*}
\end{proposition}
The proof is provided in Appendix~\ref{app:dist_bd}. Then we consider the estimate $\hatP$ based on the data generated by these two distributions. For any $\hatP$, we define a classifier $\psi$ that determines the data set is generated by $P_0$ or $P_1$ as
\begin{align*}
    \psi(\hatP)=\argmin_{i\in\{0,1\}} \tv(\hatP,P_i).
\end{align*}
If $\psi(\hatP)\neq i$ for $i\in\{0,1\}$, then the triangle inequality shows that
\begin{align*}
    \tv(\hatP,P_{i})\geq \frac{1}{2}I_{1}(X;Z|Y)=\frac{s}{2}.
\end{align*}
Thus, we have that
\begin{align}
    \inf_{\hatP}\!\!\sup_{P\in\calS(s)}\!\!P\bigg(\tv(\hatP,P)\geq\frac{s}{2}\bigg)\geq \inf_{\hatP}\!\!\sup_{i\in\{0,1\}}\!\!P_{i}\bigg(\tv(\hatP,P_{i})\geq\frac{s}{2}\bigg)\geq \inf_{\hatP}\!\!\sup_{i\in\{0,1\}}\!\!P_{i}\big(\psi(\hatP)\neq i\big).\label{ieq:lbd}
\end{align}
We denote the data distribution as $P_{i}^{\calD}$, i.e.,
\begin{align*}
    & P_{i}^{\calD}\big(\{X_i=x_i,Y_i=y_i\}_{i=1}^{N}\cap\{Y_i=y_i,Z_i=z_i\}_{i=N+1}^{2N}\big)\\
    &\quad=\prod_{i=1}^{N}P_{i}(X_i=x_i,Y_i=y_i)\cdot\prod_{i=N+1}^{2N}P_{i}(Y_i=y_i,Z_i=z_i).
\end{align*}
Then Eqn.~\eqref{eq:equal} implies that $P_{0}^{\calD}=P_{1}^{\calD}$. Neyman-Pearson Theorem shows that any test $\psi$ has the following error probability.
\begin{align*}
    P_{1}(\psi=0)+P_{0}(\psi=1)\geq \int \min(\rmd P_{0}^{\calD},\rmd P_{1}^{\calD})=1.
\end{align*}
Then we can further lower bound Eqn.~\eqref{ieq:lbd} as 
\begin{align*}
    \inf_{\hatP}\!\!\sup_{i\in\{0,1\}}\!\!P_{i}\big(\psi(\hatP)\neq i\big)\geq \inf_{\psi}\!\!\sup_{i\in\{0,1\}}\!\!P_{i}\big(\psi\neq i\big)\geq \frac{1}{2}.
\end{align*}
Thus, we conclude the proof of Theorem~\ref{thm:low_bd}.

\section{Supporting Lemmas}\label{app:lem}
\begin{lemma}[Lemma 6 in \cite{chen2023improved}]\label{lem:diff}
    Consider the following two Ito processes
    \begin{align*}
        \rmd X^{t}=F_{1}(X^{t},t)\rmd t+g(t)\rmd B^{t},\quad X^0=a\\
        \rmd Y^{t}=F_{2}(Y^{t},t)\rmd t+g(t)\rmd B^{t},\quad Y^0=a,
    \end{align*}
    where $F_1$, $F_2$, and $g$ are continuous functions and may depend on $a$, We assume the uniqueness and regularity condition:
    \begin{itemize}
        \item These two \ac{sde}s have unique solutions.
        \item The processes $X^t$ and $Y^t$ admit densities $p^t,q^t\in C^2(\bbR^{d})$ for $t>0$.
    \end{itemize}
    Define the relative Fisher information between $p^t$ and $q^t$ by
    \begin{align*}
        J(p^t\,\|\,q^t)=\int p^t(x)\bigg\|\nabla \log\frac{p^t(x)}{q^t(x)}\bigg\|^{2}\rmd x.
    \end{align*}
    Then for any $t>0$, the evolution of $\kl(p^t\|q^t)$ is given by
    \begin{align*}
        \frac{\rmd}{\rmd t}\kl(p^t\,\|\,q^t)=-g(t)^2J(p^t\,\|\,q^t)+\bbE_{p^t}\bigg[\bigg\langle F_{1}(X^t,t)-F_{2}(X^t,t),\nabla\log\frac{p^t(X^t)}{q^t(X^t)}\bigg\rangle\bigg].
    \end{align*}
\end{lemma}

\begin{lemma}[Theorem 7 in \cite{verdu2014total}]\label{lem:rev_pinsker}
    For two distributions $P,Q\in\calP(\Omega)$, we define
    \begin{align*}
        \beta_{1}^{-1}=\sup_{w\in\Omega}\frac{\rmd P}{\rmd Q}(w).
    \end{align*}
    Then we have that
    \begin{align*}
        \tv(P,Q)\geq \frac{1-\beta_{1}}{\log_{2} 1/\beta_{1}}\kl(P\,\|\,Q).
    \end{align*}
\end{lemma}

\begin{lemma}[Lemmas 9 and 11 in \cite{chen2023improved}]\label{lem:init}
    For any distribution $P_X$ on $\bbR^{d}$ that has finite second moment, i.e., $\bbE_{P_{X}}\|X\|^{2} < \infty$, and $X^{t}$ evolves as
    \begin{align*}
        \rmd X^{t}&=-\frac{1}{2}X^{t}\rmd t+\rmd B^{t},\quad X^{0}\sim P_{X},
    \end{align*}
    where $B^t$ is a Brownian motion. We denote the distribution of $X^{t}$ as $P^{t}$. Then we have that
    \begin{align*}
        \kl(P^{T}\|\calN(0,I))\leq (d+\bbE_{P_{X}}\|X\|^{2})\cdot \exp(-T).
    \end{align*}
    For any $0\leq t\leq s\leq T$, we define $\alpha_{t,s}=\exp(-\frac{1}{2}(s-t))$. Then we have that
    \begin{align*}
        &\bbE\Big[\big\|\nabla\log P^{t}(X^t)-\nabla\log P^s(X^s)\big\|^{2}\Big]\\
        &\quad\leq 4\bbE\Big[\big\|\nabla\log P^{t}(X^t)-\nabla\log P^t(\alpha_{t,s}^{-1} X^s)\big\|^{2}\Big]+2(1-\alpha_{t,s}^{-1})^{2}\bbE\Big[\big\|\nabla\log P^{t}(X^t)\big\|^{2}\Big].
    \end{align*}
\end{lemma}
\begin{lemma}[\cite{chewi2024analysis}]\label{lem:bd}
    Let $P\in C^{1}(\bbR^{d})$ be a probability distribution. If $\nabla\log P$ is $L$-Lipschitz, then we have that
    \begin{align*}
        \bbE_{P}\Big[\big\|\nabla\log P(X)\big\|^{2}\Big]\leq d\cdot L.
    \end{align*}
    
\end{lemma}
\section{Proof of Supporting Propositions}\label{app:supp}
\subsection{Proof of Proposition~\ref{prop:bless}}\label{app:bless}
We first decompose $I(X,f(X),g(X);Z|Y)$ in two ways as follows.
\begin{align*}
    I(X,f(X),g(X);Z|Y)&=I(f(X);Z|Y)+I(X;Z|Y,f(X))+I(g(X);Z|Y,f(X),X)\\
    &=I(f(X);Z|Y)+I(X;Z|Y,f(X)),
\end{align*}
where the first equality results from the chain rule, and the second equality results from the fact that $I(g(X);Z|Y,f(X),X)=0$. Similarly, we have that
\begin{align*}
    I(X,f(X),g(X);Z|Y)=I(g(X);Z|Y)+I(X;Z|Y,g(X)).
\end{align*}
Thus, we have that
\begin{align*}
    I(X;Z|Y,g(X)) - I(X;Z|Y,f(X)) = I(f(X);Z|Y) - I(g(X);Z|Y).
\end{align*}
In the following, we will show that the right-hand side of this equation is non-negative. In fact, the chain rule shows that
\begin{align*}
    I(g(X),f(X);Z|Y)= I(g(X);Z|Y)+I(f(X);Z|Y,g(X)).
\end{align*}
Similarly, we have that
\begin{align*}
    I(g(X),f(X);Z|Y)= I(f(X);Z|Y)+I(g(X);Z|Y,f(X))=I(f(X);Z|Y),
\end{align*}
where the last equality results from that $g(x)=h(f(x))$ for all $x$. Thus, we have that
\begin{align*}
    I(f(X);Z|Y) - I(g(X);Z|Y)=I(f(X);Z|Y,g(X))\geq 0.
\end{align*}
    We conclude the proof of Proposition~\ref{prop:bless}.
\subsection{Proof of Proposition~\ref{prop:h_g}}\label{app:h_g}
We begin the proof by showing the recursive relationship for $\calH$. We note that
        \begin{align}
            &\calH\Big(\big\{Y_{i+j}^{t_{j}^{\rmI}}\big\}_{j=1}^{w}\Big)= \calH\Big(\big\{Y_{i-\Delta+j}^{t_{j}^{\rmI}}\big\}_{j=1}^{w}\Big)\cup \{Y_{i-\Delta+1:i}^{0}\}\cup \big\{Y_{i-\Delta+j}^{t_{j}^{\rmI}}\big\}_{j=1}^{w},\label{eq:h_h}
        \end{align}
        the equality follows from the definition of $\calH(\cdot)$ in Eqn.~\eqref{eq:h}. Then we have
        \begin{align*}
            &\calG\Big(\big\{Y_{i-\Delta+j}^{t_{j}^{\rmO}}\big\}_{j=1}^{w}\Big)\\
            &\quad = \calH\Big(\big\{Y_{i-\Delta+j}^{t_{j}^{\rmI}}\big\}_{j=1}^{w}\Big)\cup\big\{Y_{i-\Delta+j}^{t_{j}^{\rmI}}\big\}_{j=1}^{w}\cup \big\{Y_{i-\Delta+j}^{t_{j}^{\rmO}}\big\}_{j=1}^{w}\\
            &\quad = \calH\Big(\big\{Y_{i-\Delta+j}^{t_{j}^{\rmI}}\big\}_{j=1}^{w}\Big)\cup\big\{Y_{i-\Delta+j}^{t_{j}^{\rmI}}\big\}_{j=1}^{w}\cup \big\{Y_{i-\Delta+j}^{t_{j}^{\rmO}}\big\}_{j=1}^{\Delta}\cup\big\{Y_{i-\Delta+j}^{t_{j}^{\rmO}}\big\}_{j=\Delta+1}^{w}\\
            &\quad = \calH\Big(\big\{Y_{i-\Delta+j}^{t_{j}^{\rmI}}\big\}_{j=1}^{w}\Big)\cup\big\{Y_{i-\Delta+j}^{t_{j}^{\rmI}}\big\}_{j=1}^{w}\cup \big\{Y_{i-\Delta+j}^{0}\big\}_{j=1}^{\Delta}\cup\big\{Y_{i+j}^{t_{j}^{\rmI}}\big\}_{j=1}^{w-\Delta}\\
            &\quad = \calH\Big(\big\{Y_{i+j}^{t_{j}^{\rmI}}\big\}_{j=1}^{w}\Big)\cup\big\{Y_{i+j}^{t_{j}^{\rmI}}\big\}_{j=1}^{w-\Delta},
        \end{align*}
        where the first equality follows from the definition of $\calG$ in Eqn.~\eqref{eq:g}, the third equality follows from $0-T$ boundary and circularity in Requirement~\ref{req:ar_step}, and the last equality follows from Eqn.~\eqref{eq:h_h}. Thus, we conclude the proof of Proposition~\ref{prop:h_g}.
\subsection{Proof of Proposition~\ref{prop:dist_bd}}\label{app:dist_bd}
First, we note that the KL-divergence between these two distributions is 
\begin{align*}
    \kl(P_{1}\|P_{0})=I_{1}(X;Z|Y).
\end{align*}
Lemma~\ref{lem:rev_pinsker} shows that
\begin{align*}
    \tv(P_{0},P_{1})\geq \frac{1-\beta_{1}}{\log_{2} 1/\beta_{1}}\kl(P_{1}\,\|\,P_{0}),
\end{align*}
where $\beta_{1}^{-1}=2(1-\epsilon)$. To prove the desired result, it remains to show that
\begin{align*}
    \frac{1-\beta_{1}}{\log_{2} 1/\beta_{1}}=\frac{1-2\epsilon}{2(1-\epsilon)\log_{2}\big(2(1-\epsilon)\big)}\geq \frac{1}{2}.
\end{align*}
Then we define the function
\begin{align*}
    f(\epsilon) = 1-2\epsilon-(1-\epsilon)\log_{2}\big(2(1-\epsilon)\big).
\end{align*}
The derivative of this function is that
\begin{align*}
    f^{\prime}(\epsilon)=-1+\frac{1}{\log 2}-\frac{\log (1-\epsilon)}{\log 2},
\end{align*}
which is a monotone function. Since $f^{\prime}(0)>0>f^{\prime}(1/2)$, we have that
\begin{align*}
    \inf_{x\in[0,1/2]}f(x)=\min\{f(0),f(1/2)\}\geq 0.
\end{align*}
Thus, we conclude the proof of Proposition~\ref{prop:dist_bd}.

\subsection{Proof of Proposition~\ref{prop:lips}}\label{app:lips}
For the difference between the scores to bound, we have that
\begin{align*}
    &\bbE\Big[\big\|\nabla\log P^{u}(X^u)-\nabla\log P^v(X^v)\big\|^{2}\Big]\\
        &\quad\leq 4\bbE\Big[\big\|\nabla\log P^{u}(X^u)-\nabla\log P^u(\alpha_{u,v}^{-1} X^v)\big\|^{2}\Big]+2(1-\alpha_{u,v}^{-1})^{2}\bbE\Big[\big\|\nabla\log P^{u}(X^u)\big\|^{2}\Big]\\
        &\quad \lesssim L^2\cdot \bbE\Big[\big\|X^u-\alpha_{u,v}^{-1} X^v\big\|^{2}\Big]+d\cdot L (1-\alpha_{u,v}^{-1})^2\\
        &\quad \lesssim d\cdot L^{2}\big(\exp(v-u)-1\big)+dL(v-u)^2\\
        &\quad\lesssim dL^2(v-u),
\end{align*}
where the first inequality results from Lemma~\ref{lem:init}, the second inequality results from the Lipschitzness and Lemma~\ref{lem:bd}, the third inequality results from the definition of $X^t$, and the last inequality results from that $v-u\leq 1$.
\section{Other Methods Covered By \texttt{Meta-\ac{ar}\ac{vdm}}}\label{app:other_methods}

We list some methods that are special cases of our framework here to demonstrate the universality of \texttt{Meta-\ac{ar}\ac{vdm}}. The methods we list here are a strict subset of all the methods included by our framework.
\begin{itemize}
    \item ART-V~\citep{weng2024art}: It denoises under same noise-level, i.e., $t _i^{\rmO}=0, t _i^{\rmI}=T$ for all $i\in[w]$. The reference frames are an anchor frame and the last two, i.e., $\mathcal{R}(i)=\{0,i-1,i-2 \}$.
    \item GameNGen~\citep{valevski2024diffusion}: It repurposes T2I model in AR style to generate videos, i.e., $t _1^{\rmO}=0, t _1^{\rmI}=T$ and $w=1$. It conditions on previous $64$ frames, i.e., $\mathcal{R}(i)=\{i-64,\cdots,i-1\}$.
    \item StreamingT2V~\citep{henschel2024streamingt2v}: It denoises 16 frames with the same noise level, i.e., $t _i^{\rmO}=0, t _i^{\rmI}=T$ for all $i\in[16]$. The reference frames are anchor frames and the last 8 frames, i.e., $\mathcal{R}(i)=[N_a]\cup \{i-8,\cdots,i-1\}$. Here we use $N_a$ anchor frames.
\end{itemize}

\section{Discussion of Noise Initialization Errors, Score Estimation Errors, and Discretization Errors in Theorem~\ref{thm:main}}\label{app:err_discussion}

The \emph{noise initialization error} originates from the fact that the random variables $X_{1:w}^{T}$ and $X_{w-\Delta+1:w}^{T}$ are not exactly Gaussian. The difference between them and the Gaussian random variables contributes to this error, which decreases exponentially with increasing $T$. The \emph{score estimation error} originates from the difference between the true score function and the estimate from the network. The concrete rate of this term is not relevant to the main topic of our work, which is considered in \cite{chen2023score}. Intuitively, this term should decrease with increasing number of training data points. The \emph{discretization error} originates from the fact that we discretize the trajectory with the Euler-Maruyama scheme. We can see that this term decreases with increasing discretization steps $M_{\init}$ and $M_{\ar}$. If we adopt uniform discretization, the discretization error will scale as $O(M_{\init}^{-1})$ or $O(M_{\ar}^{-1})$. This error term can be improved with more advanced numerical methods, such as DPM-solver~\citep{lu2022dpm}.

\section{Demonstration of Data Model for Lower Bounds}\label{app:lb_fig}
\begin{figure}[H]
\centering
 \includegraphics[width=0.7\textwidth]{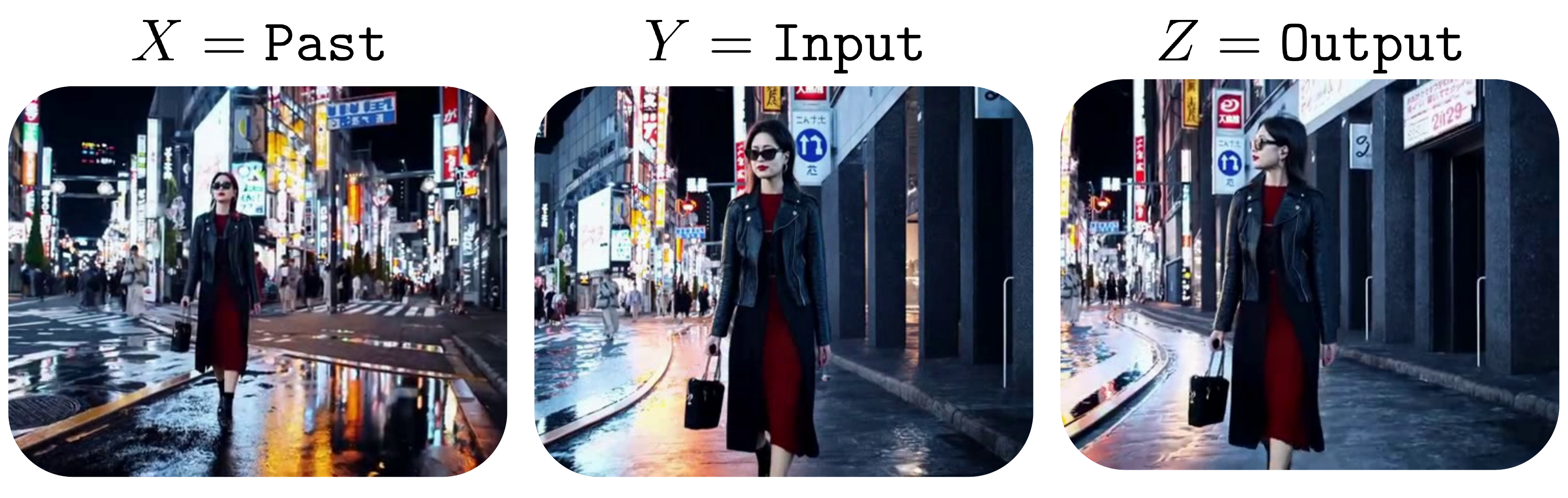}

\caption{The simplified setting for the proof of lower bounds. We adopt three random variables $X,Y,Z$ to represent \texttt{Past}, \texttt{Input}, and \texttt{Output}, respectively.}
\label{fig:lb}
\end{figure}

\section{Additional Demonstration Figures}\label{app:demo}

This section provides detailed visual demonstrations of successful and failed retrieval cases in both DMLab and Minecraft environments.

\begin{figure}[H]
    \centering
    \begin{subfigure}{0.8\linewidth}
        \includegraphics[width=\linewidth]{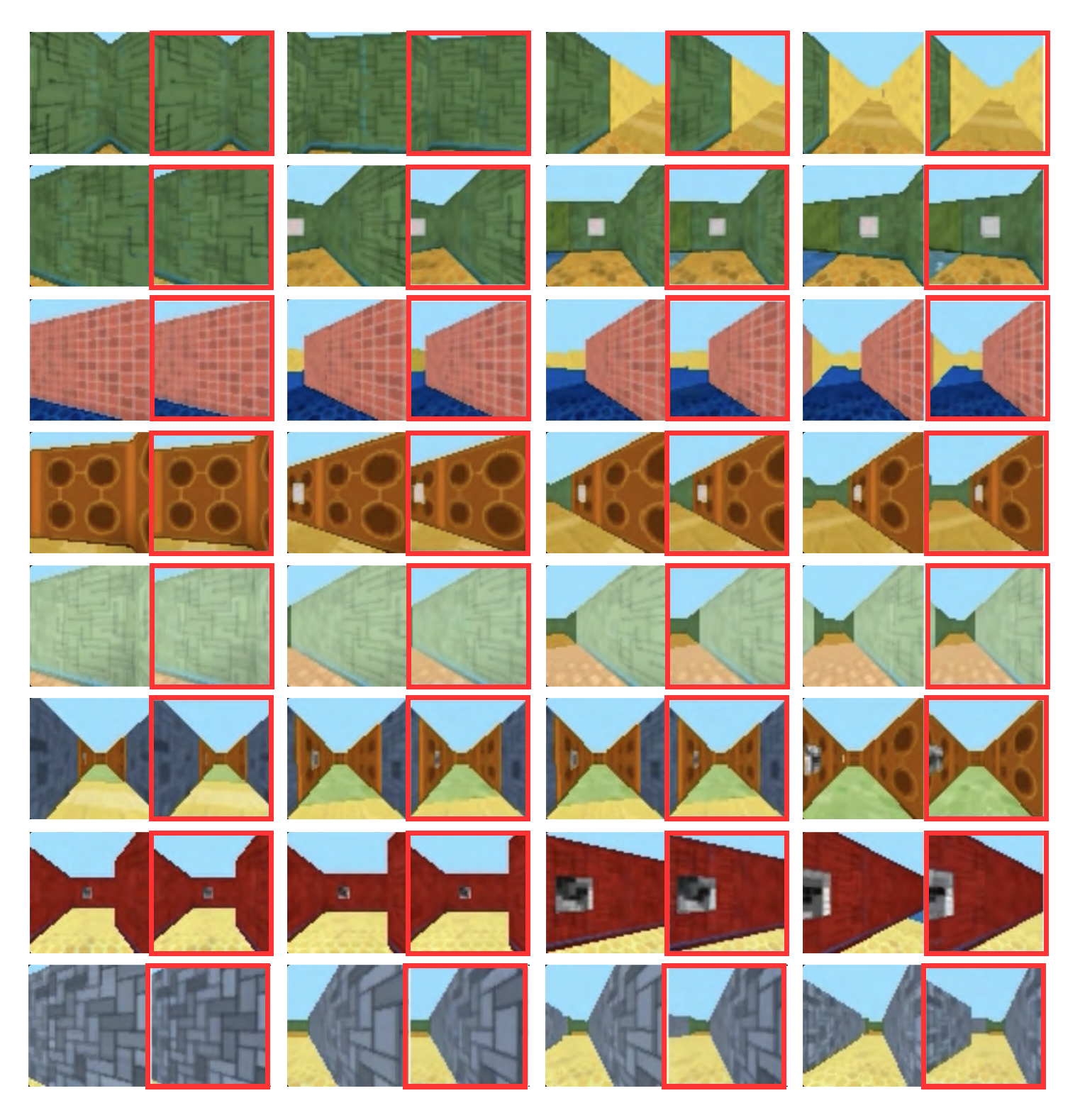}
        \caption{Successful Retrieval on DMLab}
    \end{subfigure}

    \begin{subfigure}{0.8\linewidth}
        \includegraphics[width=\linewidth]{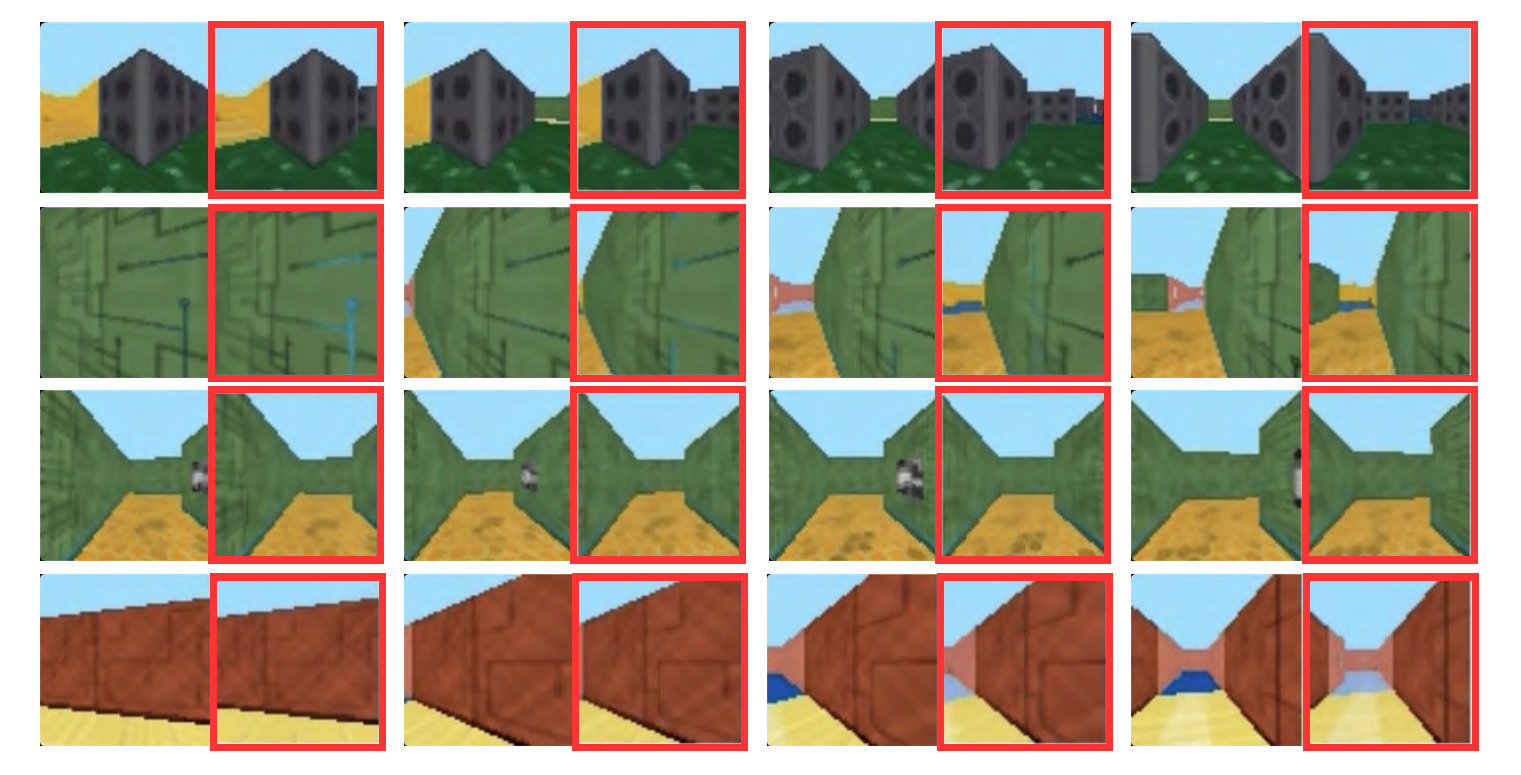}
        \caption{Failed Retrieval on DMLab}
    \end{subfigure}
    \caption{Recall demonstrations on DMLab. The left frames represent the expected
    ground truth, while the right frames, outlined with a red square, are
    generated by the model.}
    \label{fig:dmlab_recall_demo_dmlab}
\end{figure}

\begin{figure}[H]
    \centering
    \begin{subfigure}{0.8\linewidth}
        \includegraphics[width=\linewidth]{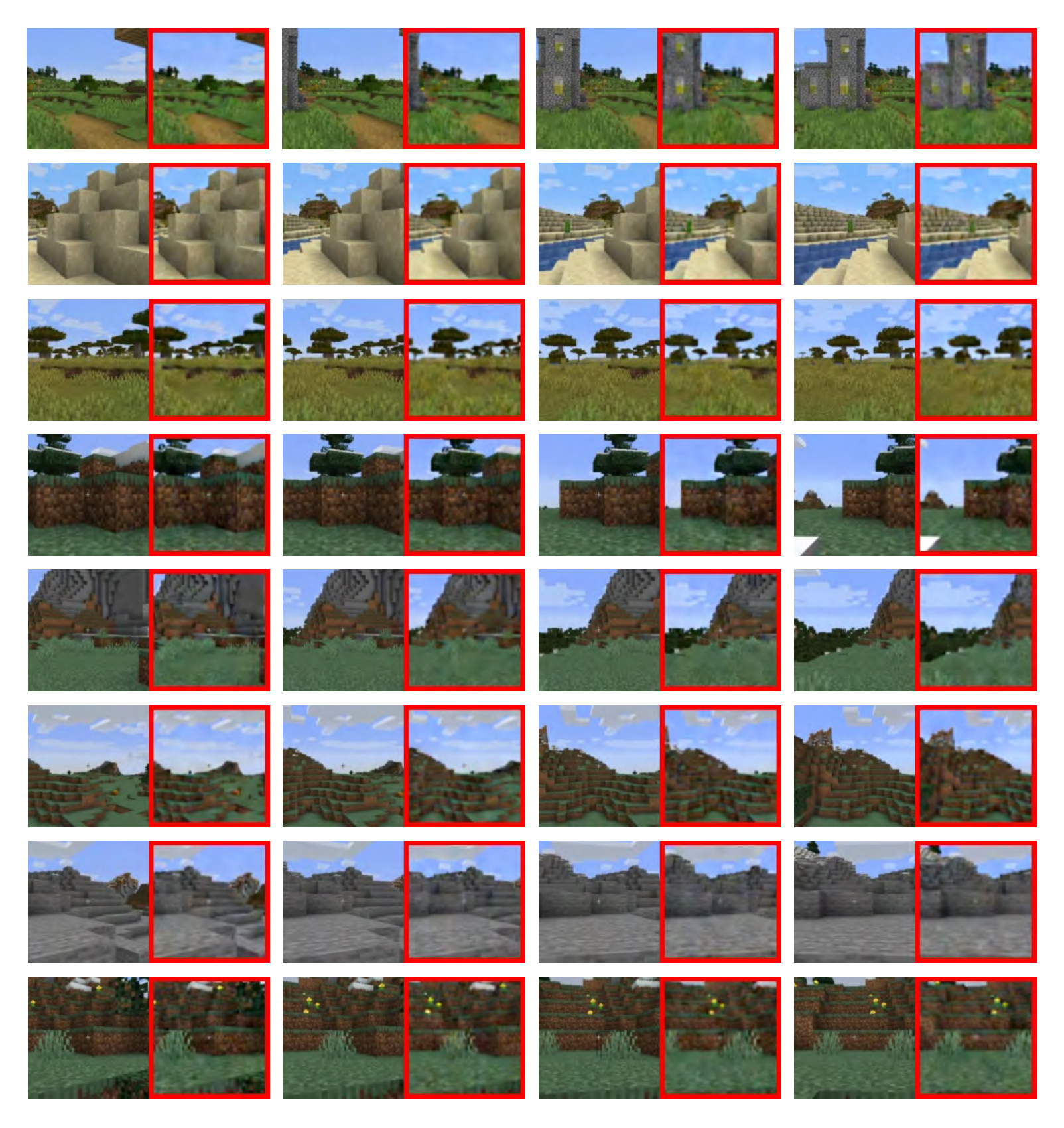}
        \caption{Successful Retrieval on Minecraft}
    \end{subfigure}

    \begin{subfigure}{0.8\linewidth}
        \includegraphics[width=\linewidth]{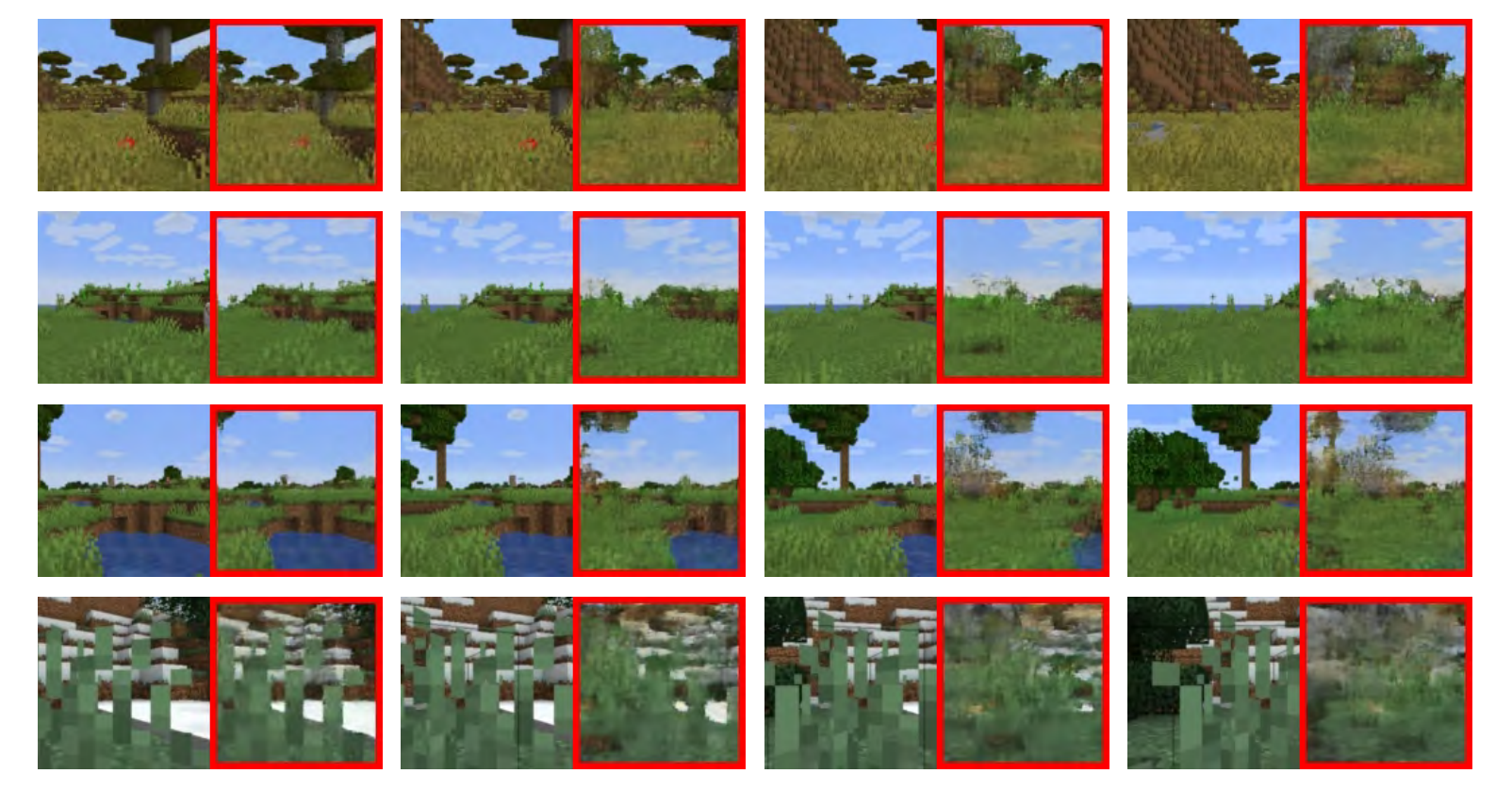}
        \caption{Failed Retrieval on Minecraft}
    \end{subfigure}
    \caption{Recall demonstrations on Minecraft. The left frames represent the expected
    ground truth, while the right frames, outlined with a red square, are
    generated by the model. The first 4 frames without red squares are provided context.}
    \label{fig:dmlab_recall_demo_minecraft}
\end{figure}

\section{Additional Experimental Results}
\label{app:additional_experiments}

This appendix contains additional experimental results for DMLab and Minecraft environments.

\begin{figure}[h]
\centering
\begin{subfigure}{0.45\textwidth}
\centering
\includegraphics[width=\textwidth]{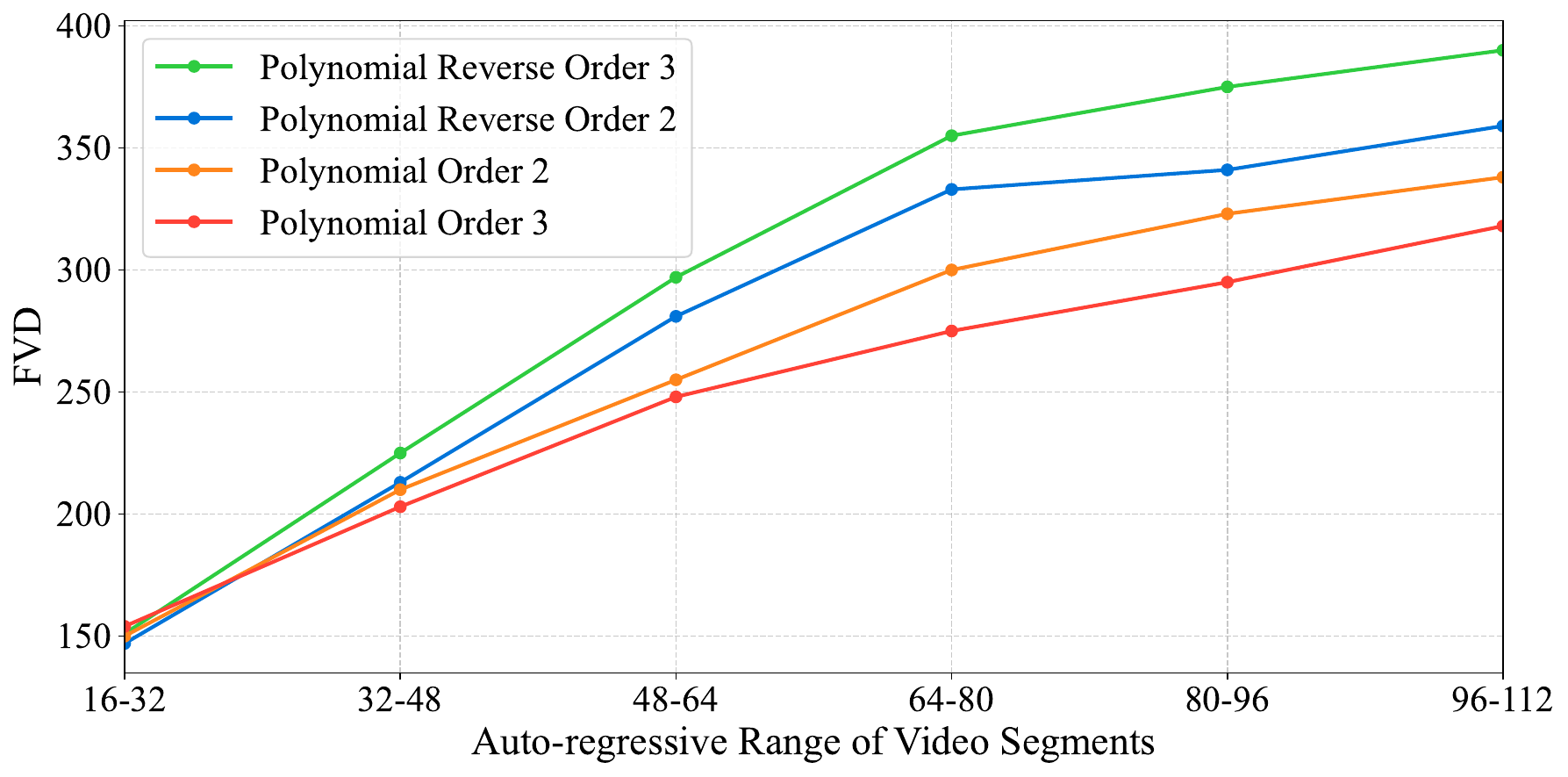}
\caption{DMLab}
\label{fig:td_diff_timesteps_dmlab}
\end{subfigure}
\hfill
\begin{subfigure}{0.45\textwidth}
\centering
\includegraphics[width=\textwidth]{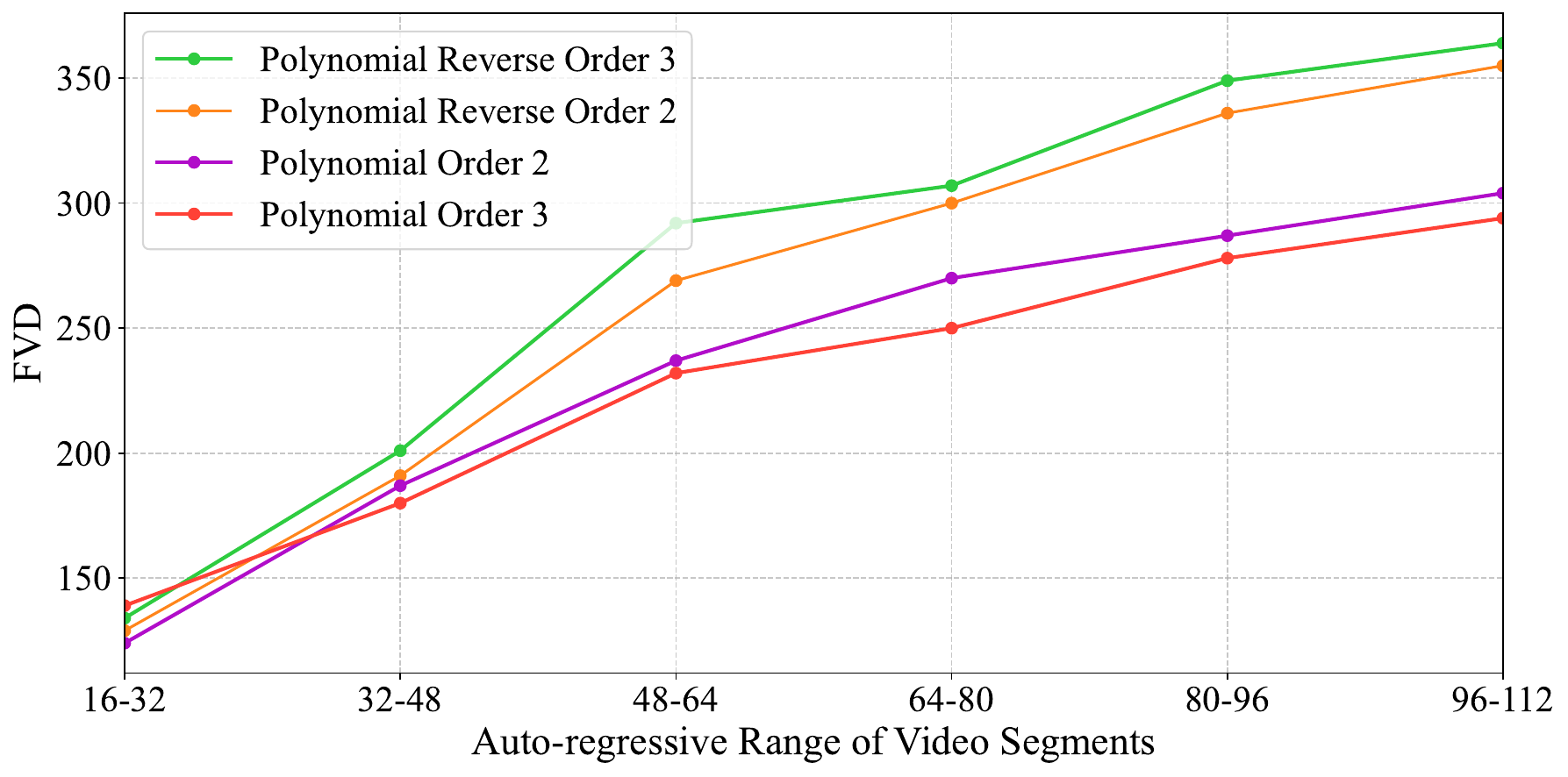}
\caption{Minecraft}
\label{fig:td_diff_timesteps_minecraft}
\end{subfigure}
\caption{Comparison of FVD across different timestep sets in DMLab and Minecraft. Polynomial Order is ($*$, 1), while Polynomial Reverse Order is ($*$, 2).}
\label{fig:td_diff_timesteps_appendix}
\end{figure}

\end{document}